\documentclass[5p]{elsarticle}

\usepackage{amsmath}
\usepackage{graphicx}
\usepackage{verbatim}
\usepackage[justification=centering]{caption}
\usepackage{enumitem}
\setlist[itemize]{noitemsep, nolistsep}
\usepackage{subcaption}
\usepackage{xcolor}
\usepackage{lineno,hyperref}
\usepackage[graphicx]{realboxes}
\usepackage{multirow}
\usepackage{rotating}
\modulolinenumbers[5]
\usepackage{mathtools}
\usepackage{amssymb}
\usepackage{float}
\usepackage{leftidx}
\usepackage{multicol}	
\usepackage{longtable}	
\usepackage{threeparttablex}    
\usepackage{array}
\usepackage{pifont}

\newcommand{\cmark}{\ding{51}}
\newcommand{\xmark}{\ding{55}}

\DeclareMathOperator{\atantwo}{atan2}

\journal{Robotics and Autonomous Systems}

\begin{document}

\begin{frontmatter}

\title{Real-Time Deep Learning Approach to Visual Servo Control and Grasp Detection for Autonomous Robotic Manipulation}

\author[address1]{Eduardo Godinho Ribeiro\corref{corauthor}}
\ead{eduardogr@usp.br}
\cortext[corauthor]{Corresponding author}
\author[address1]{Raul de Queiroz Mendes}
\ead{raulmendes@usp.br}
\author[address1]{Valdir Grassi Jr}
\ead{vgrassi@usp.br}

\address[address1]{Department of Electrical and Computer Engineering, S\~{a}o Carlos School of Engineering, University of S\~{a}o Paulo, Brazil}

\begin{abstract}

Robots still cannot perform everyday manipulation tasks, such as grasping, with the same dexterity as humans do. In order to explore the potential of supervised deep learning for robotic grasping in unstructured and dynamic environments, this work addresses the visual perception phase involved in the task. This phase involves the processing of visual data to obtain the location of the object to be grasped, its pose and the points at which the robot’s grippers must make contact to ensure a stable grasp. For this, the Cornell Grasping Dataset (CGD) is used to train a Convolutional Neural Network (CNN) that is able to consider these three stages simultaneously. In other words, having an image of the robot’s workspace, containing a certain object, the network predicts a grasp rectangle that symbolizes the position, orientation and opening of the robot’s parallel grippers the instant before its closing. In addition to this network, which runs in real-time, another network is designed, so that it is possible to deal with situations in which the object moves in the environment. Therefore, the second convolutional network is trained to perform a visual servo control, ensuring that the object remains in the robot’s field of view. This network predicts the proportional values of the linear and angular velocities that the camera must have to ensure the object is in the image processed by the grasp network. The dataset used for training was automatically generated by a Kinova Gen3 robotic manipulator with seven Degrees of Freedom (DoF). The robot is also used to evaluate the applicability in real-time and obtain practical results from the designed algorithms. Moreover, the offline results obtained through test sets are also analyzed and discussed regarding their efficiency and processing speed. The developed controller is able to achieve a millimeter accuracy in the final position considering a target object seen for the first time. To the best of our knowledge, we have not found in the literature other works that achieve such precision with a controller learned from scratch. Thus, this work presents a new system for autonomous robotic manipulation, with the ability to generalize to different objects and with high processing speed, which allows its application in real robotic systems.

\end{abstract}


\end{frontmatter}


\section{Introduction}

Object grasping is a canonical research area in robotic manipulation. However, since it is regarded as an open problem of robotics \cite{kumra2017robotic}, its relevance and importance extend to the present day, as can be noted in the lack of dexterity with which robots manipulate complex objects \cite{ribeiro2019fast}.

There are infinite candidate grasps that can be applied to an object. Thus, as identified by Bohg et al. \cite{bohg2014data}, a good subset of grasp hypotheses for finding the ideal grasp configuration comprises the type of task that the robot will execute, the features of the target object, what type of prior knowledge about the object is available, what type of hand is used, and, finally, the grasp synthesis.

Grasp synthesis is the core of the robotic grasping problem, as it refers to the task of finding points in the object that configure appropriate grasp choices. These are the points at which grippers must make contact with the object, ensuring that the action of external forces does not lead it to instability and satisfies a set of relevant criteria for the grasping task \cite{bohg2014data}.

Grasp synthesis methods can generally be categorized as analytic or data-driven. Analytic methods are those that construct force closure with a dexterous and stable multi-fingered hand with a certain dynamic behavior \cite{bohg2014data, shimoga1996robot}. Grasp synthesis is formulated as a restricted optimization problem that uses criteria such as kinematic, geometric, or dynamic formulations to measure the described properties \cite{shimoga1996robot}. These methods usually assume the existence of a precise geometric model of the object to be grasped, something that is not always possible. Besides, the surface properties or friction coefficients of the object, as well as its weight, center of mass, among others, may not be available \cite{bohg2014data}. 

Data-based methods are built upon the search of candidates for grasping and the classification of these candidates by some metric. This process generally assumes the existence of previous experience of grasping, provided by means of heuristics or learning \cite{bohg2014data}. In other words, these methods require annotations of those candidates that are considered correct \cite{rubert2017relevance}, to serve as a model for the algorithms used. Therefore, one must generate these annotations in some way, either by means of real robots \cite{levine2018learning, pinto2016supersizing}, simulations \cite{kappler2015leveraging} or direct annotations on images \cite{lenz2015deep}.

Among the works that use empirical methods, many deal with some kind of visual information in their algorithms. The main benefit of using images for grasp synthesis is its independence from 3D models of the target objects. In real scenarios, it is often difficult to obtain an accurate and complete 3D model of a first time seen object \cite{saxena2008robotic}.

Du, Wang and Lian \cite{du2019vision} assume that the vision-based robotic grasping system is composed of four main steps, namely, target object localization, object pose estimation, grasp detection (synthesis) and grasp planning. The first three steps can be performed simultaneously considering a system based on Convolutional Neural Network (CNN) that receives an image of the object as input and predicts a grasping rectangle as output.

Regarding the grasp planning phase, in which the manipulator finds the best path to the object, the robotic grasping system must overcome some challenges to be applicable in the real-world. It should be able to adapt to changes in the workspace and consider dynamic objects, using visual feedback. Therefore, the robot must track the object, so that it does not leave the camera's field of view, allowing localization, pose estimation and grasp synthesis to be performed in a reactive way.

Most of the approaches for the robotic grasping task perform one-shot grasp detection and cannot respond to changes in the environment. Thus, the insertion of visual feedback in grasp systems is desirable, since it makes it robust to perceptual noise, object movement, and kinematic inaccuracies \cite{viereck2017learning}. However, even with modern hardware, classical approaches require a prohibitive amount of time for closed-loop applications, and only perform adjustments of force, without visual feedback \cite{lampe2013acquiring}.

Thus, some works started to include a stage of Visual Servoing (VS) to deal with possible disturbances during grasp execution \cite{kragic2002survey}. However, this technique heavily relies on the extraction and tracking of features, often based on 3D models, camera parameters and other information that must be known \textit{a priori}. Although recent approaches aim to minimize the work of feature design, control modeling, and the need for \textit{a priori} information, they fail to achieve all of these advantages simultaneously. 

A grasping system that can combine the benefits of automatic grasp detection and that receives visual feedback to deal with unknown dynamic objects, in real-time, is one of the goals in robotics. This work aims to advance towards this goal by designing a real-time, real-world, and reactive autonomous robotic manipulation system.

According to the classification proposed by Bohg et al. \cite{bohg2014data}, our grasping system can be classified as data-driven, based on labeled training data. The object features are 2D (images) and, after training, no prior knowledge of the objects is required. We only consider parallel grippers, hence dexterous manipulation is beyond our scope.  Using a CNN, we detect grasps as grasp rectangles in the image, which indicate the position, orientation and opening of the robot’s gripper.

We also propose a visual servoing system to handle visual feedback. It is trained from scratch using a neural network and can be applied to unknown objects without any 3D model or camera parameters information. With an image of the target object seen from a particular pose, the network predicts velocity signals to be applied to the camera so that it keeps the object in the field of view during grasping.

Our design choices are motivated by three principles: simplicity, generalization and modularization. The networks developed for grasping and visual servoing are extremely lightweight and fast, and can be applied in scenarios other than those found in training. Moreover, we have strong inspirations to divide the complete system into two distinct modules. Firstly, this division makes the method agnostic to robotic platforms, unlike reinforcement learning \cite{saxena2017exploring}, and allows the developed network for visual servoing to be applied to other tasks besides grasping. Also, errors in the prediction of one of the networks do not induce errors in the other, so that modularization leads to an increase in the system's robustness.

The main contributions of this work are:
\begin{itemize}[noitemsep, nosep]
    \item Comprehensive review of works that use the Cornell Grasping Dataset (CGD) for static grasping; works that introduce learning algorithms for visual servoing; and works that address dynamic and reactive grasping.
    \item Design of a convolutional network for grasp rectangle prediction that achieves state-of-the-art speed in the CGD, and implementation of the network in a real robot, as an extension of a published conference paper \cite{ribeiro2019fast}.
    \item Design of four models of convolutional neural networks that can execute visual servoing in different target objects, by generating a proportional velocity signal based only on the reference image and the current image. The simplest model achieves state-of-the-art positioning error considering a controller learned from scratch dealing with a first-seen object.
    \item Construction of a large dataset for end-to-end visual servo control using a Kinova Gen3 robot. The dataset is publicly available. \footnote{https://github.com/RauldeQueirozMendes/VSDataset}
    \item Implementation of all algorithms in a final grasping system able to consider distinct and dynamic target objects subject to different lighting changes.
    \item Detailed analysis of processing speed, efficiency, and comparison with related works for all developed algorithms, both offline and in the Kinova Gen3 robot.
\end{itemize}

This paper is organized as follows: Section II presents a brief survey of grasp detection, visual servoing, and dynamic grasping. Section III describes the methodology of the proposed system. Section IV presents the achieved results and discussion. Finally, Section V presents the conclusions.

\section{Brief Survey}

In this work, target object localization and object pose estimation are sub-tasks of the grasp detection stage and are processed through deep learning in an approach called end-to-end grasp detection \cite{du2019vision}. Object localization and object pose estimation are problems with their own literature and will not be addressed here, hence greater emphasis is placed on grasp detection strategies.

The grasp planning stage is performed in two steps. First, as a visual servo controller to reactively adapt to changes in the object's pose. Then, as an internal problem of the robot’s inverse kinematics, with no restrictions on the movement of the robot to the object, except for those related to singularities. 

Thus, a brief review of the works on visual servo control is presented, with greater emphasis on the ones that use learning to synthesize the controller. Finally, works that use visual sevoing specifically in the task of grasping, here called dynamic grasping, are presented.

\subsection{Grasp Detection}
\label{grasp_detection}

The early methods of grasp detection, known as analytic methods, rely on the geometric structure of the object to be grasped and have many problems with execution time and force estimation. Besides, they differ from data-driven methods in many ways. For a complete review of these methods, one may read the work of Bicchi and Kumar \cite{bicchi2000robotic}.

\begin{figure}[]
\centering
\includegraphics[width=\linewidth]{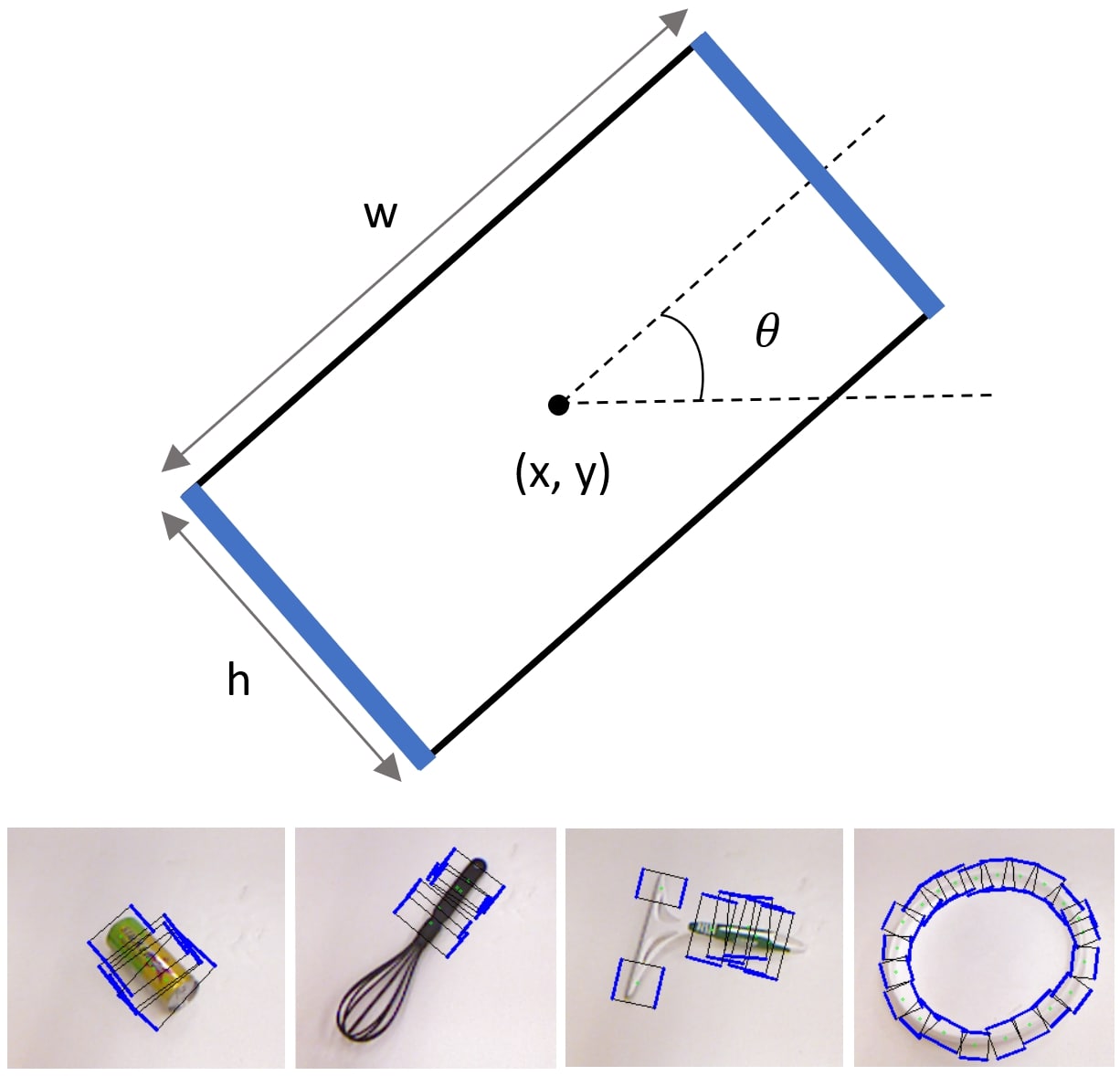}\\  
\caption{Representation of the robot’s grippers using a grasp rectangle and instances of the CGD with their respective grasp rectangles.} 
\label{fig:retpree}
\end{figure}

In the context of data-driven approaches, using only images, Jiang, Moseson and Saxena \cite{jiang2011efficient} proposed a representation of the position and orientation of a robot’s grippers, just before closing for grasping, from five dimensions. The scheme is illustrated in Fig. \ref{fig:retpree}. In the image, $(x, y)$ is the center of the oriented rectangle, $w$ represents the opening of the grippers and $h$ its size. In other words, the edges shown in blue represent the grippers of the robot. Finally, there is the angle $\theta$ that represents the grippers’ orientation. 
This five-dimensional representation is enough to encode the seven-dimensional representation of the grasp pose \cite{jiang2011efficient} since a normal approximation to the image plane is assumed, so that the 3D orientation is given only by $\theta$. The images required for the algorithm design were later updated by Lenz, Lee and Saxena \cite{lenz2015deep}, generating the Cornell Grasping Dataset. The current version of the dataset is composed of 885 images of domestic objects and their respective ground truth grasp rectangles, some of which are illustrated in Fig. \ref{fig:retpree}.

Lenz, Lee and Saxena \cite{lenz2015deep} were one of the first to test their algorithms with this dataset and to use deep learning for grasp detection. Although they did not obtain a relevant accuracy, it encouraged other researchers to investigate the grasp problem employing the rectangle representation.

Redmon and Angelova \cite{redmon2015real}, for example, also used the CGD to train an AlexNet architecture \cite{krizhevsky2012imagenet} that receives images with the blue channel replaced with depth information as input, and retrieves grasp rectangles. Furthermore, they introduced two different learning methods to accomplish grasp detection, which are called direct regression and multigrasp detection. The former is based on the presence of a single observed object, and the latter consists of estimating one rectangle for each fraction of the input image. Thus, using the multigrasp detection method, the AlexNet model is able to predict more rectangles associated with multiple objects. 

Kumra and Kanan \cite{kumra2017robotic} developed two ResNet \cite{he2016deep} based models that work in parallel. While one of them is trained with Red-Green-Blue (RGB) inputs, the other one receives depth data from the CGD. The accuracy of their results surpassed the ones obtained by Redmon and Angelova \cite{redmon2015real} and, although the joint networks have a higher number of parameters, the authors reported a prediction time of 103 ms.

Based on the Faster R-CNN \cite{ren2015faster}, Chu, Xu and Vela \cite{chu2018real} proposed a two-stage framework to address the grasp task. In the first stage, the grasp points are detected and, in the second stage, the orientation of the rectangle is obtained by classification methods. This network, from the object detection point of view, is a two-stage detector, whereas single-stage detectors perform a direct regression of the rectangle. Therefore, although the authors achieved good detection results in the CGD, the grasp detection runs at only 8 frames per second (fps).

Inspired by the same object detection network, Zhou et al. \cite{zhou2018fully} proposed a new approach that regards the variation of the rectangle angles, instead of its scale and size. According to the authors, the definition of the grasp rectangle heavily relies on its angles, which justifies their choice. They achieved state-of-art results in terms of grasp detection accuracy. In addition, the built network is single-stage, but the prediction speed is increased by only 0.2 fps compared to Chu, Xu and Vela \cite{chu2018real}.

Morrison, Corke and Leitner \cite{Morrison_2018} introduced a generative neural network (GG-CNN) that detects grasps, and an associated quality index, in all pixels of the input image, allowing a fully convolutional architecture to be used. The authors applied the CGD to transform the rectangular representation into parameterized images of grasp quality, angle and width, and use them for training. For testing, the network is evaluated in a real scenario with a prediction speed of 52.6 fps. However, the authors did not present the network efficiency in the CGD.

Following the same approach, Gu, Su and Bi \cite{gu2019attention} reparameterized the instances from the CGD into three images representing the ground truth grasp and used a fully convolutional architecture as well. They introduced an attention mechanism in the network, so that the model can focus on salient features, achieving state-of-the-art results in the dataset. However, the authors did not make it clear which set of the dataset was used to obtain this result. Furthermore, the developed network is slightly slower than the GG-CNN and no real tests are presented.

Our work focuses on the development of a simple and efficient CNN that predicts grasp rectangles. In the training and testing steps, the proposed network is light enough to allow the joint application of a second CNN that addresses the task of visual servo control. Thus, the entire system can be executed in real-time for robotic applications without decreasing the accuracy of both tasks.

\subsection{Visual Servo Control}

Classic VS strategies demand the extraction of visual features as an input to the control law. These features must be correctly selected since the robustness of the control is directly related to this selection. However, the use of projected features requires manual engineering for each new system and it avoids the exploitation of statistical regularities that make the developed VS system more flexible \cite{lee2017learning}. Assuming that notable convergence and stability issues can appear if the VS is performed in a region away from the projected operating point \cite{chaumette1998potential}, the classic VS techniques mainly deal with controller convergence and stability analysis. 

Relying on projected features, some studies sought to investigate the applicability of machine learning algorithms in VS \cite{miller1987sensor}. Other authors also sought to automate the interaction matrix estimation using a multilayer perceptron \cite{wei1999multisensory,ramachandram2003short} or support vector machines \cite{pandya2015servoing}.

The first step towards the independence of feature extraction and tracking was taken with the development of Direct Visual Servoing (DVS) by Deguchi \cite{Deguchi2000}. The technique has been improved to better deal with nonlinearities present in the minimized cost function and convergence domain \cite{caron2013photometric,silveira2012direct,silveira2014intensity}. However, although the control law is obtained directly from pixel intensity, information about the objects is still necessary and the complexity of the image processing algorithms tends to compromise their use.

In order to create a control system that uses neither projected features, nor metric information about objects, and is still robust to imaging conditions, Silveira and Malis \cite{silveira2012direct} developed a new DVS model. The authors explored the projective and geometric parameters that relate the current and desired image using pixel intensity. However, the system had initialization issues, since obtaining these parameters is computationally expensive. In a later work \cite{silveira2014intensity}, the authors presented three optimization methods to try to overcome computational complexity. However, cheaper methods did not have satisfactory convergence properties and generally need prior knowledge of the system. Furthermore, when evaluated in the real-world, the algorithms converge only with reduced gains and the cheaper method did not converge.

In this sense, the most recent VS techniques have explored deep learning algorithms to simultaneously overcome the problems of feature extraction and tracking, generalization, prior knowledge of the system, and processing time. The first work that demonstrated the possibility of generating a controller from raw-pixel images, without any prior knowledge of configuration, was developed by Zhang et al. \cite{zhang2015towards}. The authors used a Deep Q-Network to perform a target reaching task controlling 3 joints of a robot by means of deep visuomotor policies. The training was done in simulation and failed to be confronted with real images. However, when the camera images were replaced by synthetic images, the robot was controlled.

Other works that follow the Reinforcement Learning (RL) approach use deterministic policy gradients to design a new image based VS \cite{sampedro2018image} or Fuzzy Q-Learning, relying on feature extraction \cite{shi2016decoupled}, to control multirotor aerial robots. In a different approach, some works that explore deep learning in visual servoing rely on CNNs. Although RL appears to be the best approach to the problem, by mapping joint angles and camera images directly to the joint torques, CNNs can also be trained to perform end-to-end visual servoing. Moreover, the generalization power achieved by CNNs is superior to that of RL, since the parameters learned by RL are specific to the environment and task \cite{saxena2017exploring}.

Saxena et al. \cite{saxena2017exploring} developed a CNN trained to perform visual servoing in different environments, without knowledge of the scene’s geometry. To this end, the authors trained a network, based on the FlowNet architecture \cite{dosovitskiy2015flownet}, with the 7-Scenes dataset \cite{glocker2013real}, which has several images taken in sequence through transformations in the camera. Thus, having the current image $I_c$ and the desired image $I_d$, the network can predict the homogeneous transformation that relates the camera pose in $I_d$ and the camera pose in $I_c$. The authors performed tests on a quadcopter and achieved promising results in both indoor and outdoor environments.

Bateux et al. \cite{bateux2018training}, like the previous authors, developed a CNN capable of predicting the transformation that occurred in a camera through two images. Some essential differences are in the architecture used, in the robot operated and, mainly, in the dataset used. The networks are based on fine-tunning the AlexNet and VGG \cite{simonyan2014very} networks, and the operated robot is a 6 Degrees of Freedom (DoF) manipulator. The authors developed their own dataset using only one image. Starting from virtual cameras, it was possible to generate thousands of images and their associated transformations based on homography.

In our work, four models of convolutional neural networks were designed as potential candidates for end-to-end visual servo controllers. The networks do not use any type of additional information, other than a reference image and the current image, to regress the control signal. Thus, the proposed network works as a \textit{de facto} controller, predicting a velocity signal, not a relative pose.

\subsection{Dynamic Grasping}

Learning visual representations for perception-action \cite{piater2011learning} that adhere to reactive paradigm and that generates a control signal directly from sensory input, without high-level reasoning \cite{lampe2013acquiring}, can help in dynamic grasping. Therefore, some authors developed variants of VS specifically for reaching and grasping tasks, by using an uncalibrated camera \cite{piepmeier2004uncalibrated}, calculating the visual-motor Jacobian without knowledge of the system \cite{shademan2010robust}, or with policy iteration and imitation learning techniques \cite{ratliff2007imitation,stulp2011learning}

One of these works, developed by Lampe and Riedmiller \cite{lampe2013acquiring}, uses reinforcement learning to determine the outcome of a trajectory and to know if the robot will reach a final position in which closing the grippers results in a successful grasp. Moreover, the authors proposed a grasp success prediction that anticipates grasp success, based on current visual information, to avoid multiple attempts of grasping.

However, since they are applied to specific types of objects, and since they still rely on some prior knowledge, these systems do not have full capacity for generalization. Thus, more recently, a significant number of works have explored the use of deep learning as an approach to the problem of closed-loop grasping.

In this context, Levine et al. \cite{levine2016learning} proposed a grasp system based on two components. The first part is a CNN that receives an image and a motion command as inputs and outputs the probability that, by executing such a command, the resulting grasp will be satisfactory. The second component is the visual servoing function which uses the previous CNN to choose the command that will continuously control the robot towards a successful grasp. By separating the hand-eye coordination system between components, the authors could train the grasp CNN with standard supervised learning and design the control mechanism to use the network prediction, so that grasp performance is optimized. The resulting method can be interpreted as a form of deep reinforcement learning \cite{levine2016learning}.

Aiming to improve the work of Levine et al. \cite{levine2016learning}, Viereck et al. \cite{viereck2017learning} developed a system based on CNN to learn visuomotor skills using depth images in simulation. The learned controller is a function that calculates the distance-to-nearest-grasp, making it possible to react to changes in the position and orientation of objects. Starting from simulation, and mounting the depth sensor near the robot end-effector, the authors managed to adapt the algorithm to the real-world, avoiding the two months of training experience performed by Levine et al. \cite{levine2016learning}. However, the developed CNN calculates the distance in relation to a grasp pose given \textit{a priori}. This grasp pose is obtained using the algorithm developed by Pas and Platt \cite{ten2018using}, so that errors in the grasp detection can induce errors in the control.

Inspired by the work of Viereck et al. \cite{viereck2017learning} and Levine et al. \cite{levine2018learning}, whose designed network receives actions as inputs, Wang et al. \cite{wang2019homography} developed a Grasp Quality CNN that relates these actions to the images that the robot would see when executing them, using homography. Thus, from a dataset of only 30K grasp trials, collected in a self-supervised manner, the authors removed the need to generalize over different actions and predict the grasp success metrics in a large number of poses.

In a more similar way to the methodology developed in our work, Morrison, Corke and Leitner \cite{morrison2019learning} developed a system for closed-loop grasping in which the grasp detection and visual servoing are not simultaneously learned. The authors used a Fully CNN to obtain the grasp points and apply a position-based visual servoing to make the pose of the grippers match the predicted grasp pose.

\section{Methodology}
\label{metodologia}

To address the generalization challenge, a convolutional neural network is designed and trained for grasp detection on a large number of objects (240 objects considered in the CGD). To avoid one-shot grasp and react to changes in the workspace, a visual controller is designed. Furthermore, to also satisfy the generalization condition, the controller is learned from scratch using a CNN capable of end-to-end visual servoing. The real-time execution of the algorithms is guaranteed, since the developed networks are, by design, both light and fast.

The purpose of the VS is to guide the manipulator, through comparisons between the image continuously obtained by the camera and a reference image, to a position where the robot has a full view of the object, so that the grasp detection conditions are met. Thus, the application of the method encompasses all situations in which a robotic manipulator, with a camera mounted in eye-in-hand configuration, must track and grasp an object.

The system's development consists of three phases: design stage, test stage, and operational stage. The first is based on designing the CNN's architectures and training them, as well as gathering and processing the datasets. In the second phase, offline results are obtained using test sets and evaluated based on their accuracy, speed and application domain. The third phase involves the implementation of the trained networks on a robot to assess their adequacy in real-time and real-world applications.

\begin{figure}[!htb]
\centering
\includegraphics[width=\linewidth]{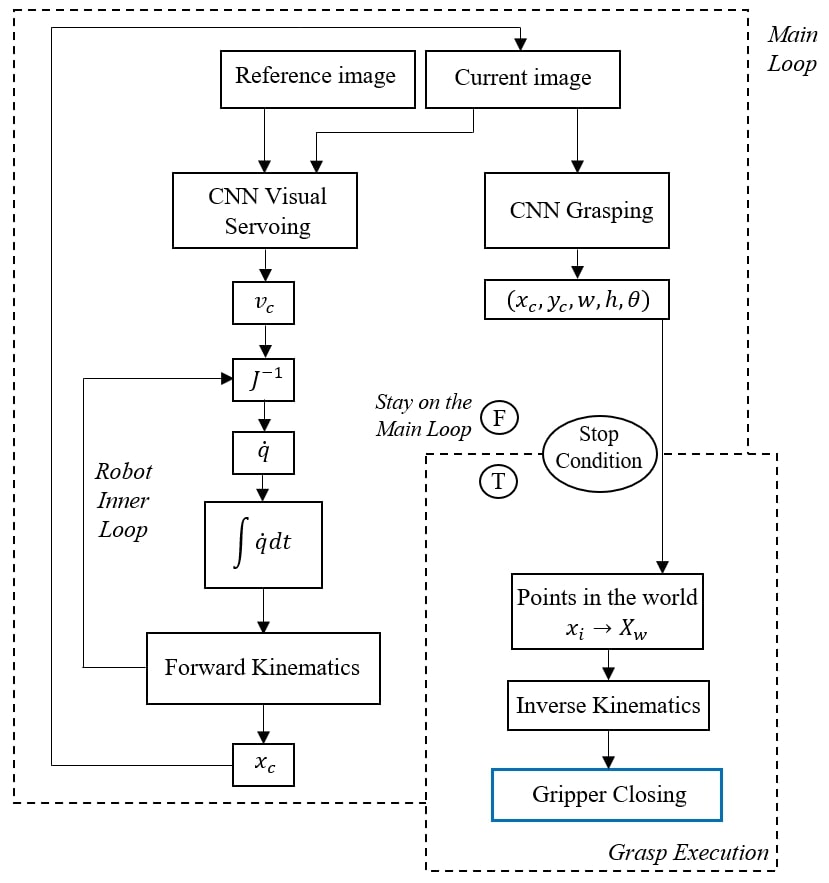}\\  
\caption{Schematic summary of the proposed dynamic grasp system} 
\label{fig:dynamic_grasp_system}
\end{figure}

The requirement for the system operation in the operational phase is to obtain an image of the target object \textit{a priori}, which will be used as setpoint by the VS. The control loop is executed as long as the L1-norm (Manhattan Distance) of the control signal is greater than a certain threshold. The operational phase is outlined in Fig. \ref{fig:dynamic_grasp_system}.

A single reference image is presented to the system as one of the Visual Servoing CNN’s inputs. The image obtained by the camera at present time serves as the second input in this network and as the input for the grasp CNN. Both networks run continuously, as the grasp CNN predicts the rectangle in real-time for monitoring purposes and the VS network executes real-time control of the robot’s pose.

The VS CNN predicts a velocity signal, which must be multiplied by a proportional gain $\lambda$, to be applied in the camera. The robot’s internal controller performs all necessary calculations to find the joints velocities that guarantee the predicted velocity in the camera. At each loop execution, the current image is updated according to the robot’s current position and, as long as the control signal is not small enough to guarantee convergence, this loop is repeated.

When the stop condition is satisfied, the prediction of the grasp network is mapped to world coordinates. Then, the robot executes the inverse kinematics to reach the predicted points and close the grippers. Since we deal with single-view grasping, we only have information about $x$ and $y$ coordinates of the grasping rectangle. Determining a grasp using multiple views of the object requires additional thought \cite{jiang2011efficient}, which is not within the scope of this work. Therefore, we approach the object to a predefined $z$ and impose restrictions on the height of the objects so that there is no collision. Note that the $z$ axis is not neglected in the experiment, it is considered throughout the process and can vary freely. We only define its value when the final grasping configuration has been obtained.

The methodology used in all phases of the proposed system is presented in the following sections. All network illustrations were designed with the \textit{PlotNeuralNet} open software \cite{haris_iqbal_2018_2526396}.

\subsection{Grasping Network Architecture}

\begin{figure*}[!htb]
\centering
\includegraphics[width=\textwidth]{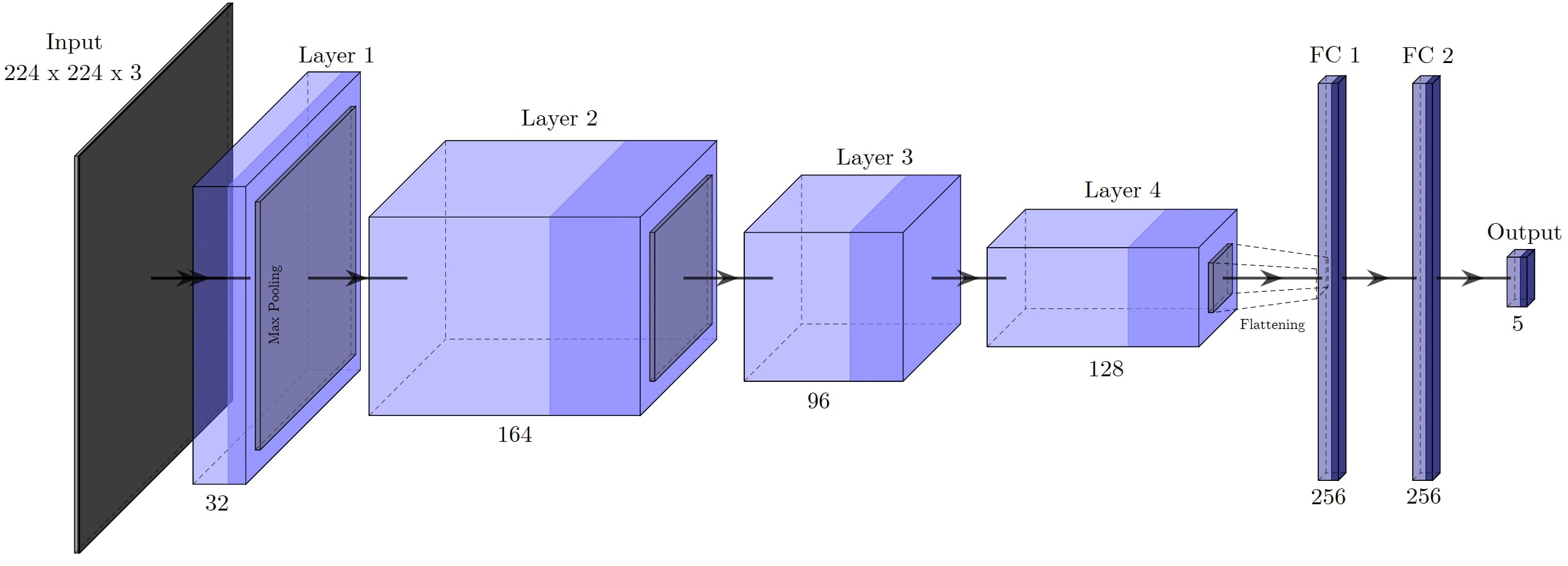}\\  
\caption{Convolutional neural network architecture for grasping} 
\label{fig:cnn}
\end{figure*}

The convolutional network architecture illustrated in Fig. \ref{fig:cnn} is proposed for grasp detection. The network receives as input a $224 \times224 \times3$ image, in which the third dimension refers to the number of channels, \textit{i.e.} RGB, without any depth information. The first layer is composed of $32$ convolution filters with dimensions $3\times3$ and the second layer contains $164$ of these filters. In both, the convolution operations are performed with stride $2$ and zero-padding, and are followed by batch normalization and $2\times2$ max pooling.

The third layer contains $96$ convolution filters in which the convolutions are performed with stride 1 and zero-padding, and followed only by batch normalization. The fourth, and last, convolutional layer is composed of $128$ filters performed with stride $1$ and followed by $2\times2$ max pooling. After the last layer of convolutions, the resulting feature map is flattened in an one-dimensional vector with 4,608 elements. This vector is further delivered to two Fully Connected (FC) layers with 256 neurons each. Between these layers, a dropout rate of 50\% is considered during training. 

Finally, the output layer consists of 5 neurons corresponding to the $(x,y,w,h,\theta)$ values that encode the grasp rectangle. In all layers, the activation function used is the Rectified Linear Unit (ReLU), except in the output layer, in which the linear function is used. The implemented network is noticeably smaller than others described in the literature for solving the grasping problem, with 1,548,569 trainable parameters.

\subsubsection{Cornell Grasping Dataset}

To proceed with the grasp network training, the CGD is used. It consists of a set of $885$ images and associated point clouds of $240$ common household objects, which would potentially be found by a personal robot. These objects have the appropriate size and shape to be grasped by a robotic arm equipped with parallel grippers capable of opening up to 4 inches. Such features make it possible to apply the trained agent beyond domestic environments, as seen in the related works. 

To obtain the data, a Kinect sensor mounted in the robot’s end-effector was used. During the data acquisition, the positions of the robot were chosen so that the manipulator can grasp the object from a path normal to the image plane. For encoding the dataset ground truths, the grasp rectangles were compiled with the \textit{x} and \textit{y} coordinates of the four vertices.

\begin{figure}[!htb]
\centering
\includegraphics[width=0.96\linewidth]{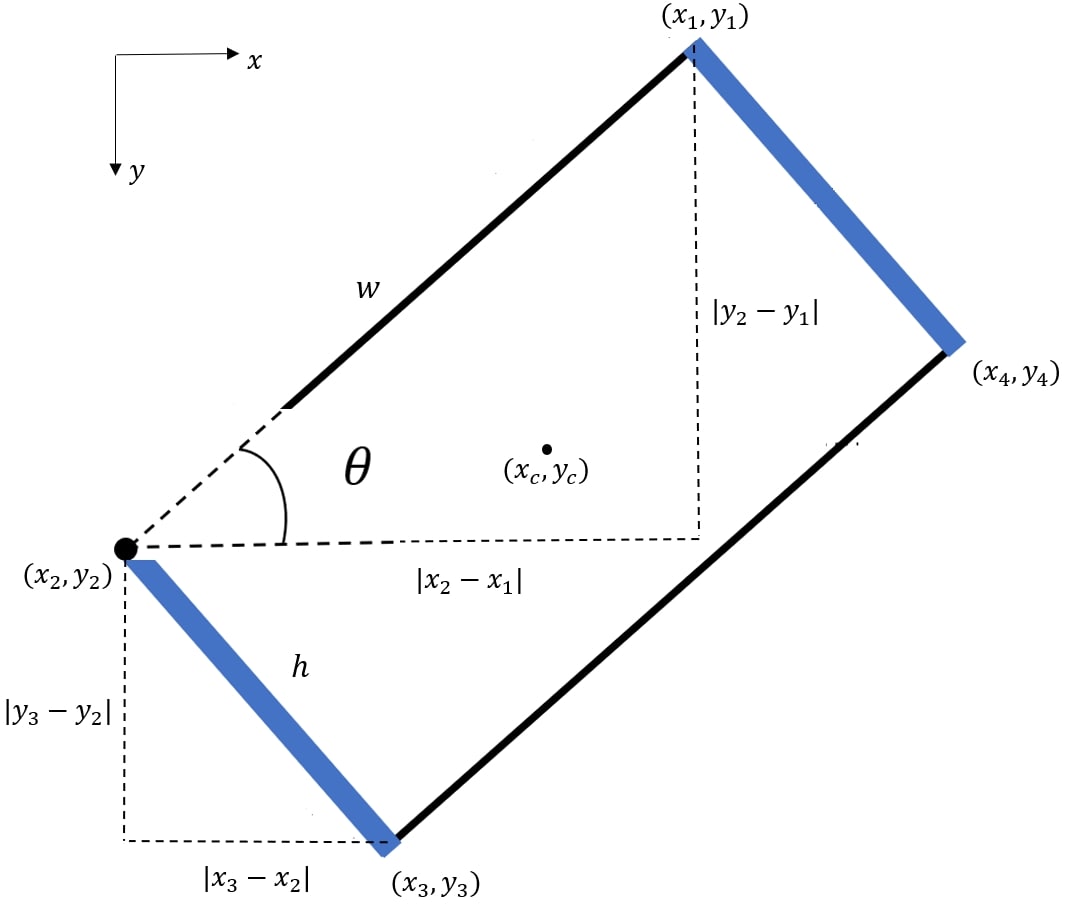}\\  
\caption{Rectangle encoding using its vertices coordinates}
\label{fig:retvar}
\end{figure}

Based on the vertices, five parameters are obtained for their representation, according to the methodology found in Jiang, Moseson and Saxena \cite{jiang2011efficient}. Fig. \ref{fig:retvar} and Equations \ref{g1}, \ref{g2} and \ref{g3} show how the dataset is reparameterized into $(x_c,y_c,w,h,\theta)$.

The $x_c$ and $y_c$ parameters, that represent the $x$ and $y$ coordinates of the rectangle's center point, respectively, are obtained from:

\begin{equation}
    x_c=\frac{\sum_{i=1}^4{x_i}}{4},
    \quad\quad\quad\quad
    y_c=\frac{\sum_{i=1}^4{y_i}}{4}.
    \label{g1}
\end{equation}

The grippers opening, $w$, and height, $h$, are calculated from the vertices as follows:

\begin{equation}
    w=\sqrt{(x_2-x_1)^2+(y_2-y_1)^2},
    \label{g2}
\end{equation}
\begin{equation*}
    h=\sqrt{(x_3-x_2)^2+(y_3-y_2)^2}.
\end{equation*}

Finally, $\theta$, which represents the orientation of the grippers relative to the horizontal axis, is given by:

\begin{equation}
    \theta=\atantwo(y_2-y_1,x_2-x_1).
    \label{g3}
\end{equation}

\subsubsection{Data Augmentation}

\begin{figure}[!htb]
\centering
\includegraphics[width=\linewidth]{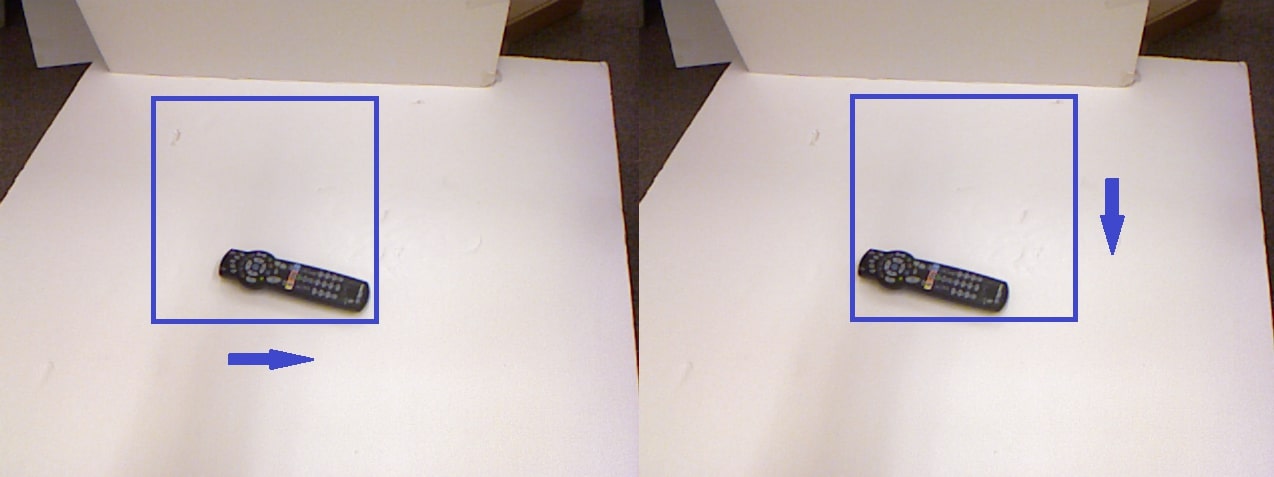}\\  
\caption{Cropping window used to generate new images} 
\label{fig:window}
\end{figure}

Data Augmentation (DA) \cite{tanner1987calculation} is a strategy for artificially increasing training data to improve the network’s learning ability. This technique is especially important if the relationship between the amount of available data and the number of trainable parameters in the network is too small. The applied method differs from the traditional translations and rotations usually applied in DA. Assuming that a much larger number of transformed images (335 times the original), each one with particular visual cues, may aid the network’s learning process, we use a $320 \times320$ cropping window that slides in the image, ensuring that the object of interest is always in the cropped area. Fig. \ref{fig:window} illustrates the strategy.

To crop the image without part of the object being outside the representation, we considered the values of the labeled grasp rectangles. Thus, the horizontal ($i$) and vertical ($j$) displacements of the window were limited by the minimum $x$ and $y$ coordinates of the associated grasp rectangles, respectively. Based on the position of the crop window, appropriate adaptations on the original ground truth rectangles was also made, to serve as label for the new dataset instance.

The applied DA resulted in a new dataset that is more than 335 times larger than the original: 297,338 training instances obtained from just 885 images.

\subsubsection{Training and Evaluation}
\label{grasptraintest}

The network is implemented in Python 3.6.8 through the Keras 2.2.4 framework (with Tensorflow 1.13.1 backend), in the Ubuntu 16.04 operating system. The hardware used is an Intel Core i7-7700 processor with 8 cores of 3.6GHz and an Nvidia Graphics Processing Unit (GPU) GeForce GTX 1050 Ti. We used Mean Squared Error (MSE) as loss function, Nesterov-accelerated Adaptive Moment Estimation (Nadam) \cite{dozat2016incorporating} with a learning rate of 0.002 as optimization algorithm, batch size of 144 and 30 epochs of training. All hyperparameters were empirically fine-tuned. Alexnet \cite{krizhevsky2012imagenet} was influential in the design of the lightweight network structure.

The dataset is divided into training (84\%), validation (1\%), and test (15\%) sets. The number of samples in each of them is presented in Table \ref{divisao}. Another division performed in the dataset comprises the type of images contained in the above mentioned sets. In the case called \textit{Image-Wise} (\textit{IW}), images are randomly organized into sets. In the \textit{Object-Wise} (\textit{OW}) case, all images containing the same object are placed in the same set.

\begin{table}[h]
\centering
\caption{Grasping dataset division}
\label{divisao}
\begin{tabular}{@{}c|ccc|c@{}}
\hline
\hline
   Set         & Training & Validation & Test & Total\\ 
\hline
\textit{Image-Wise}  & 250,211      & 2,527   & 44,600  & 297,338\\
\textit{Object-Wise} & 251,951      & 2,545   & 42,842  & 297,338\\ 
\hline
\hline
\end{tabular}
\end{table}

The evaluation of the result is made with the following metric, as used in related works of grasp detection by means of the rectangle representation:

\begin{itemize}[noitemsep, nosep]

    \item the difference between the angle $\theta$ of the predicted rectangle ($r_p$) and the correct rectangle ($r_c$) must be within $30^\circ$;
    
    \item the Jaccard index of the predicted rectangle relative to the correct one must be greater than 0.25. This index represents the percentage of overlap between the rectangles, according to the following formula:
    
    \begin{equation}
        J(r_p, r_c) = \frac{r_p \cap r_c}{r_p \cup r_c} > 0.25
    \end{equation}
    
\end{itemize}

If the network prediction satisfies these two conditions for one or more of the labeled rectangles associated with the object, then this prediction is considered to be correct. Note that the overlap between label and prediction (0.25) is less rigorous than in other computer vision problems (0.5) since the labeled rectangles represent only a fraction of all possible grasping rectangles associated with an object. Furthermore, Redmon and Angelova \cite{redmon2015real} claim that the limit of 0.25 is often sufficient to enable a successful grasp.

\subsection{Visual Servoing Network Architectures}

Unlike grasping, the network designed to perform visual servo control of a robotic manipulator receives two images as input and must regress six values, which can also be categorized into two outputs, considering linear and angular camera velocities. To this end, 3 network architectures are designed. Since two models were created from one of these architectures, a total of 4 models is presented to approach the VS task.

All networks, however, are generated from the remarkably simple structure of the grasping network. Therefore, the number of filters and the operations of each convolutional layer, the number of FC layers and the respective number of neurons, as well as the dropout rate, are similar to the grasping network and common to all developed models. All networks receive two images as inputs in the format $640 \times 360 \times 3$ and output the 6-dimensional velocity vector without any intermediate step. The size of the input images differs from the grasping network as the datasets used were collected in different formats: 4:3 for the CGD and 16:9 for the VS dataset. The inputs do not carry temporal information, only Cartesian relations. Thus, we do not approach any architecture with recurrent structures.

\begin{figure*}[t]
\centering
\includegraphics[width=\textwidth]{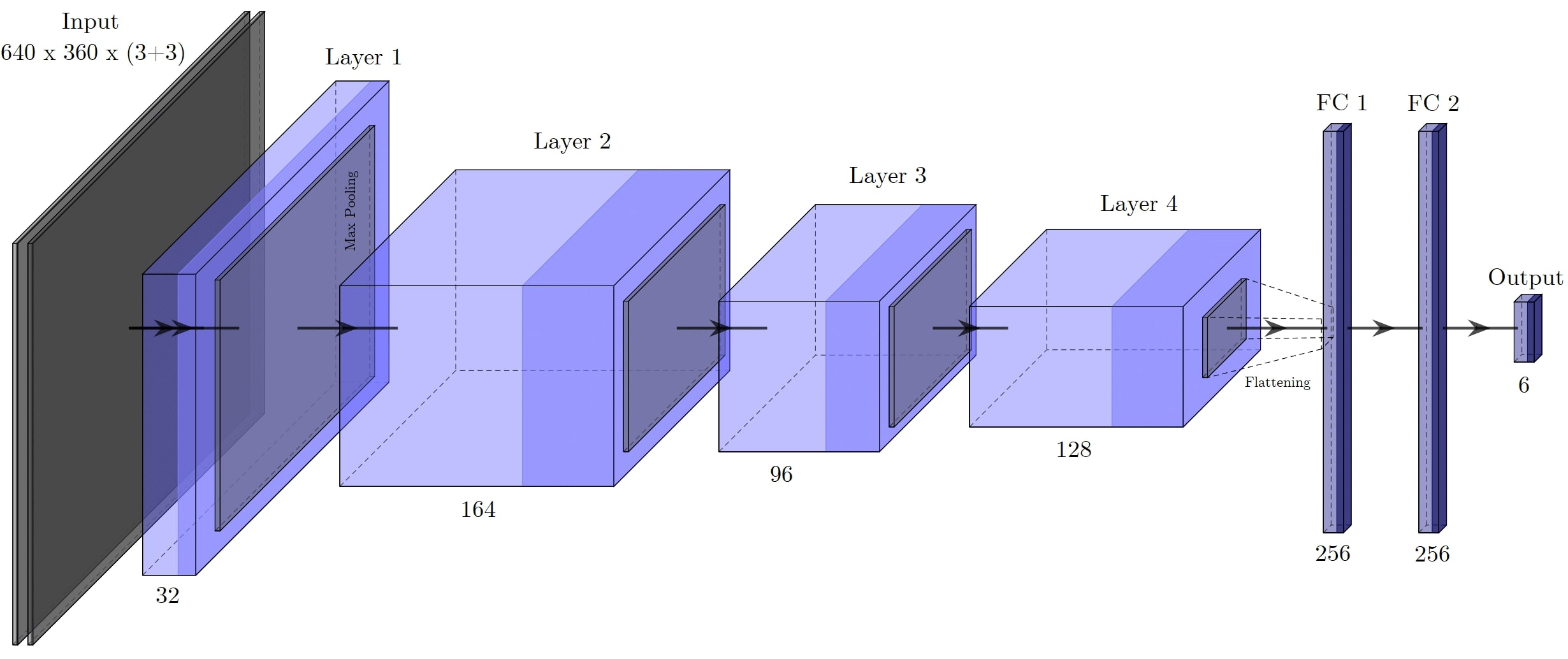}\\  
\caption{Model 1 - Direct regression} 
\label{fig:modvs1}
\end{figure*}

The first model for the VS task, called \textit{Model 1 - Direct Regression}, is presented in Fig. \ref{fig:modvs1}. It is similar to the grasping network, except for the inclusion of max-pooling in the third convolutional layer and for the different input dimensions. The inputs are simply concatenated and the data flow inside the network is always ahead, without any association or division of feature maps. Model 1 has only 1,549,690 trainable parameters.

\begin{figure*}[t]
\centering
\includegraphics[width=\textwidth]{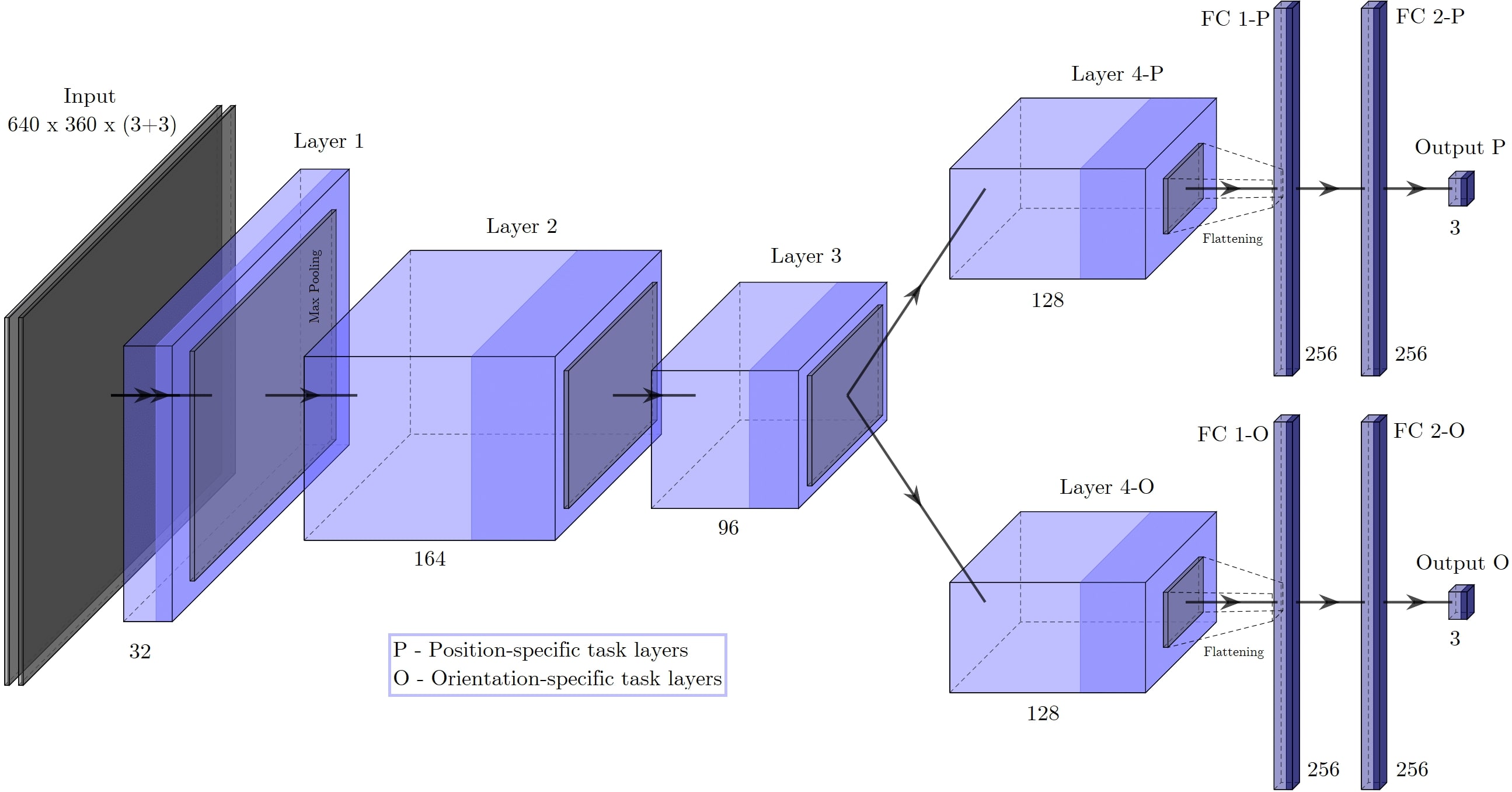}\\  
\caption{Model 2 - Task-specific regression} 
\label{fig:modvs2}
\end{figure*}

The second model, called \textit{Model 2 - Task-specific Regression}, is illustrated in Fig. \ref{fig:modvs2}. The network inputs are concatenated and the third set of feature maps is processed by two separate layers sequences. Thus, the 6D velocity vector is predicted by the network in the form of two 3D vectors. Specifically, this structure consists of a shared encoder and two specific decoders - one for linear velocity and another for angular velocity. Model 2 has 2,906,106 trainable parameters.

Therefore, the particularities of each domain can be learned by the network at the same time that its similarities can help in generalization. This structure is known as a multi-task network and, as presented by Kendall et al. \cite{kendall2018multi}, the cues from one task can be used to regularize and improve generalization of another one through inductive knowledge transfer. The purpose of this network is to evaluate how the dissociation of the velocity control problem, in two specific tasks of linear and angular velocities control, can affect the network accuracy.

\begin{figure*}[!htb]
\centering
\includegraphics[width=\textwidth]{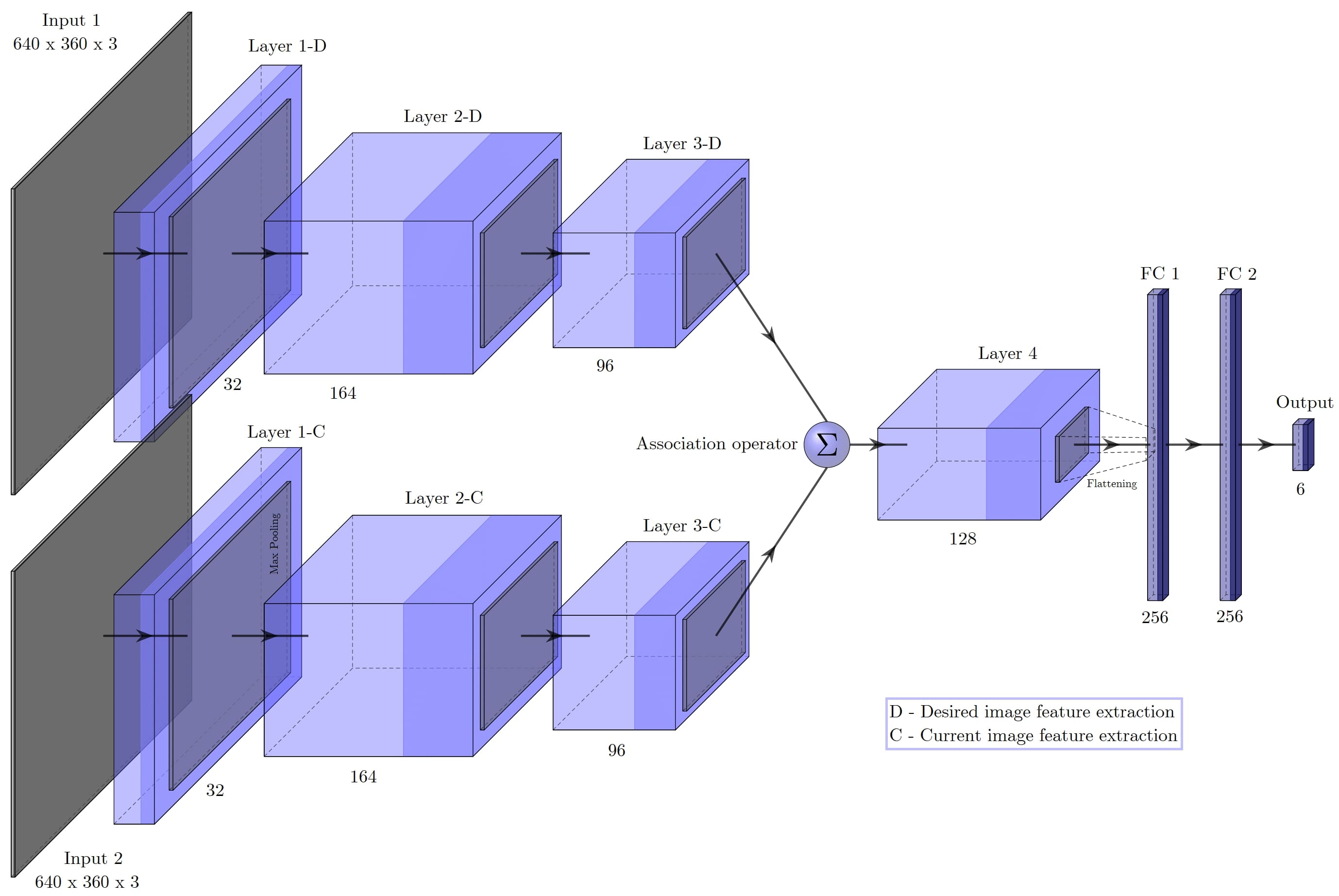}\\  
\caption{Models 3 and 4 - Direct regression from associated features} 
\label{fig:modvs3_4}
\end{figure*}

The third and fourth models, called \textit{Model 3 - Direct Regression from Concatenated Features} and \textit{Model 4 - Direct Regression from Correlated Features}, are created using the architecture presented in Fig. \ref{fig:modvs3_4}. This structure has two encoders and one decoder, which makes it possible to obtain high-level representations of each image before associating them. Two different association operators ($\Sigma$) were considered: a simple concatenation, which defines Model 3, and a correlation, proposed by Dosovitskiy et al. \cite{dosovitskiy2015flownet}, which defines Model 4.

Model 3 considers the pure concatenation of the feature maps resulting from the third convolutional layers, so that the input to the fourth layer is twice as deep. Model 4, on the other hand, has a correlation layer that aids the network to find correspondences between the feature representations of each image. The original correlation layer is a structural unit of the optical flow network FlowNet \cite{dosovitskiy2015flownet}. Models 3 and 4 have 1,850,286 and 1,533,358 trainable parameters respectively.

\subsubsection{Visual Servoing Dataset}

To train a model that can perform visual servoing on different target objects, without the need to design features, it is necessary to have a dataset that efficiently captures the attributes of the environment in which the robot operates, be representative of the VS task and diverse enough to ensure generalization. To this end, the data is collected by the Kinova Gen3 robot, in a way that approximates the self-supervised approach. Human interventions are associated with the assembly of the workspace and the setup of the robot, which involves determining a reference pose from which the robot captures the images and labels them.

The robot is programmed to assume different poses from a Gaussian distribution centered in the reference pose, with different Standard Deviations (SD). This approach is inspired by the work of Bateux et al. \cite{bateux2018training}, which does the same through virtual cameras and homography instead of a real environment. The reference pose (mean of the distribution) and the sets of SD, assumed by the robot, are presented in Table \ref{Gaussianparameters}.

\begin{table}[h]
\centering
\caption{Gaussian distribution parameters to build the VS dataset}
\label{Gaussianparameters}
\begin{tabular}{c|c|c|ccc}
\hline
\hline
\multicolumn{2}{c|}{\multirow{2}{*}{Dimension}} & \multirow{2}{*}{Mean} & \multicolumn{3}{c}{Standard Deviation} \\
\cline{4-6}
\multicolumn{2}{c|}{} & & High & Mid & Low \\
\hline
\multirow{3}{*}{\begin{tabular}[c]{@{}c@{}}Position\\(meters)\end{tabular}} & $x$ & 0.288 & 0.080 & 0.030 & 0.010     \\ 
& $y$          & 0.344  & 0.080 & 0.030 & 0.010                \\
& $z$        & 0.532   & 0.080 & 0.030 & 0.005               \\
\hline
\multirow{3}{*}{\begin{tabular}[c]{@{}c@{}}Orientation\\(degrees)\end{tabular}}&  $\alpha$    & 175.8 & 5.0 & 2.0 & 1.0 \\
& $\beta$     & -5.5   & 5.0 & 2.0 & 1.0                  \\
& $\gamma$   & 90.0    & 5.0 & 2.0 & 1.0                \\
\hline
\hline
\end{tabular}
\end{table}

\begin{figure}[!htb]
\centering
     \begin{subfigure}[b]{0.8\linewidth}
         \centering
         \includegraphics[width=\linewidth]{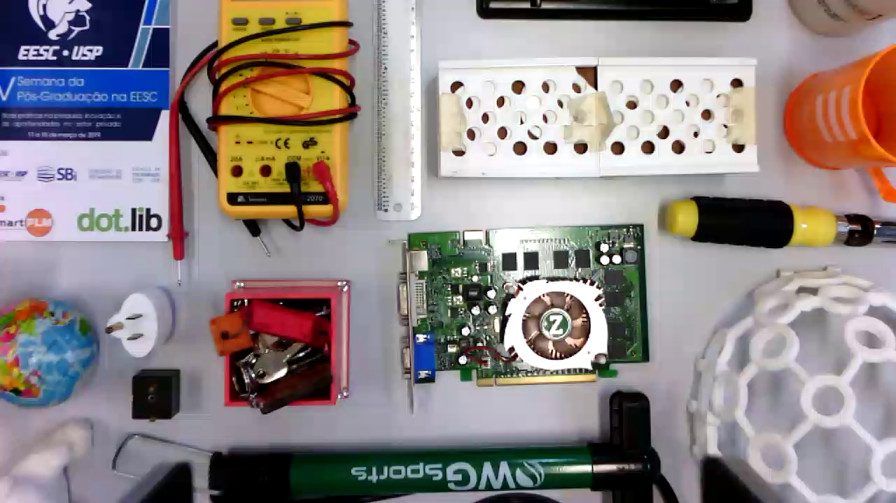}
         \caption{Image taken from camera in pose (0.326m, 0.356m, 0.503m, $178.0^\circ$, $1.1^\circ$, $91.5^\circ$)}
         \label{fig:23}
     \end{subfigure}
     \hfill
     \begin{subfigure}[b]{0.8\linewidth}
         \centering
         \includegraphics[width=\linewidth]{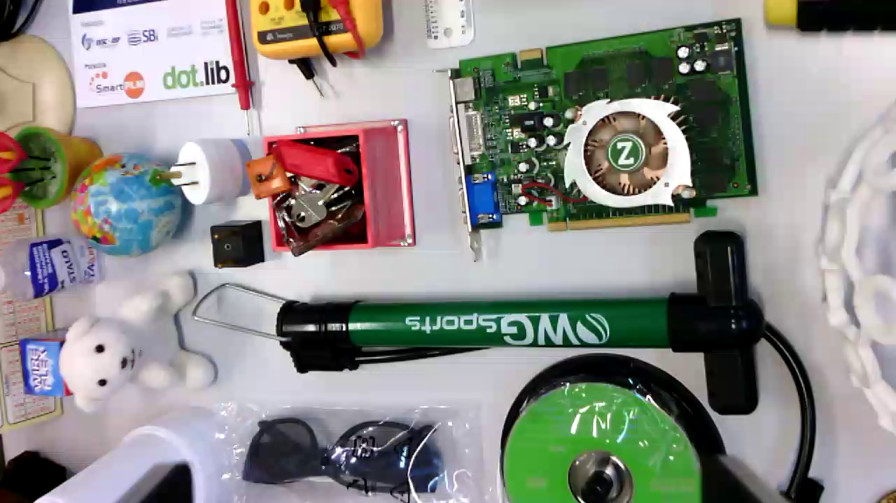}
         \caption{Image taken from camera in pose (0.258m, 0.207m, 0.402m, $-175.8^\circ$, $-23.0^\circ$, $87.2^\circ$)}
         \label{fig:26}
     \end{subfigure}
\caption{Instance examples from the VS dataset. Generated from a Gaussian distribution with mean in the reference pose $[x, y, z, \alpha, \beta, \gamma]$: (0.228m, 0.344m, 0.532m, $175.8^\circ$, $-5.5^\circ$, $90.0^\circ$)} 
\label{fig:datasetVS}
\end{figure}

The SD choices consider the expected displacement values that the robot must perform during the VS. Images obtained from a high SD help the network to understand the resulting changes in the image space when a large displacement is made by the robot. The instances obtained from a low SD enable the reduction of the error between the reference image and the current one when they are close, for a submillimeter precision in steady state. The mean SD values help the network to reason during most of the VS execution. Two dataset instance examples and their respective labels are illustrated in Fig. \ref{fig:datasetVS}.

Regarding the choice of objects and their distribution in the workspace to build the dataset, two factors were considered. On the one hand, the network must learn the features that make it possible to correlate a reference image to the current one, therefore, the greater the number of visual cues in the images, the better the network's learning ability. Thus, some scenes comprise a large number of objects (20 to 40), with varied structures, scattered across the table (as in Fig. \ref{fig:datasetVS}). On the other hand, some scenes are constructed so that the dataset is representative of the grasping task. Therefore, only one object is placed on the table, hence little visual information is available, but a good representation of the task is obtained.

In addition to single and multiple objects, we consider two other types of scenes - a framed map with a reflective surface and sets of books. Hence, the network may be robust to distortions in the captured image and may learn relevant features in flat objects. Brightness changes were considered during the dataset generation, since it was captured at different times of the day, without the concern of keeping luminosity constant. Shadows in random regions of the image were considered since the environment was not controlled and intermittent transit of people was allowed. The only post-processing performed on the obtained images is the exclusion of those that do not contain any part of the object. Since the VS dataset is large, we consider only 85\% of it for the experiment, 80\% are used for training and 5\% for testing. Table \ref{division_VS} shows the details.

\begin{table*}[!htb]
\centering
\caption{VS dataset: Composition, labels generation and division}
\label{division_VS}
\resizebox{0.8\textwidth}{!}{%
\begin{tabular}{@{}c|c|ccc|c|c|c@{}}
\hline
\hline
\multirow{2}{*}{Description} & \multicolumn{4}{c|}{Composition}& \multirow{2}{*}{Generation} & \multicolumn{2}{c}{Division}\\
\cline{2-5}
\cline{7-8}
 & Scenes       & High SD & Mid SD & Low SD & & Training & Test \\ 
\hline
Multiple objects & 4 & 767      & 630  & 600  & 361,315 & 289,052 & 18,065\\
 Single objects &  13   & 877   &910 & 780 & 161,625 & 129,300 &8,081 \\ 
 Books & 2 & 208   & 210 & 180  & 62,935& 50,348 &3,146\\ 
 Framed map & 1 & 140   & 140  & 120  & 50,470 & 40,376 &2,523\\ 
\hline
 \textbf{Total} & \textbf{20} & \textbf{1,992} & \textbf{1,890} & \textbf{1,680} & \textbf{636,345}& \textbf{509,076} & \textbf{31,817} \\
\hline
\hline
\end{tabular}
}
\end{table*}

After obtaining the data, the dataset is structured in the form $(I, [x, y, z, \alpha, \beta, \gamma])$, in which $I$ is the image, and $[x, y, z, \alpha, \beta, \gamma]$ is the associated camera pose when this image was captured, with $\alpha, \beta, \gamma$ represented in the Tait-Bryan extrinsic convention. To use this dataset to train a Position-Based VS neural network, according to the Visual Servo Control Tutorial by Chaumette and Hutchinson \cite{chaumette2006visual}, two images and the relative pose between them must be considered. Then, each instance of the processed dataset takes the form $(I_d, I_c,  \leftidx{^{d}}{\boldsymbol{H}_c})$, in which $I_d$ is a random instance chosen as the desired image, $I_c$ is another instance chosen as current image, and $\leftidx{^{d}}{\boldsymbol{H}_c}$ is the transformation that relates the current frame to the desired camera frame. This is done by expressing each pose, represented by translations and Euler angles, in an homogeneous transformation matrix form ($\leftidx{^{0}}{\boldsymbol{H}_d}$ and $\leftidx{^{0}}{\boldsymbol{H}_c}$), and then obtaining $\leftidx{^{d}}{\boldsymbol{H}_c} = \leftidx{^{0}}{\boldsymbol{H}_d}^{-1} \: \leftidx{^{0}}{\boldsymbol{H}_c}$. 

Finally, for the network to be, in fact, a controller, we need its prediction to be the velocity signal of the camera, \textit{i.e.}, the control signal. Hence, the data structured as $(I_d, I_c,  \leftidx{^{d}}{\boldsymbol{H}_c})$ is transformed to $(I_d, I_c, \boldsymbol{v_c})$, in which $\boldsymbol{v_c}$ is the proportional camera velocity. The proportional term is used since the gain $\lambda$ is not considered in determining the labeled velocity, and must be tuned \textit{a posteriori}, during the control execution. The velocity $\boldsymbol{v_c}$ is obtained from $\leftidx{^{d}}{\boldsymbol{H}_c}$ with Equations \ref{angle_axis} and \ref{finalVS},

\begin{equation}
    \theta = cos^{-1}\left(\frac{tr(\boldsymbol{R})-1}{2}\right),
    \quad
    \boldsymbol{u} =\frac{1}{2sen\theta}\begin{bmatrix} 
  r_{32} - r_{23} \\ 
  r_{13} - r_{31} \\ 
  r_{21} - r_{12} 
  \end{bmatrix},
  \label{angle_axis}
\end{equation}

\begin{equation}
 \boldsymbol{v}_c=  \begin{bmatrix} 
  -\lambda\boldsymbol{R}^T \; \leftidx{^{c^*}}{\boldsymbol{t}_c} \\ 
  -\lambda\theta\boldsymbol{u}
  \end{bmatrix},
  \label{finalVS}
\end{equation}

\noindent where $\boldsymbol{R}$ is the rotation matrix, $r_{ij}$ is the element at line $i$ and column $j$ of the same matrix,  $^{c^*}\boldsymbol{t}_c$ is the translation vector of the current camera frame related to the frame of the desired pose, $\lambda$ is the proportional gain (set to 1 in the dataset and later adjusted during the operational stage) and $\theta\boldsymbol{u}$ is the angle-axis representation of the rotation matrix.

\subsubsection{Training and Evaluation}

The networks are implemented using the same software and hardware setups as before. We used MSE as cost function, Adaptive Moment Estimation (Adam) \cite{kingma2014adam} with learning rate of 0.001 as optimization algorithm, batch size of 32 and 10 epochs of training. As in the grasping network, all hyperparameters were empirically fine-tuned.

For evaluation, the MSE is used as the performance metric, according to Eq. \ref{mseVS}, where $ n $ is the number of instances in the test set, $\boldsymbol{v_c^l}$ is the labeled camera velocity and $\boldsymbol{v_c^p}$ is the network prediction.

\begin{equation}
E_{MSE}=\frac{1}{n}\sum_{k=1}^n{(\boldsymbol{v_c^l}-\boldsymbol{v_c^p})^2}
\label{mseVS}
\end{equation}

For the purpose of illustration, predictions of all models for the same test instance are implemented as a velocity command in the robot. No feedback is considered, since we aim to show the effect of applying the velocity predicted by the network in open-loop, so that the MSE of this prediction is better understood in terms of the difference between poses.

\section{Results and Discussion}
\label{results}

In this section we present all the results obtained with the developed algorithms and evaluate them in two different ways: (a) considering the tasks of grasp detection and visual servoing in isolation and (b) the joint application in the dynamic grasping system.

\subsection{Offline Results}

This section presents the results achieved for grasping and visual servoing from the test sets developed for each task.

\subsubsection{Grasp Detection}

\begin{figure}[!htb]
\centering
\includegraphics[width=0.8\linewidth]{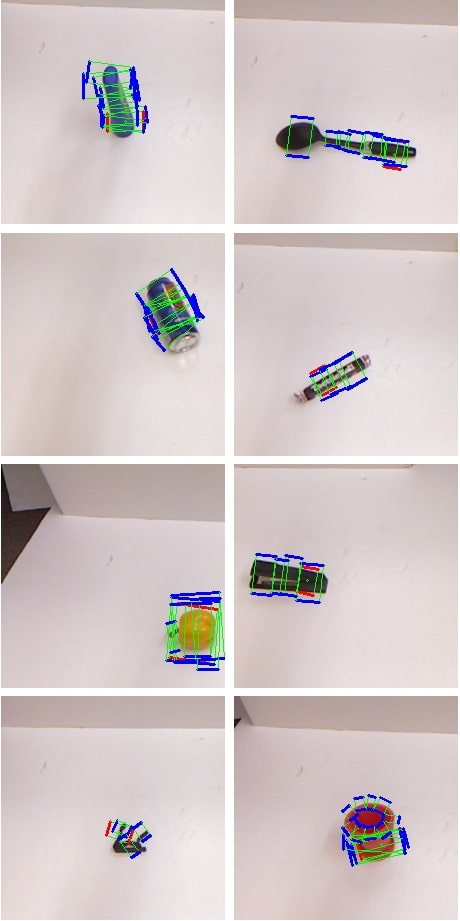}\\ 
\caption{Grasp network prediction (red) and labeled rectangles (blue)} 
\label{fig:resultimagens}
\end{figure}

The visual results of the developed grasp detection algorithm are illustrated in Fig. \ref{fig:resultimagens}. It is possible to note the network's ability to predict a grasp rectangle consistent with the object's shape even if it requires a small gripper opening or a choice from several options of grasp points.

Since several labeled rectangles are associated with each dataset instance, the way training is conducted must be chosen between using all of them as labels or using only one, chosen at random. Although the first option allows a more reliable characterization of the grasping process, it is not suitable in the case where the network predicts only one rectangle, as it learns to predict the average of the labeled rectangles. Thus, in the case of a circular object, for example, the network processes all labels arranged on its circumference and returns a central rectangle, which will certainly not result in a good grasp. Therefore, we carry out all training steps with only one label, that is, once we choose a label at random for each instance, we keep it for the entire training process.

\begin{table*}[h]
\centering
\caption{Accuracy, speed and implementation details of state-of-the-art approaches on the Cornell Grasping Dataset. *(Approximate number of parameters based on the models presented by the authors)}
\label{tab:res}
\resizebox{\linewidth}{!}{
\begin{tabular}{@{}l|c|c|c|c|c|c@{}}
\hline
\hline
\textit{Approach} & \textit{Image-Wise} & \textit{Object-Wise} & \textit{Hardware} & \textit{Input} & \textit{Parameters} & \textit{Speed} $(fps)$\\ 
\hline
Jiang, Moseson and Saxena (2011) \cite{jiang2011efficient} & 60.5\% & 58.3\% & - & RGB-D & - & 0.02 \\
Lenz, Lee and Saxena (2015) \cite{lenz2015deep} & 73.9\% & 75.6\% & - & RGB-D & - & 0.07 \\
Redmon and Angelova (2015) (1) \cite{redmon2015real} & 84.4\% & 84.9\%  & Tesla K20 & RGB-D & 7.3M* & 13.15\\
Redmon and Angelova (2015) (2) \cite{redmon2015real} & 88.0\% & 87.1\% & Tesla K20 & RGB-D & 7.3M* & 13.15\\
Wang et al. (2016) \cite{Wang2016robot} & 81.8\% & - & GeForce GTX 980 & RGB-D & - & 7.10 \\
Asif, Bennamoun and Sohel (2017) \cite{Asif2017rgbd} & 88.2\% & 87.5\% & - & RGB-D & 144M* & - \\
Kumra and Kanan (2017) \cite{kumra2017robotic} & 89.2\% & 88.9\% & GeForce GTX 645 & RGB-D & 50M* & 16.03\\
Oliveira, Alves and Malqui (2017) \cite{cbic-paper-97} & 86.5\% & 84.9\% & Tesla C2070 & RGB & - & - \\
Guo et al. (2017) \cite{guo2017robotic} & 93.2\% & 89.1\% & - & RGB, tactile & - & - \\
Asif, Tang and Harrer (2018) \cite{asif2018graspnet} & 90.2\% & 90.6\% & Jetson TX1 & RGB-D & 3.71M & 41.67\\
Park and Chun (2018) \cite{park2018classification} & 89.6\% & - & GeForce GTX 1080 Ti & RGB-D & - & 43.48\\
Chu, Xu and Vela (2018) \cite{chu2018real} & 96.0\% & 96.1\% & Titan-X & RGB-D & 25M* & 8.33 \\
Zhou et al. (2018) \cite{zhou2018fully} & \textbf{97.7\%} & \textbf{96.6\%} & Titan-X & RGB & 44M* & 8.51 \\ 
Zhang et al. (2018) \cite{zhang2018roi} & 93.6\% & 93.5\% & GeForce GTX 1080 Ti & RGB & 44M* & 25.16\\
Ghazaei et al. (2018) \cite{ghazaei2018dealing} & 91.5\% & 90.1\% & Titan-Xp & RGB & - & 17.86\\
Chen, Huang and Meng (2019) \cite{chen2019convolutional} & 86.4\% & 84.7\% & - & RGB-D & - & - \\
Proposed & 94.8\%  & 86.9\%  & GeForce GTX 1050 Ti & RGB & \textbf{1.55M} & \textbf{74.07}\\ 
\hline
\hline
\end{tabular}
}
\end{table*}

Table \ref{tab:res} presents the speed and accuracy obtained by our approach and compares it with related works. Hardware details, input information and number of parameters are also highlighted so that our results are better understood. Our accuracy of $ 94.8 \% $ in the image-wise division is in the top three, and the $ 86.9 \% $ accuracy for the object-wise division is comparable to the listed publications, except for the works of Chu, Xu and Vela \cite {chu2018real} and Zhou et al. \cite {zhou2018fully}, which are far behind in speed.

Considering the extremely lightweight architecture of the projected network (4 convolutional layers and 2 FC layers), we can state that it is quite efficient. With a much lower computational cost, expressed in high prediction speed, our proposed network can achieve similar or better results than state-of-art methods that employ more complex architectures. Moreover, 12 out of the 17 listed approaches use depth or tactile information in addition to RGB images, making their application inflexible to scenarios where such information cannot be obtained.

Our network's prediction time surpasses the one reported in the work of Park and Chun \cite{park2018classification} by approximately 10ms, updating the state-of-the-art prediction time on the CGD from 23ms to 13.5ms. Considering that the faster approaches in Table \ref{tab:res} use more powerful GPUs than ours (\cite{park2018classification, zhang2018roi}), the prediction speed comes primarily from the simplicity of the developed architecture. One exception is the work of Asif, Tang and Harrer \cite{asif2018graspnet}, which employs a less powerful GPU for an embedded application and achieves high speed. However, our method is lighter and does not require depth information. Moreover, the small penalty on accuracy compared with the simplicity of our architecture is achieved with the robust data augmentation strategy employed.

This high prediction speed allows us to use the network in real-time, real-world robotic systems in order to make them able to react to changes in the environment and dynamic target objects. We demonstrate the execution speed in the next sections, where we implement the trained network on a less computational power device for real-time control of a robot.

For a final comparison in terms of accuracy and prediction speed using the same hardware, we implemented a ResNet-50 network \cite {he2016deep}, which is employed by Kumra and Kanan \cite{kumra2017robotic} and Chu, Xu and Vela \cite{chu2018real}. With the ResNet as encoder, our framework achieved accuracy of $ 95.44 \% $ in the image-wise division, but was unable to generalize learning to new objects in the object-wise division, reaching only $46.92\%$ of accuracy. ResNet's large number of trainable parameters (25 million) specialized on the training data and lost the ability to generalize, resulting in overfitting.

Regarding the training time and prediction speed, the framework using ResNet took twice as long as our network to be trained and presented a prediction speed about three times lower than ours (25.64fps). Therefore, we can imply that, even with the same hardware, the approaches proposed in \cite{kumra2017robotic} and \cite{chu2018real} do not surpass ours in speed.

\subsubsection{Visual Servoing}

Prediction speeds and accuracy for the developed networks are presented in Table \ref{VStimes}. The speeds shown in the table were obtained from the average of the prediction speed of each model for the test data. The MSE is the evaluation metric. 

\begin{table}[h]
\centering
\caption{Prediction speed and accuracy of the VS CNN models}
\label{VStimes}
\begin{tabular}{@{}ccccc@{}}
\hline
\hline
& Model 1 & Model 2 & Model 3 & Model 4 \\
\hline
Speed & \textbf{86.77 fps}  & 82.49 fps & 70.12 fps & 59.86 fps    \\
MSE & 1.892 & \textbf{1.479} & 1.717 & 2.585 \\
\hline
\hline
\end{tabular}
\end{table}

Model 2 is the one that leads to the best offline results in the visual servoing task, demonstrating that the dissociation between linear and angular velocities and multi-task learning can boost performance. Model 4, on the other hand, had the worst performance, which suggests that a correlation layer designed for the optical flow task does not produce significant features for the visual servoing task.

By offline results, we mean only the difference between the predicted and labeled velocities. To get a better idea of how the MSE expresses the quality of the designed networks, the prediction of each model for an instance of the test set is presented in Table \ref{exampletable}. Thus, it is possible to make a quantitative comparison between the predicted and the labeled velocity values.
            
\begin{table}[h]
\centering
\caption{Labeled and predicted velocities using two input images from the test set, in which the current and desired ones are obtained at poses [0.344m, 0.326m, 0.372m, $-179.089^{\circ}$, $-3.201^{\circ}$, $91.907^{\circ}$] and [0.276m, 0.390m, 0.411m, $177.734^{\circ}$, $-4.326^{\circ}$, $90.252^{\circ}$], respectively. Values should be multiplied by $10^{-2}$}.
\label{exampletable}
\resizebox{\linewidth}{!}{
\begin{tabular}{@{}c|c|c|c|c|c|c@{}}
\hline
\hline
 Prediction $(\times10^{-2})$ & $\dot{x}$ [m/s] & $\dot{y}$ [m/s] & $\dot{z}$ [m/s] & $\dot{\alpha}$ [$^{\circ}$/s] & $\dot{\beta}$ [$^{\circ}$/s] & $\dot{\gamma}$ [$^{\circ}$/s] \\
 \hline
 Label   & -6.756 & 6.617 & 3.415 & 5.731 & -1.928 & -2.904 \\
 Model 1 & -6.533 & 5.085 & 1.388 & 4.949 & -2.232 & -3.141 \\
 Model 2 & -5.687 & 5.536 & 2.175 & 5.489 & -2.420 & -3.593 \\
 Model 3 & -5.355 & 7.007 & 2.625 & 6.492 & -2.363 & -1.751 \\
 Model 4 & -6.104 & 5.798 & 4.011 & 6.032 & -1.450 & -3.802 \\
\hline
\hline
\end{tabular}
}
\end{table}

Going further, we use the predictions to illustrate the robot's behavior when these speeds are applied to the camera. First, the robot is positioned in a pose equivalent to the one it had when it captured the current image ($I_c$) of the test instance. Then the speed command is applied (without feedback), considering $ \lambda_{lin} = 0.1 $ and $ \lambda_ {ang} = 10 $, until the L1-norms of the linear and angular velocities are greater than the L1-norms of the previous iteration. When this condition is met, the applied speed is set to zero and the robot stops moving. This is not the actual visual servoing, which is a closed-loop control, but rather an illustration of how the velocity predicted by the network can affect the robot's pose.

\begin{figure*}[!htb]
\centering
     \begin{subfigure}[b]{0.35\textwidth}
         \centering
         \includegraphics[width=\textwidth]{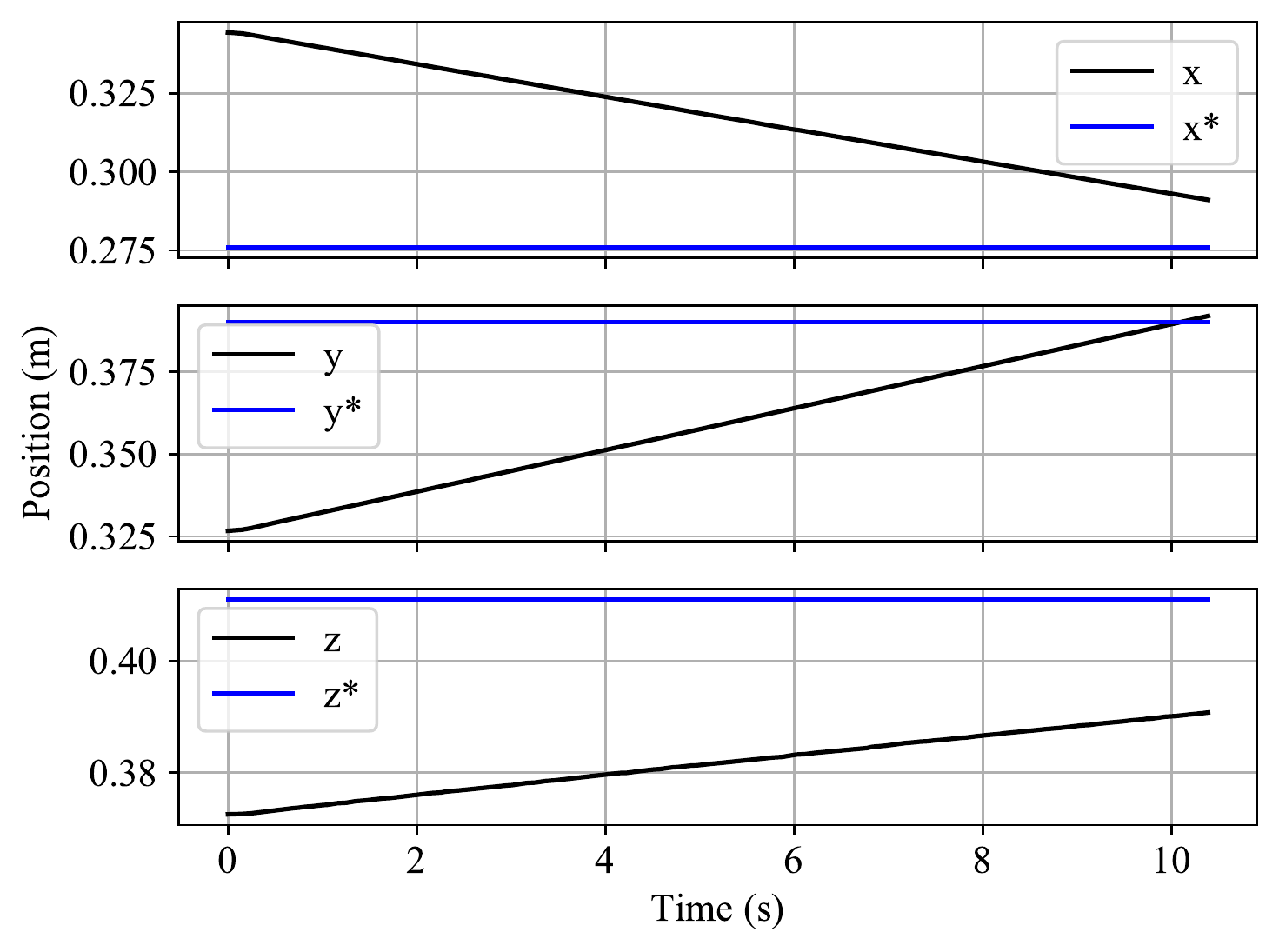}
         \caption{Model 1 prediction}
         \label{fig:posol2}
     \end{subfigure}
     \begin{subfigure}[b]{0.35\textwidth}
         \centering
         \includegraphics[width=\textwidth]{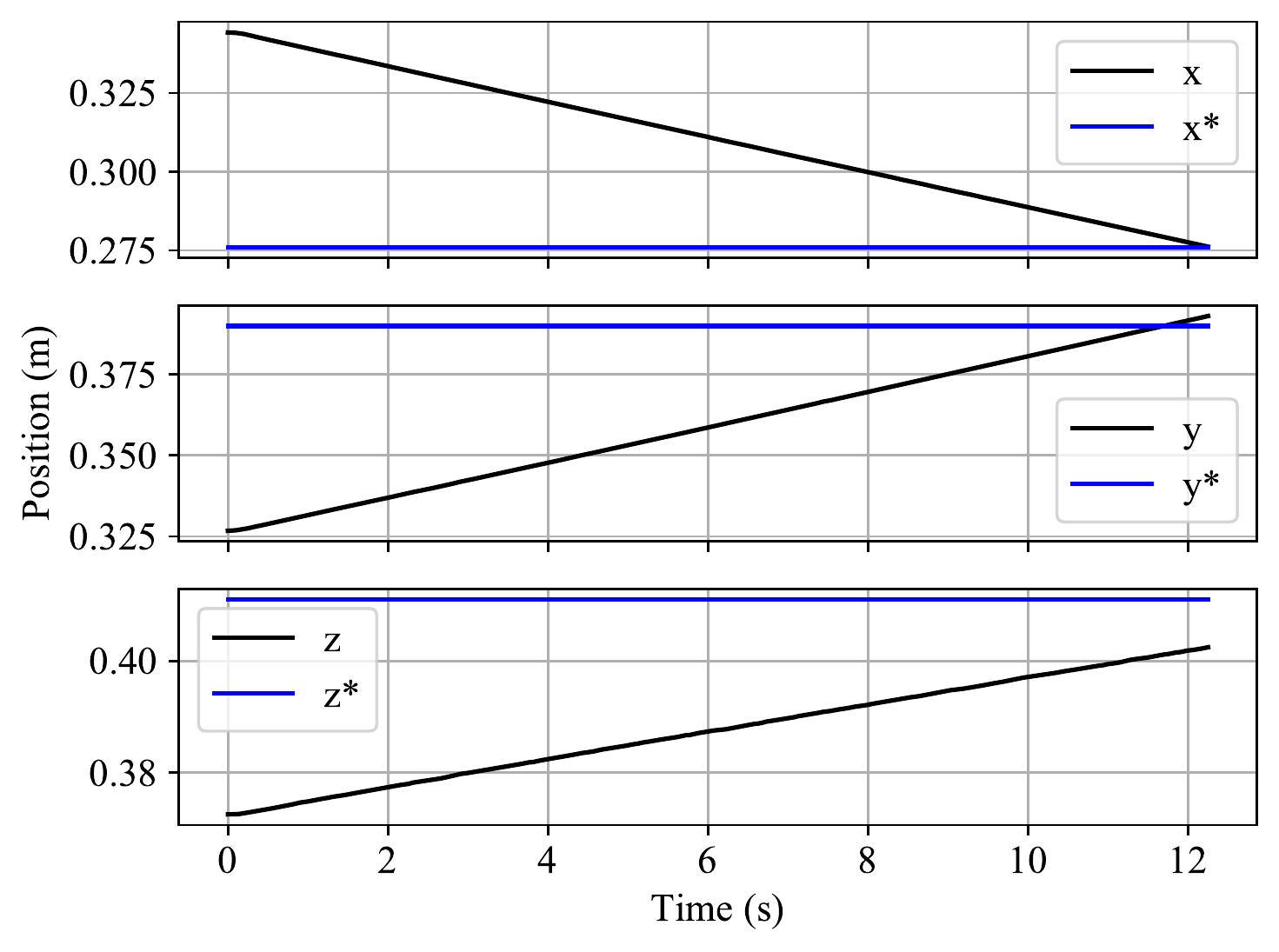}
         \caption{Model 2 prediction}
         \label{fig:posol3}
     \end{subfigure}
     \begin{subfigure}[b]{0.35\textwidth}
         \centering
         \includegraphics[width=\textwidth]{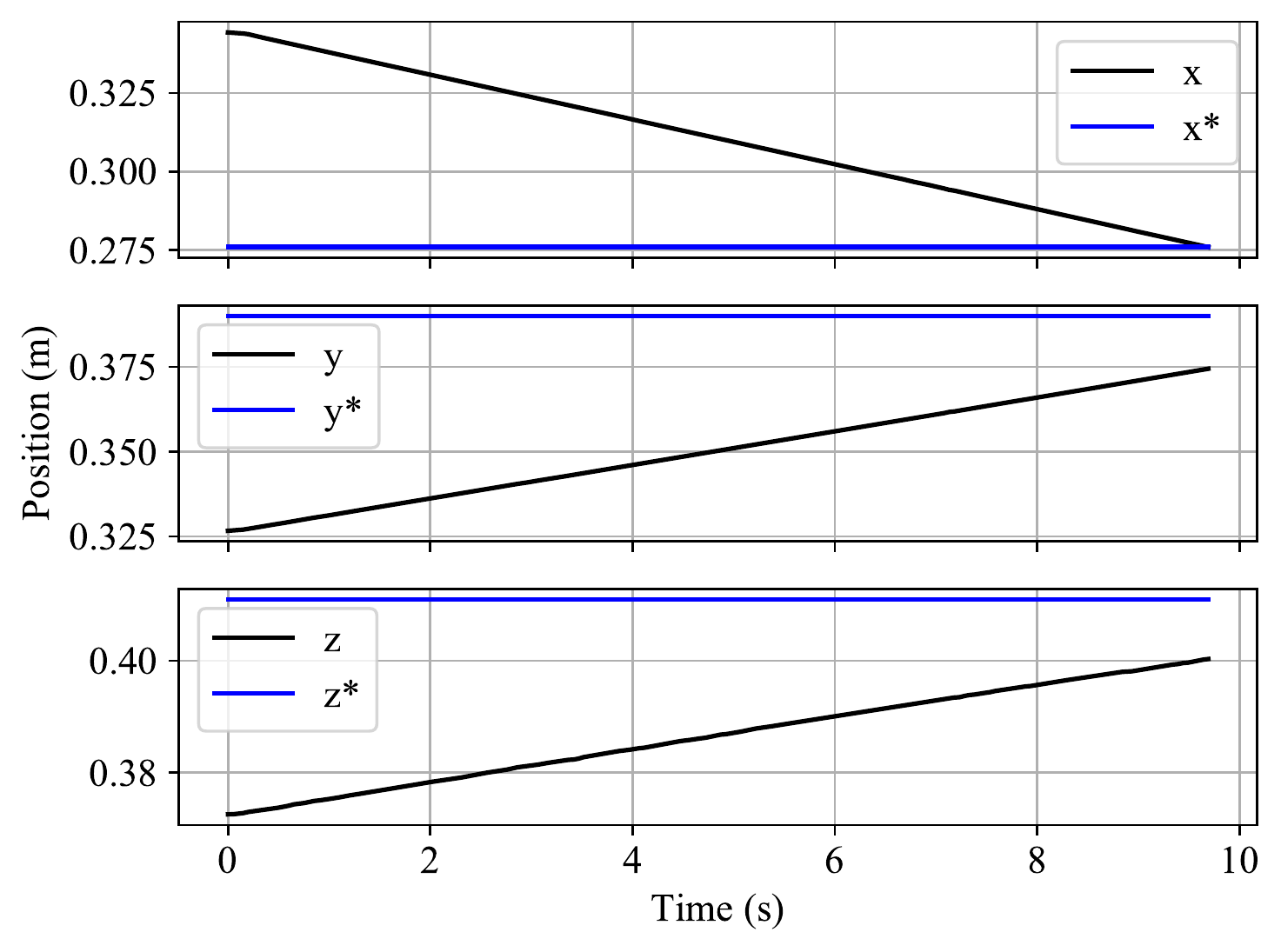}
         \caption{Model 3 prediction}
         \label{fig:posol4}
     \end{subfigure}
     \begin{subfigure}[b]{0.35\textwidth}
         \centering
         \includegraphics[width=\textwidth]{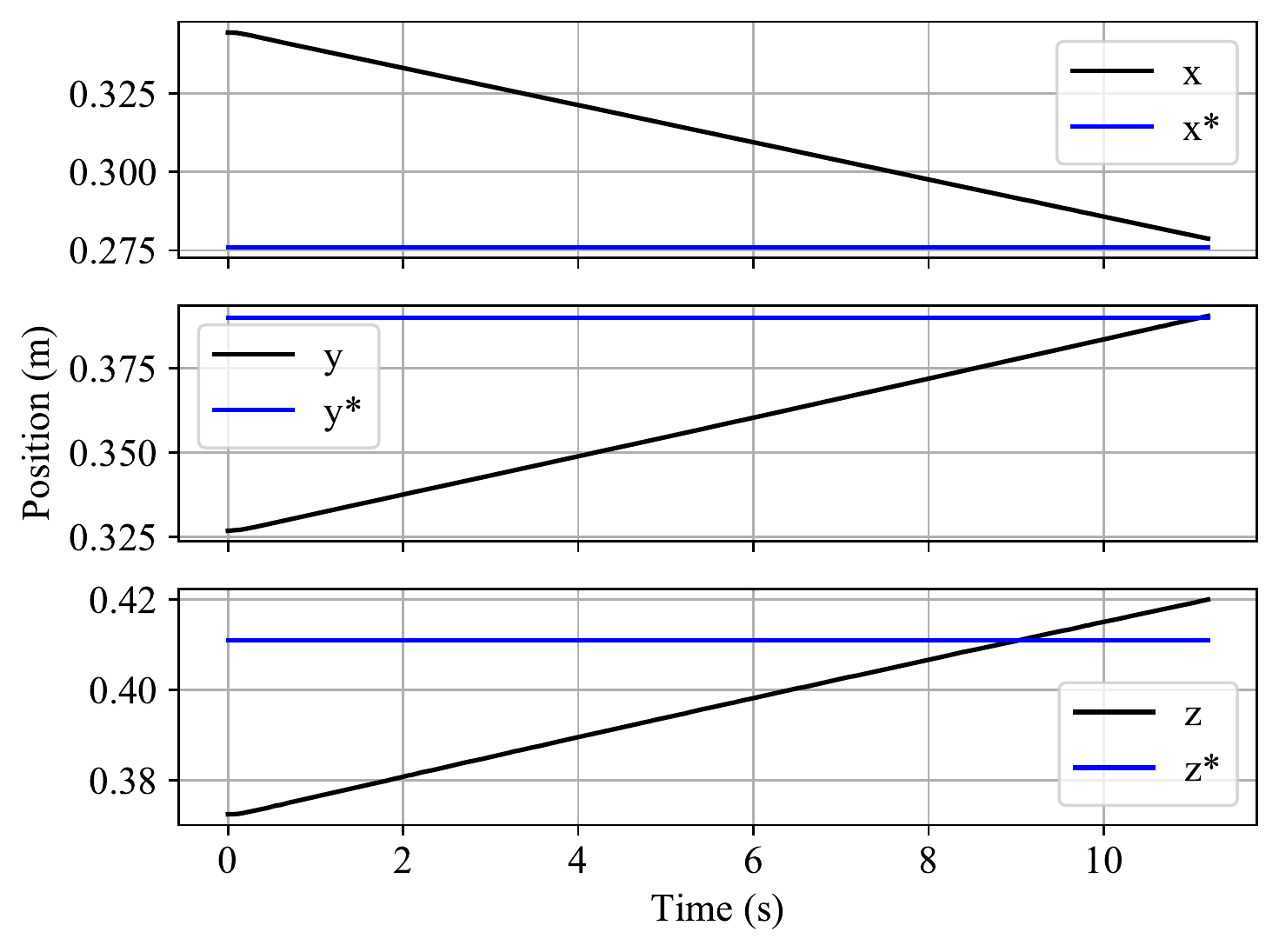}
         \caption{Model 4 prediction}
         \label{fig:posol5}
     \end{subfigure}
     \caption{Evolution of the robot's position in time (black) and its desired value (blue), when the network prediction (weighted by $\lambda_{lin}=0.1$ and $\lambda_{ang}=10$) for an instance of the test set is sent to the robot, which executes it without feedback.} 
\label{fig:posol}
\end{figure*}

\begin{figure*}[!htb]
\centering
     \begin{subfigure}[b]{0.35\textwidth}
         \centering
         \includegraphics[width=\textwidth]{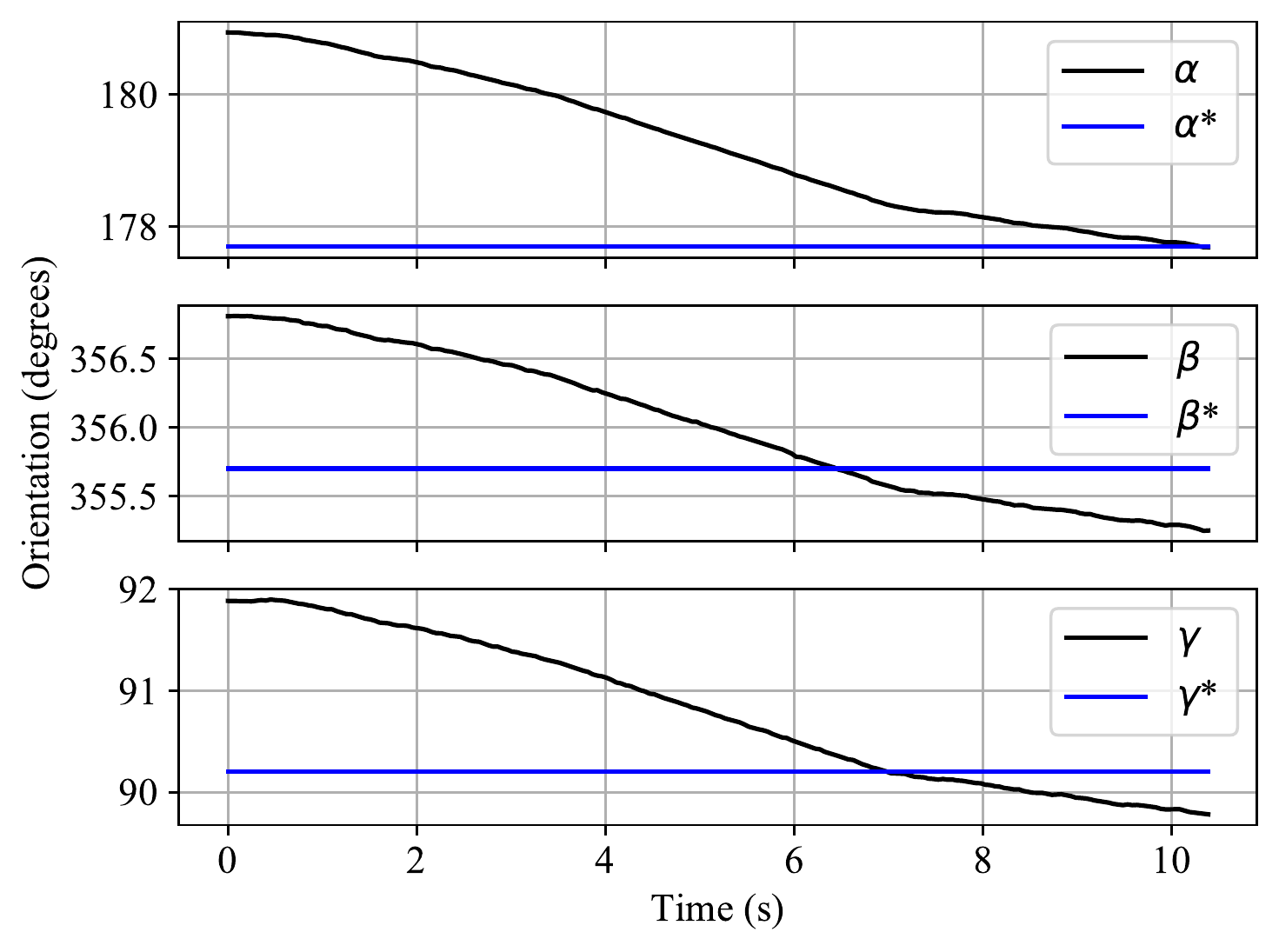}
         \caption{Model 1 prediction}
         \label{fig:oriol2}
     \end{subfigure}
     \begin{subfigure}[b]{0.35\textwidth}
         \centering
         \includegraphics[width=\textwidth]{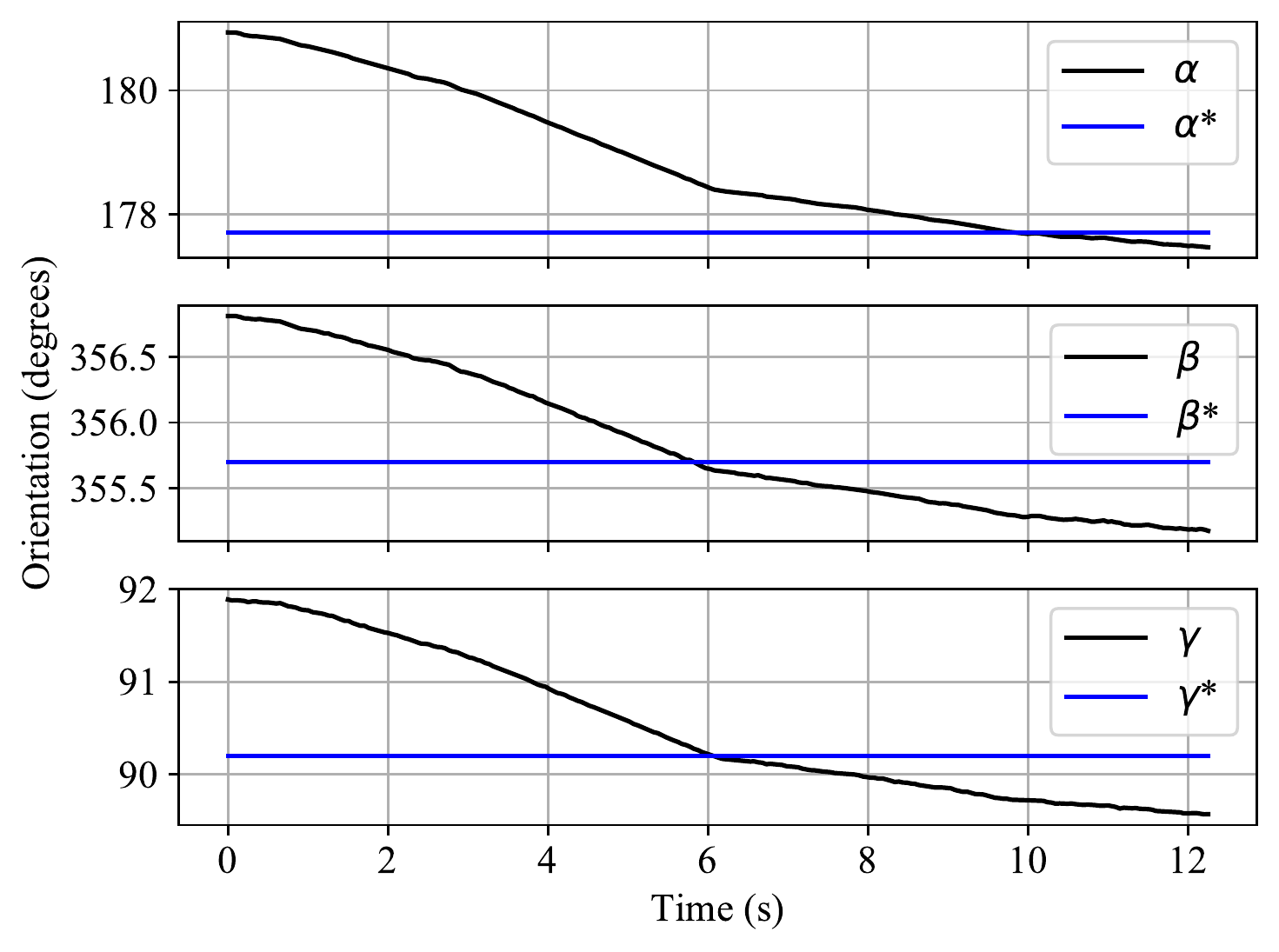}
         \caption{Model 2 prediction}
         \label{fig:oriol3}
     \end{subfigure}
     \begin{subfigure}[b]{0.35\textwidth}
         \centering
         \includegraphics[width=\textwidth]{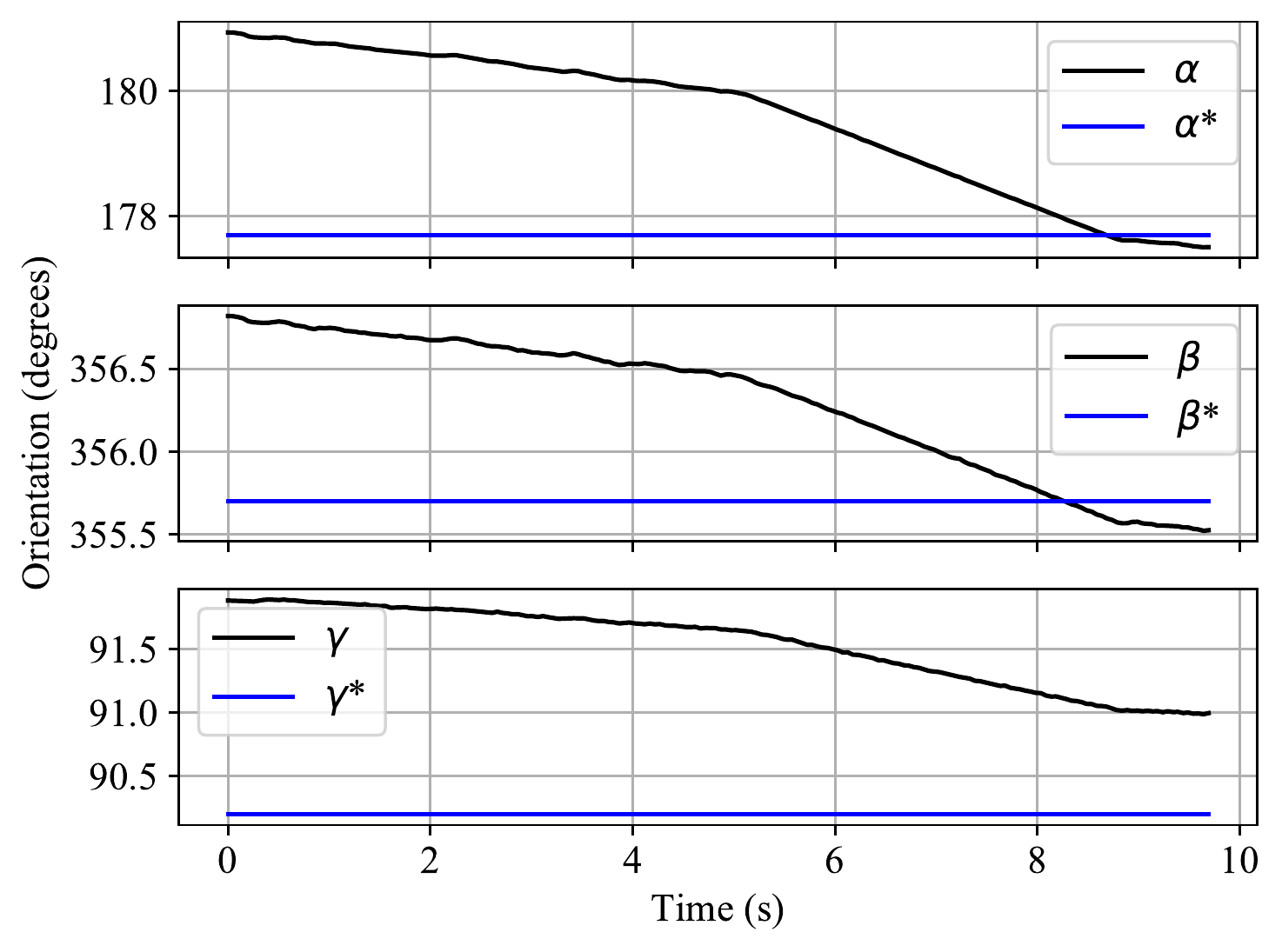}
         \caption{Model 3 prediction}
         \label{fig:oriol4}
     \end{subfigure}
     \begin{subfigure}[b]{0.35\textwidth}
         \centering
         \includegraphics[width=\textwidth]{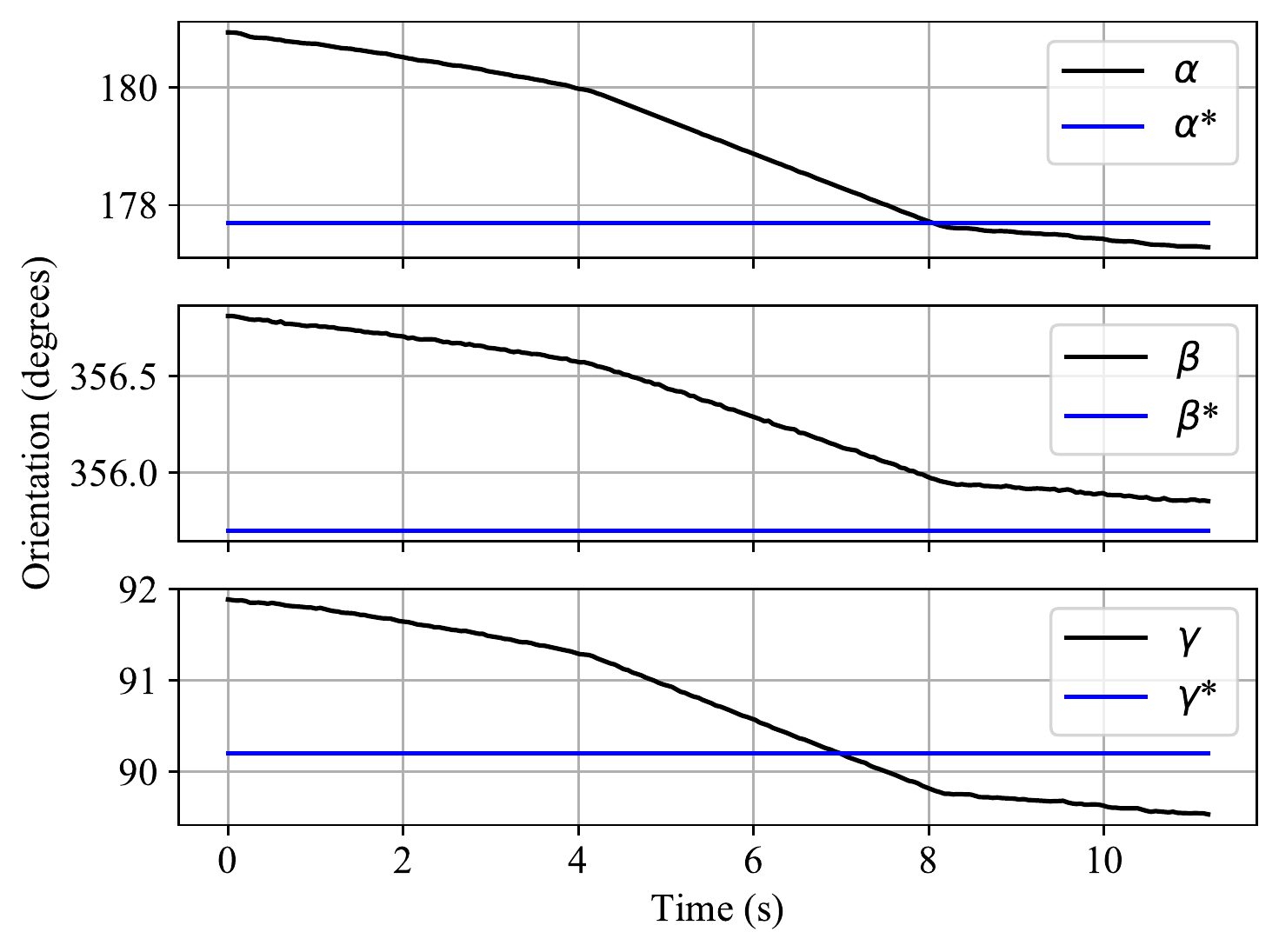}
         \caption{Model 4 prediction}
         \label{fig:oriol5}
     \end{subfigure}
     \caption{Evolution of the robot's orientation in time (black) and its desired value (blue), when the network prediction (weighted by $\lambda_{lin}=0.1$ and $\lambda_{ang}=10$) for an instance of the test set is sent to the robot, which executes it without feedback.}
\label{fig:oriol}
\end{figure*}

The pose of the robot is read throughout the process and illustrated in Figs. \ref{fig:posol} and \ref{fig:oriol}. The best scenario is represented by the application of the prediction given by Model 2 (Figs. \ref{fig:posol3} and \ref{fig:oriol3}).

It must be noted that these results may be different according to the chosen $\lambda$, which interferes both in the convergence time and in the final positioning of the robot. However, as the same experimental conditions were kept for all models, it is possible to make a comparative analysis of efficiency. The largest positioning error for Model 1 is in $z$, which differs by 21mm from the expected. The biggest error in Model 2 is also in $z$, of only 9mm. Model 3 has an error of 16mm in $y$. Although Model 4 has the largest MSE among the models, it is the one that achieves the least positioning error, of 8mm, in $z$. This may indicate that, although the values predicted by Model 4 are more distant from the expected values, they are proportional, which plays a fundamental role in visual servoing.

Concerning orientation, the convergence of control occurs when the line changes its slope. Motion from that point on is just noise, which would be suppressed if there were feedback. The angular displacement that the robot must make in this test instance is not so significant. However, it still shows that the CNNs, especially Model 2, can achieve an error of less than 1 degree in all directions.

Figs. \ref{fig:posol} and \ref{fig:oriol} clearly show how well adapted the networks are for the visual servoing task, as they indicate the difference between the final pose achieved by the robot and the desired one. However, the figures do not accurately express the robot's behavior when the networks are used as actual controllers, in a closed-loop manner, with visual feedback. Furthermore, although they demonstrate generalization for tuples of images not seen in training, they do not demonstrate generalization for unseen objects. These two scenarios are explored in section \ref{vsonlinesection}.

\subsection{Online results}

This section presents the achieved results when the developed algorithms are evaluated in the Kinova Gen3. Therefore, real-world challenges and time constraints can be assessed.
To operate the robot we implemented the trained models in a different hardware setup: an Intel Core i7-7500U processor with 2.9GHz and an Nvidia GPU GeForce 940MX. Even with less computational power, the algorithms run in real-time due to the simplicity of our models.

\subsubsection{Grasp Detection}

\begin{figure*}[t!]
     \centering
     \begin{subfigure}[t]{0.31\textwidth}
         \centering
         \includegraphics[width=\textwidth]{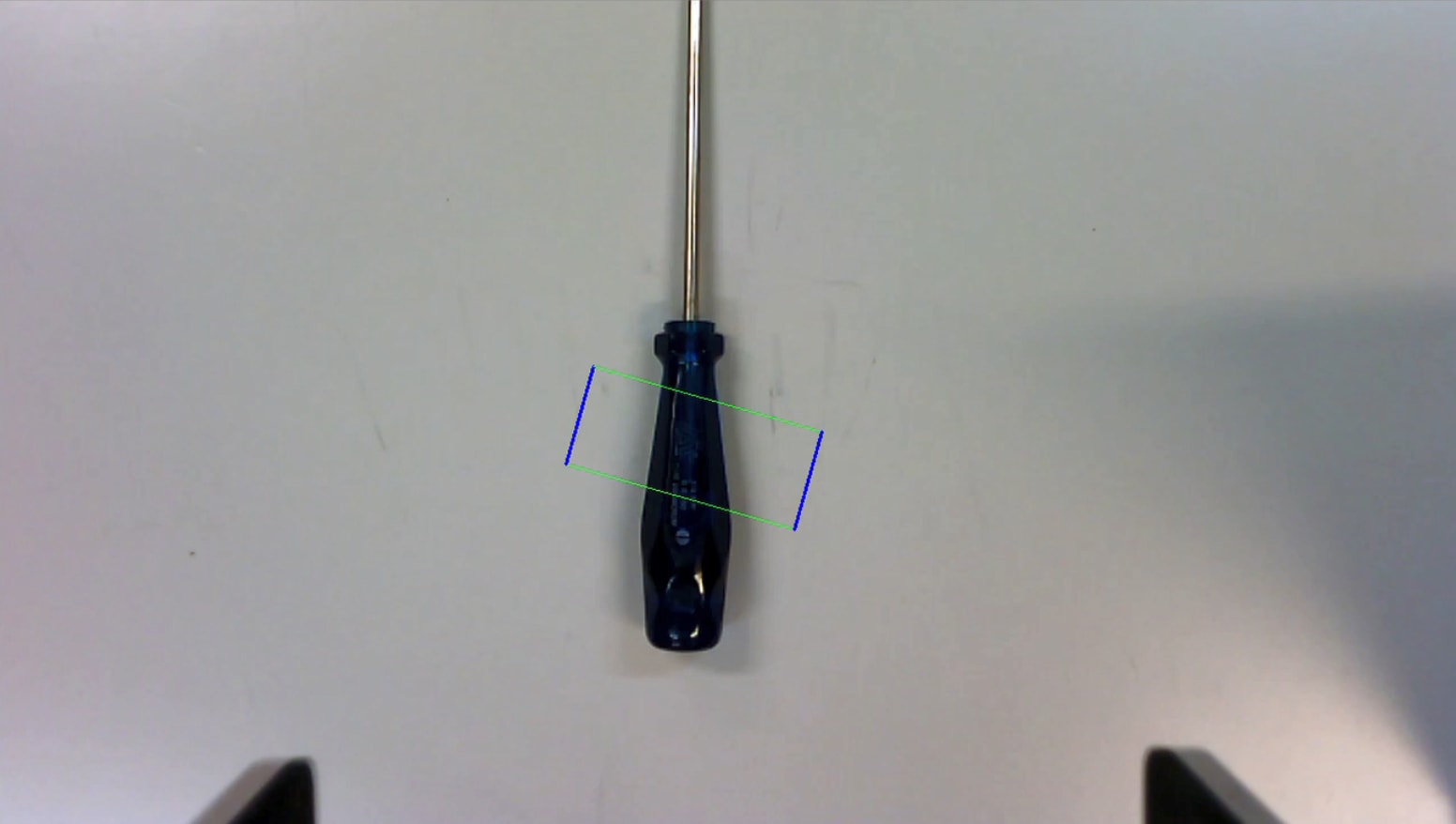}
         \caption{Screwdriver}
         \label{fig:fenda}
     \end{subfigure}
     \hfill
     \begin{subfigure}[t]{0.31\textwidth}
         \centering
         \includegraphics[width=\textwidth]{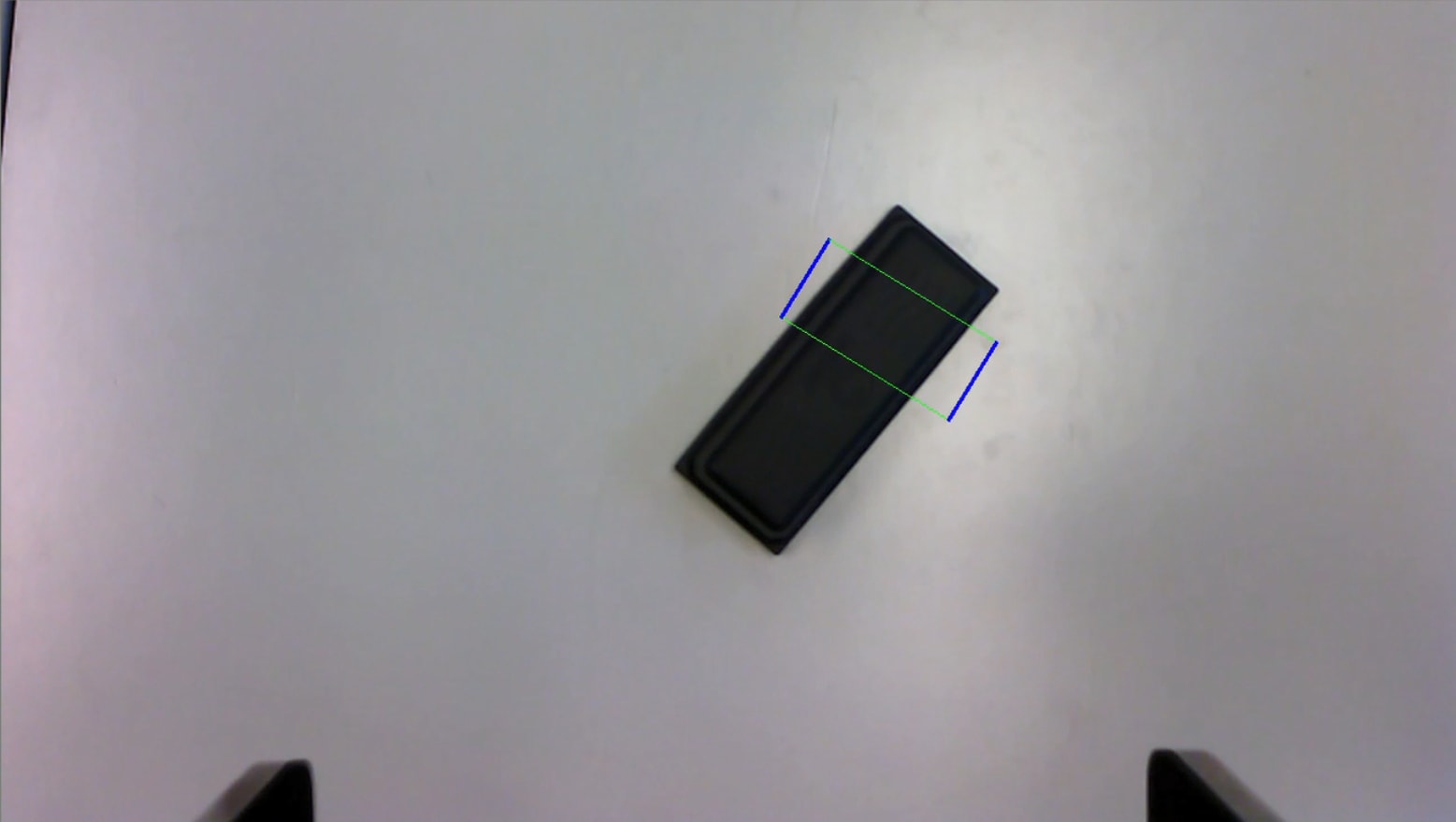}
         \caption{Chalkboard eraser}
         \label{fig:apagador}
     \end{subfigure}
     \hfill
     \begin{subfigure}[t]{0.31\textwidth}
         \centering
         \includegraphics[width=\textwidth]{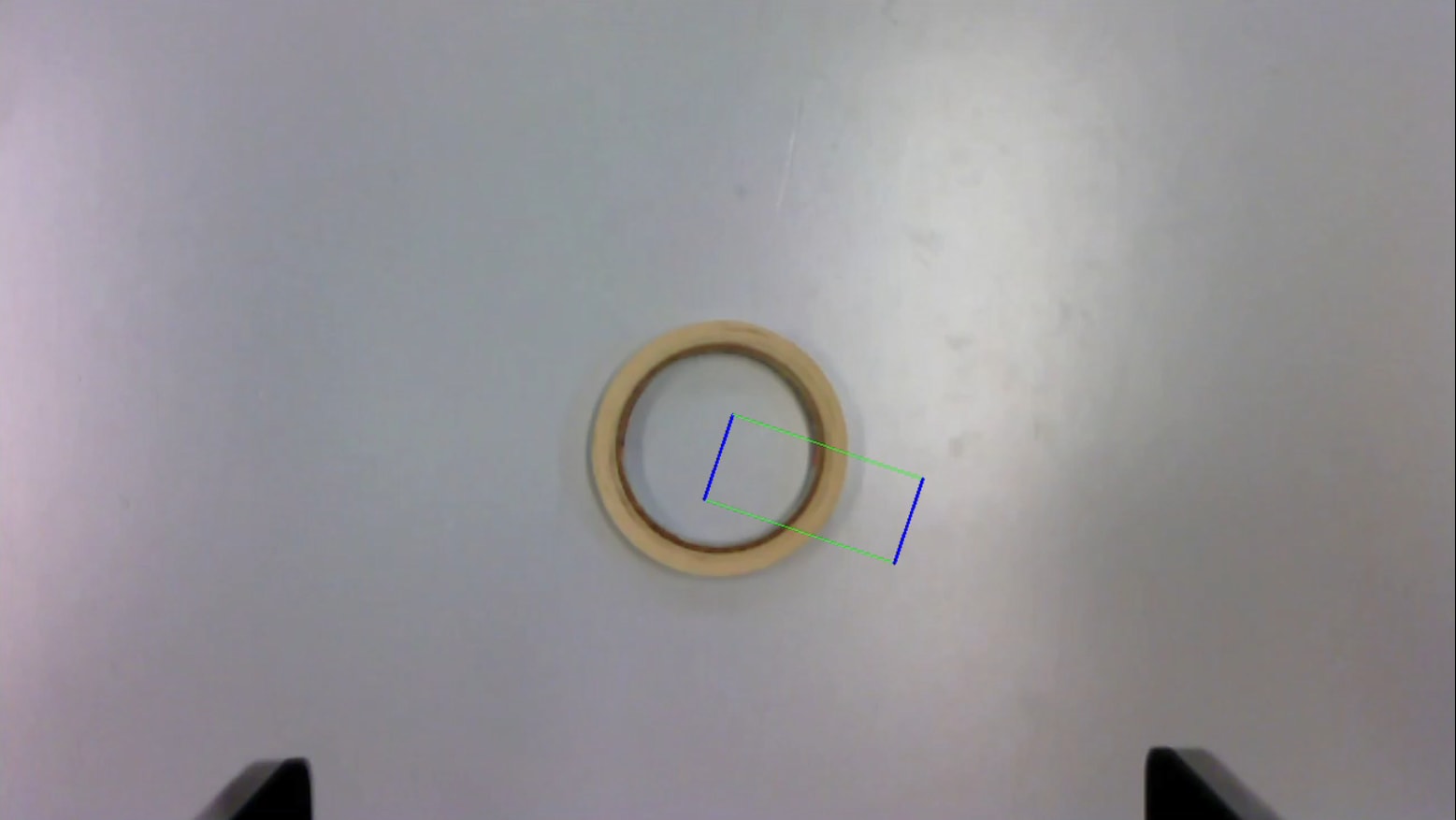}
         \caption{Thin masking tape}
         \label{fig:crepefina}
     \end{subfigure}
     \begin{subfigure}[t]{0.31\textwidth}
         \centering
         \includegraphics[width=\textwidth]{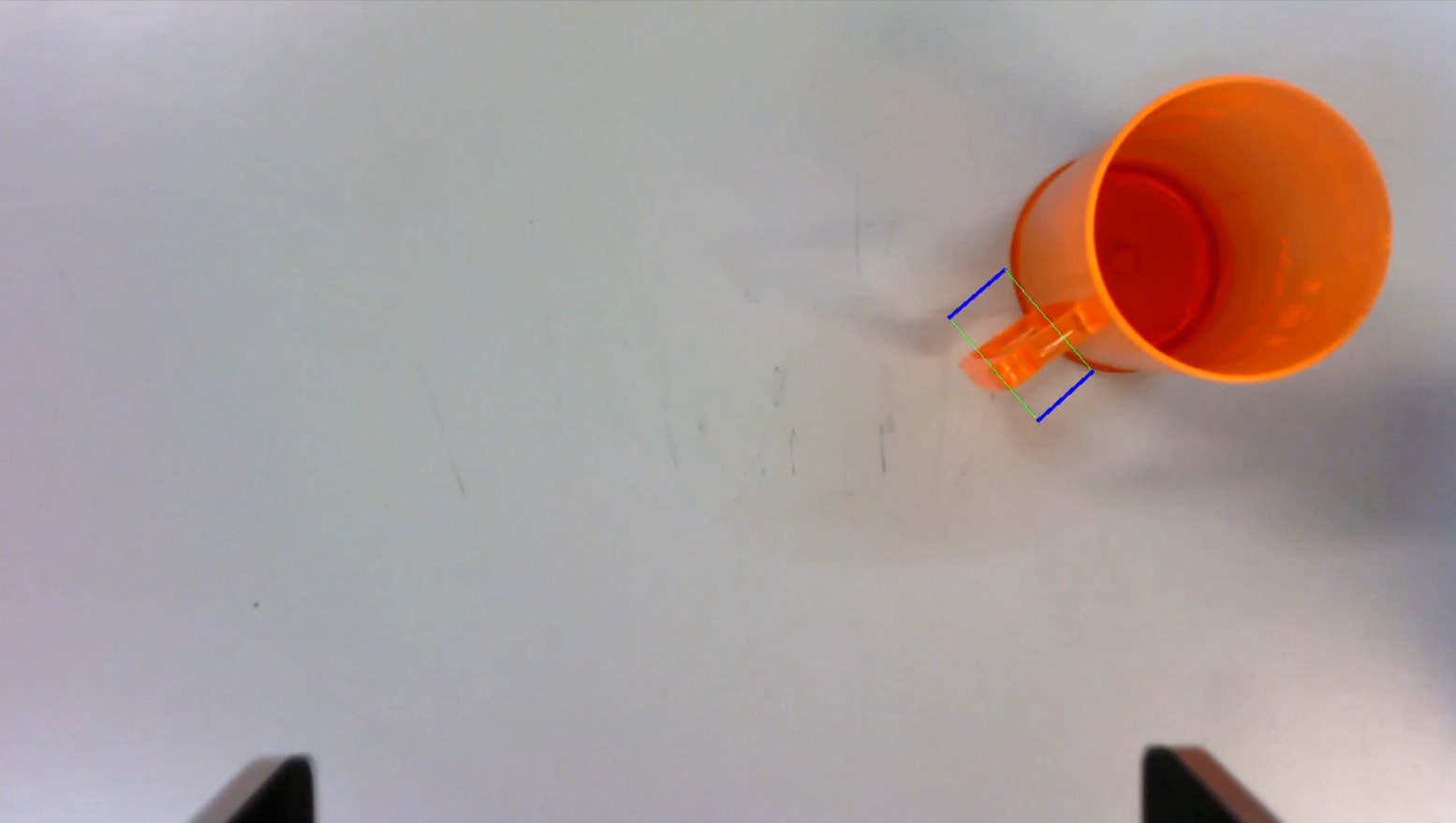}
         \caption{Cup}
         \label{fig:copo}
     \end{subfigure}
     \hfill
     \begin{subfigure}[t]{0.31\textwidth}
         \centering
         \includegraphics[width=\textwidth]{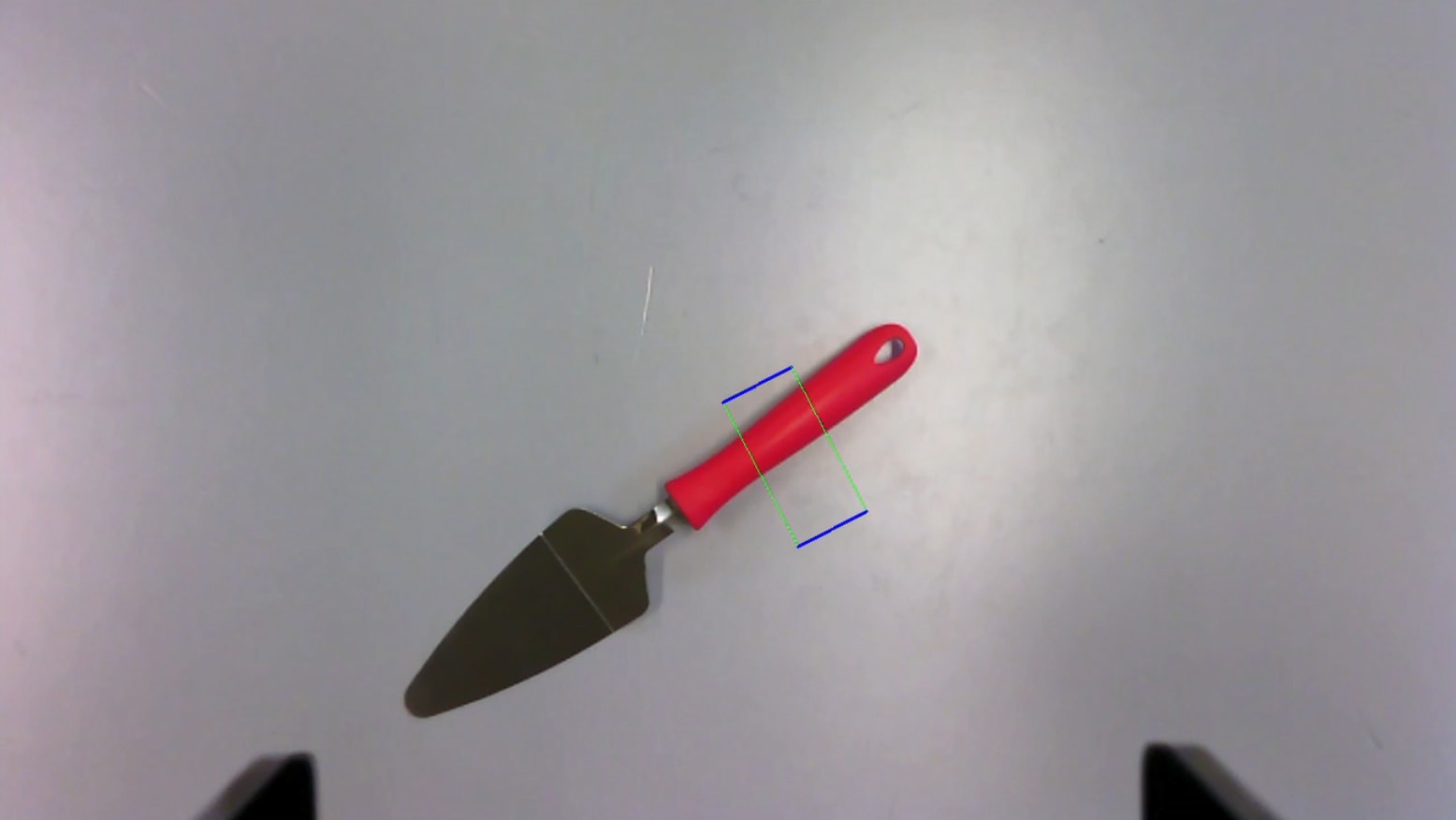}
         \caption{Pie knife}
         \label{fig:espatula}
     \end{subfigure}
     \hfill
     \begin{subfigure}[t]{0.31\textwidth}
         \centering
         \includegraphics[width=\textwidth]{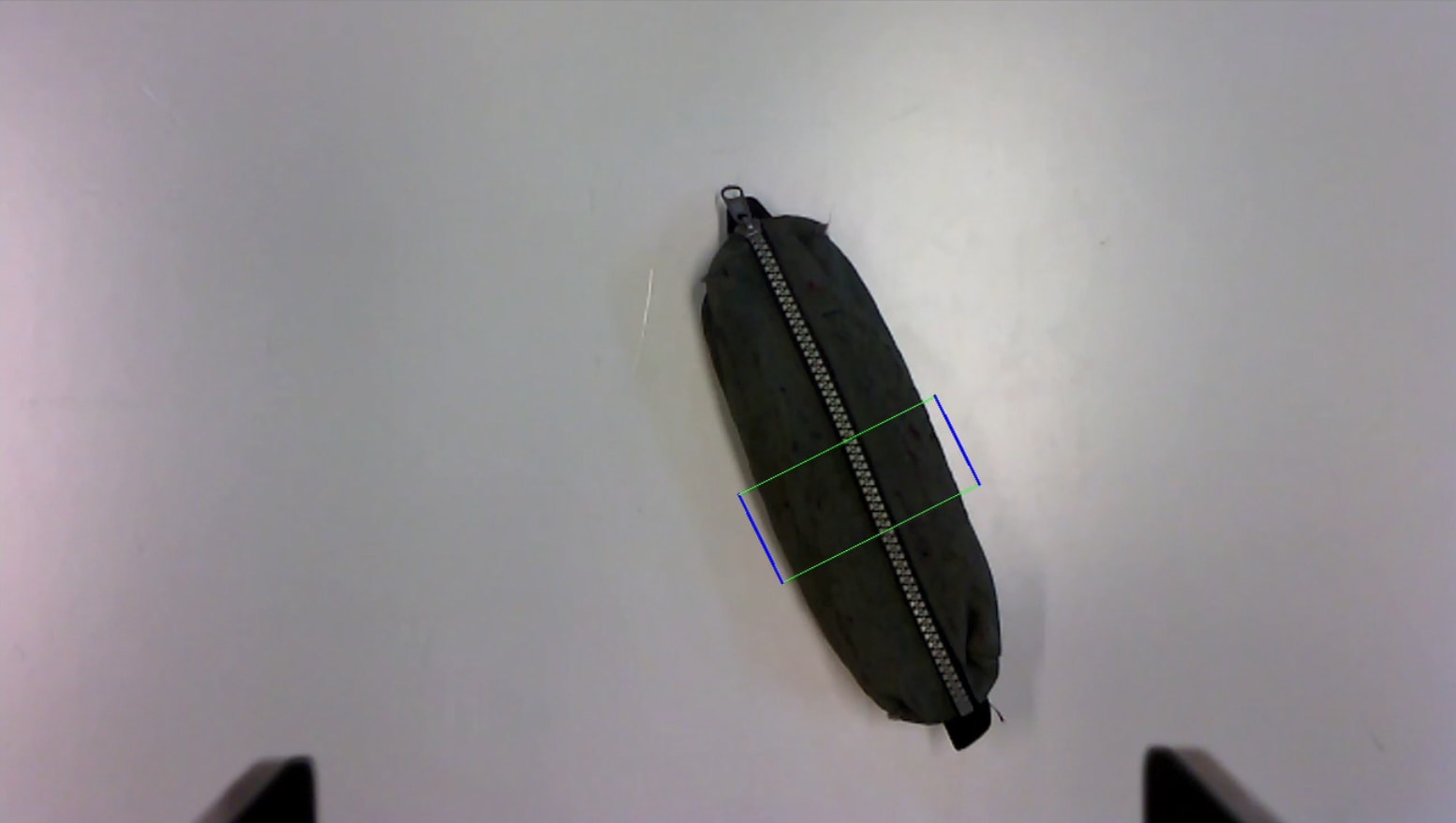}
         \caption{Pencil case}
         \label{fig:estojo}
     \end{subfigure}
     \begin{subfigure}[t]{0.31\textwidth}
         \centering
         \includegraphics[width=\textwidth]{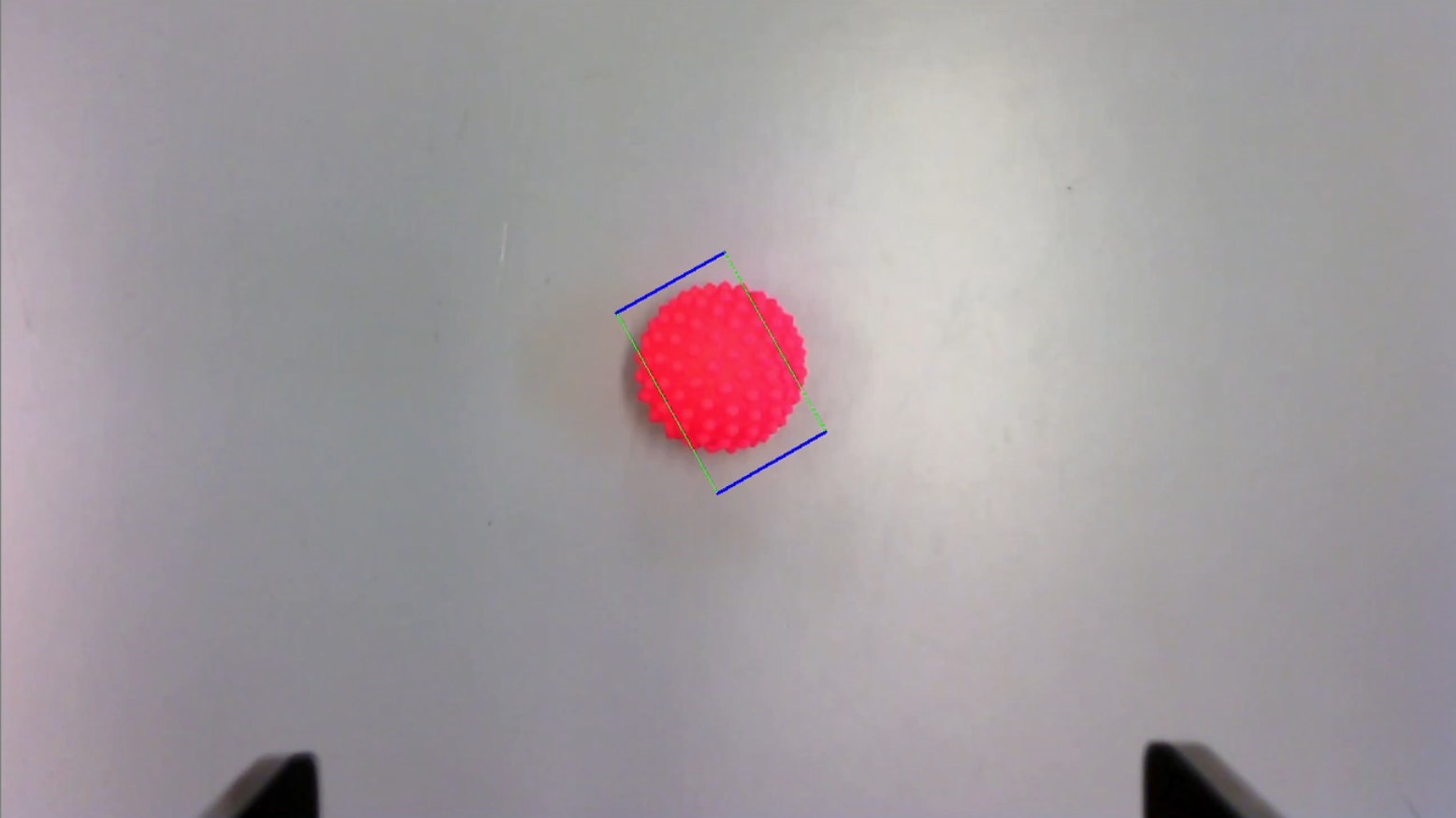}
         \caption{Hand massage ball}
         \label{fig:bola}
     \end{subfigure}
     \hfill
     \begin{subfigure}[t]{0.31\textwidth}
         \centering
         \includegraphics[width=\textwidth]{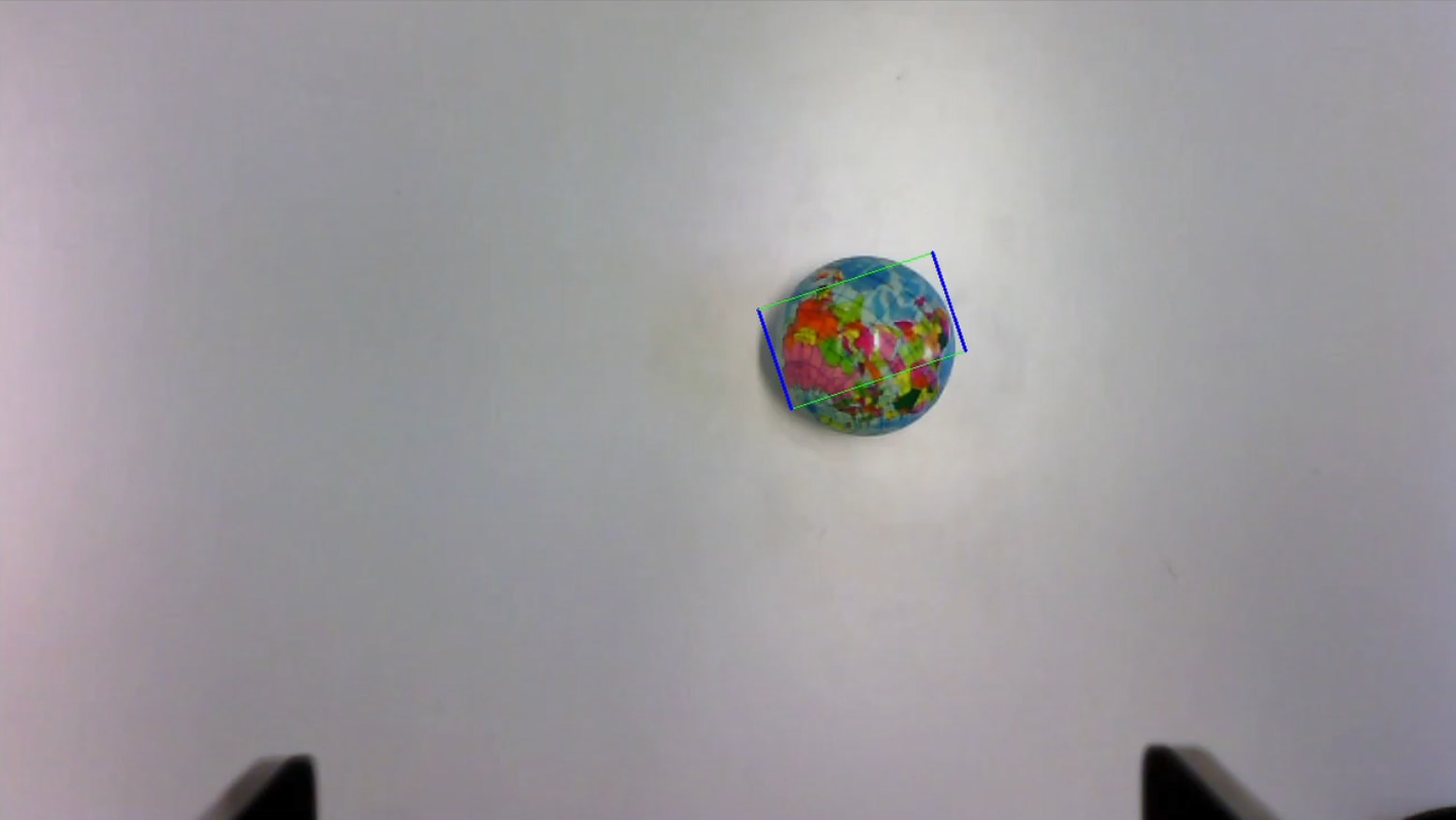}
         \caption{Anti stress ball}
         \label{fig:globo}
     \end{subfigure}
     \hfill
     \begin{subfigure}[t]{0.31\textwidth}
         \centering
         \includegraphics[width=\textwidth]{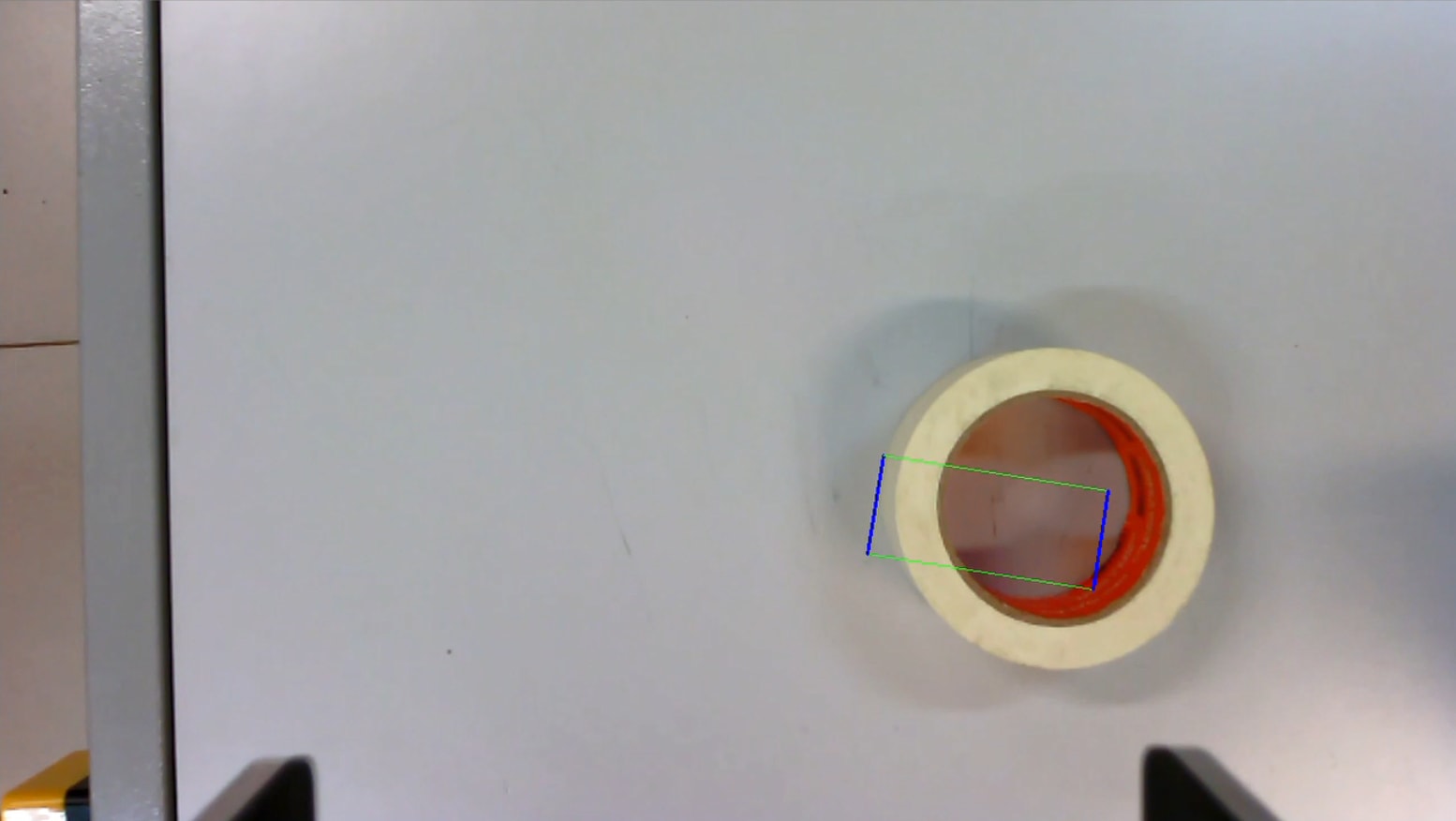}
         \caption{Thick masking tape}
         \label{fig:crepe}
     \end{subfigure}
     \begin{subfigure}[t]{0.31\textwidth}
         \centering
         \includegraphics[width=\textwidth]{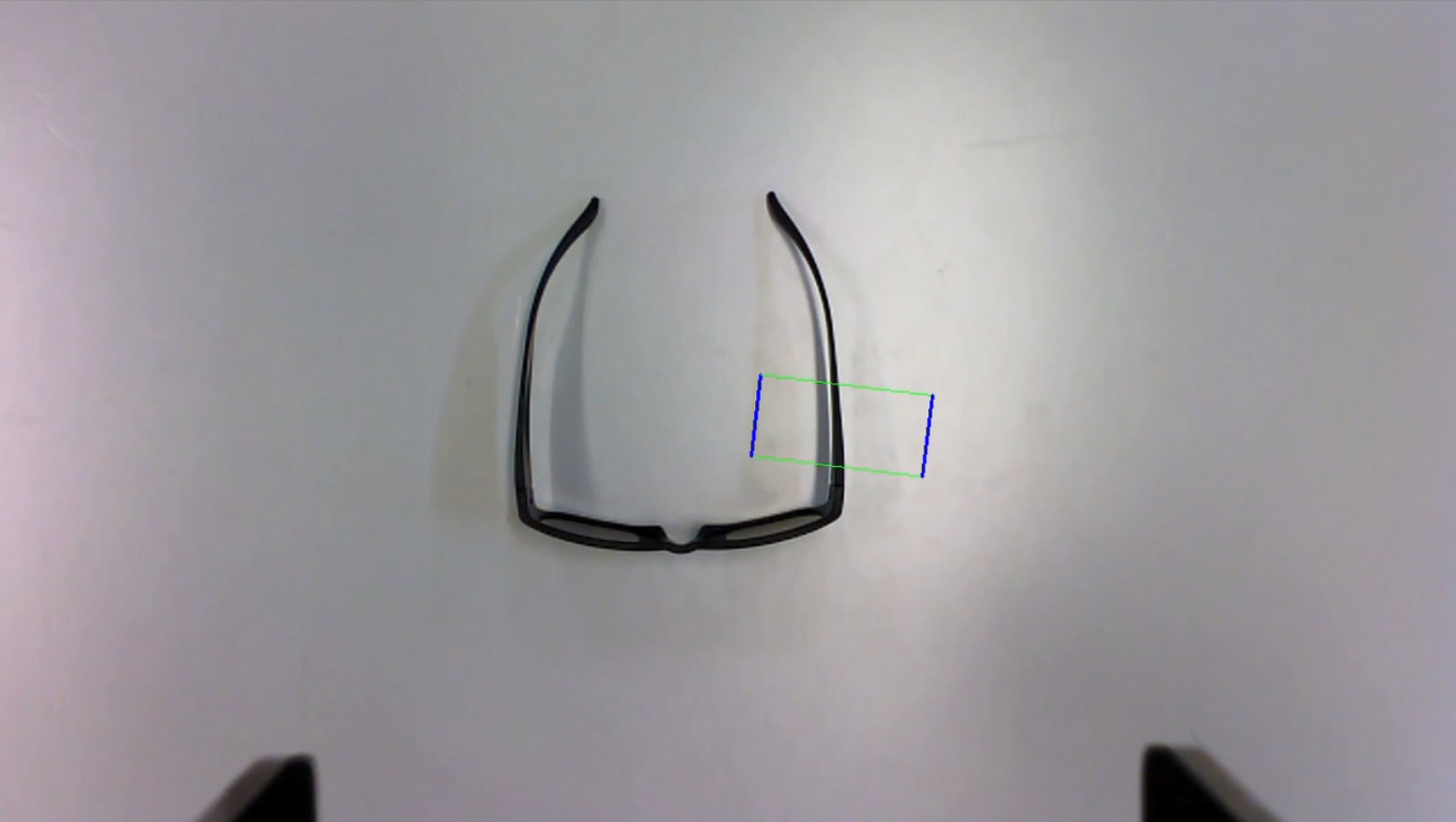}
         \caption{3D glasses}
         \label{fig:oculos}
     \end{subfigure}
     \hfill
     \begin{subfigure}[t]{0.31\textwidth}
         \centering
         \includegraphics[width=\textwidth]{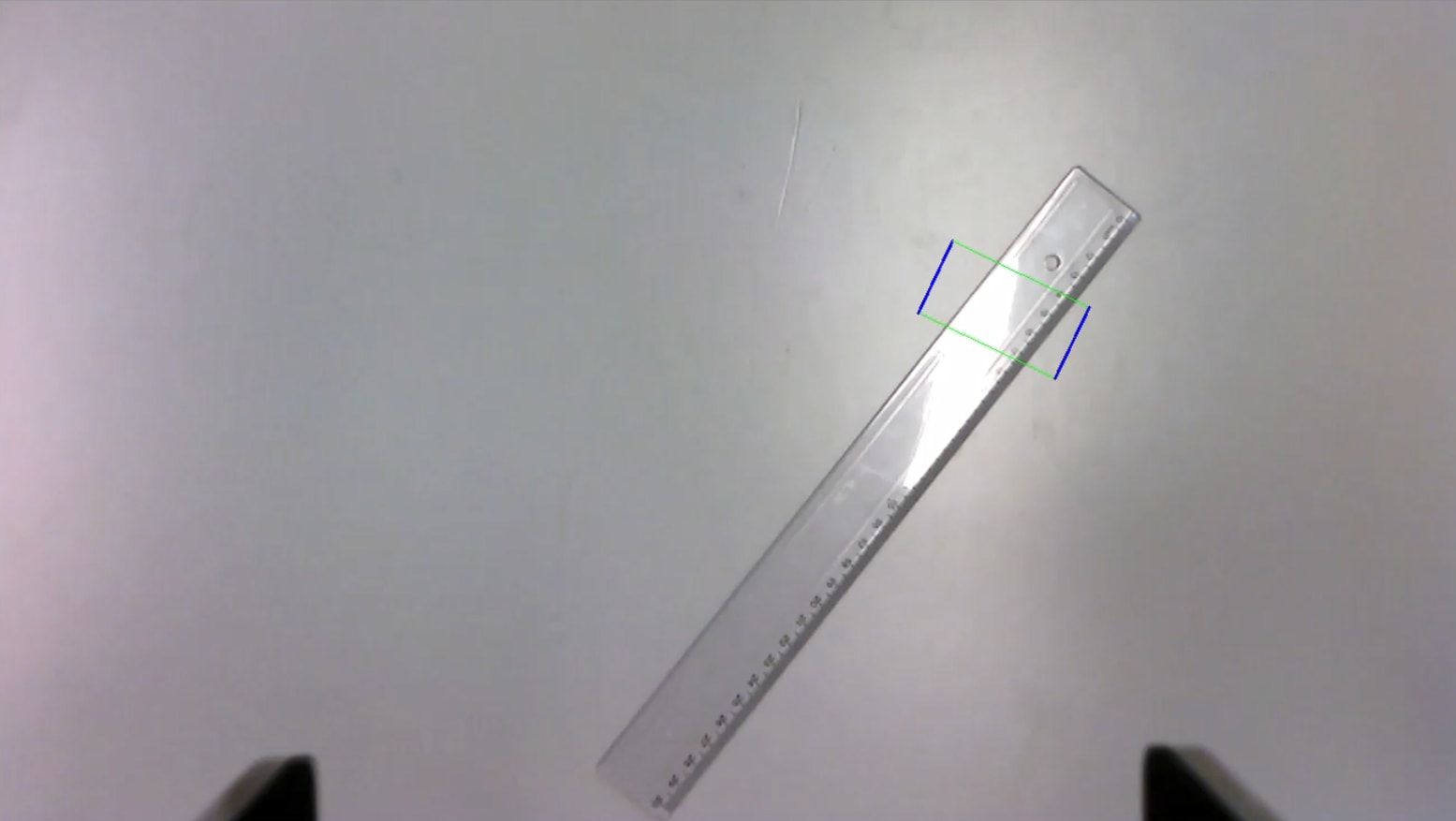}
         \caption{Ruler}
         \label{fig:regua}
     \end{subfigure}
     \hfill
     \begin{subfigure}[t]{0.31\textwidth}
         \centering
         \includegraphics[width=\textwidth]{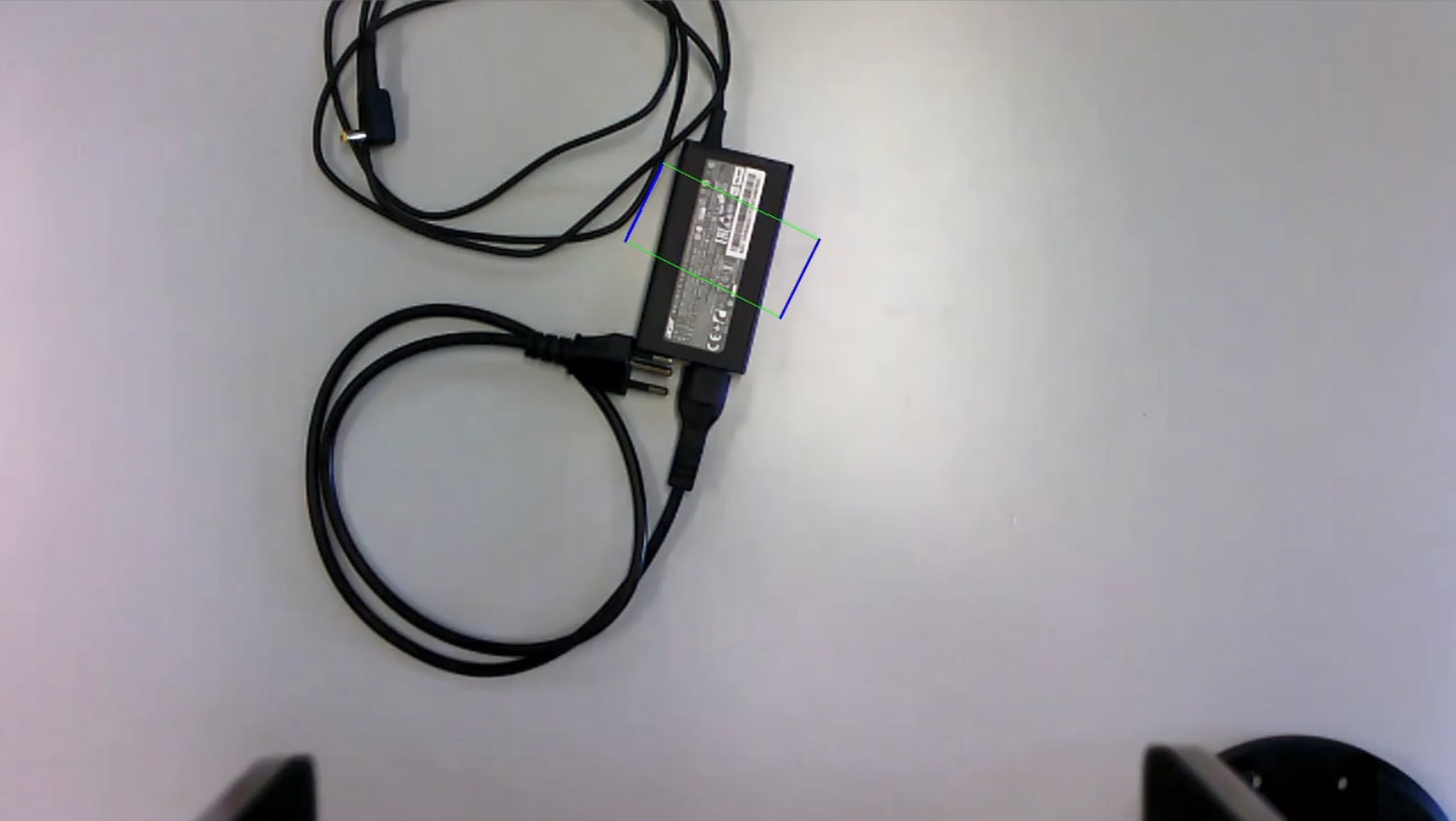}
         \caption{Laptop charger}
         \label{fig:carregador}
     \end{subfigure}
     \begin{subfigure}[t]{0.31\textwidth}
         \centering
         \includegraphics[width=\textwidth]{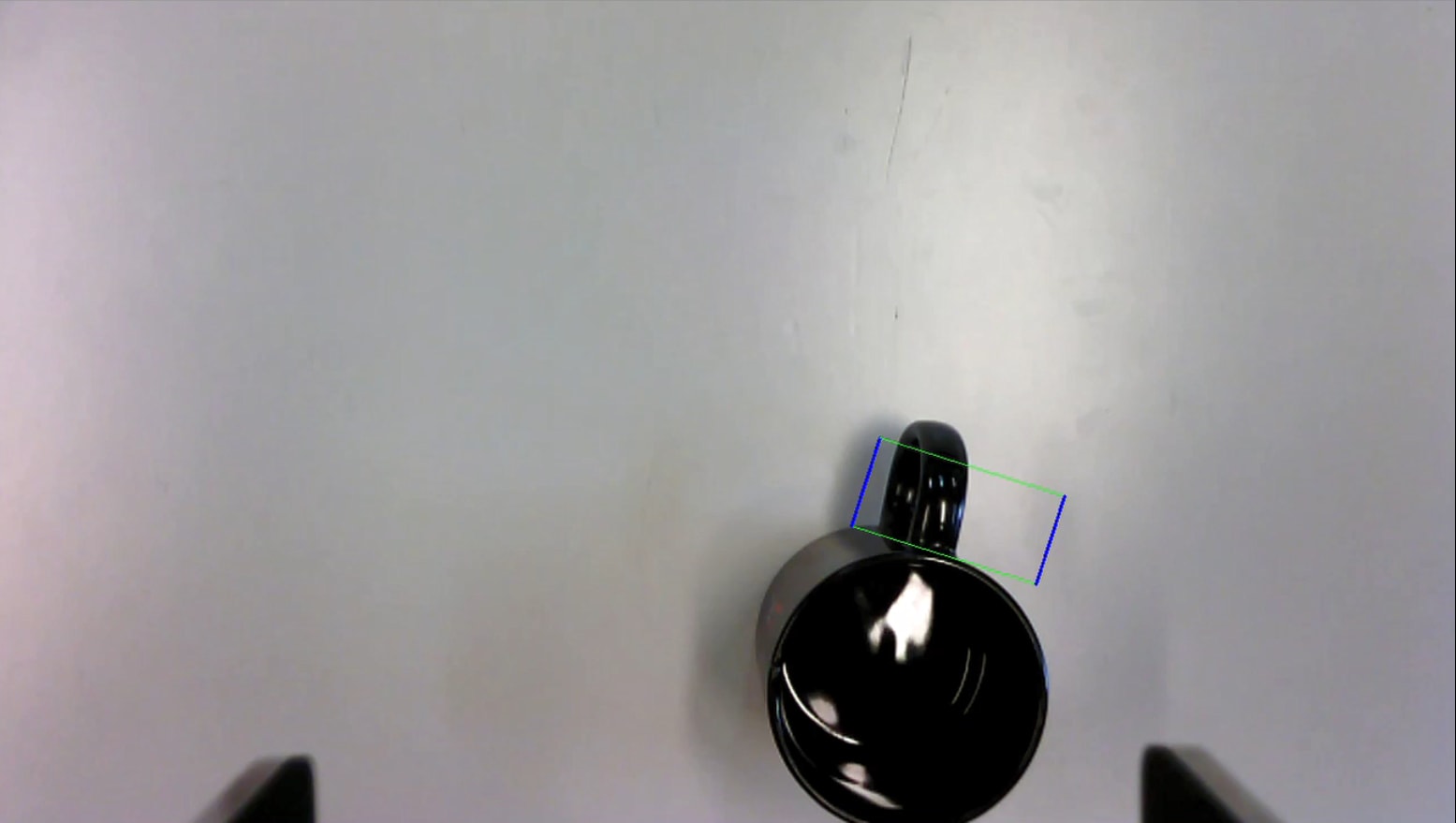}
         \caption{Mug}
         \label{fig:caneca}
     \end{subfigure}
     \hfill
     \begin{subfigure}[t]{0.31\textwidth}
         \centering
         \includegraphics[width=\textwidth]{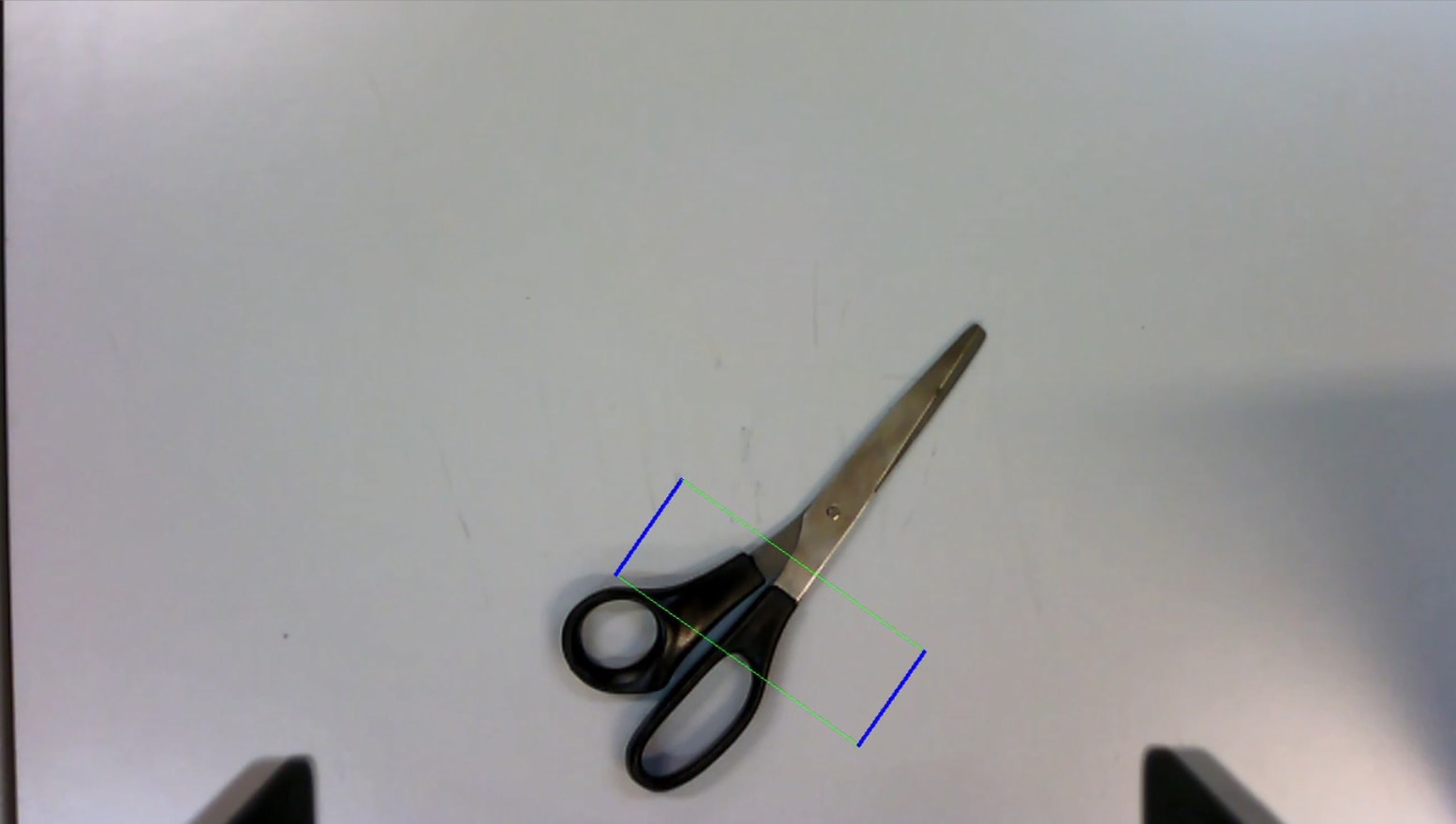}
         \caption{Scissors 1}
         \label{fig:tesoura}
     \end{subfigure}
     \hfill
     \begin{subfigure}[t]{0.31\textwidth}
         \centering
         \includegraphics[width=\textwidth]{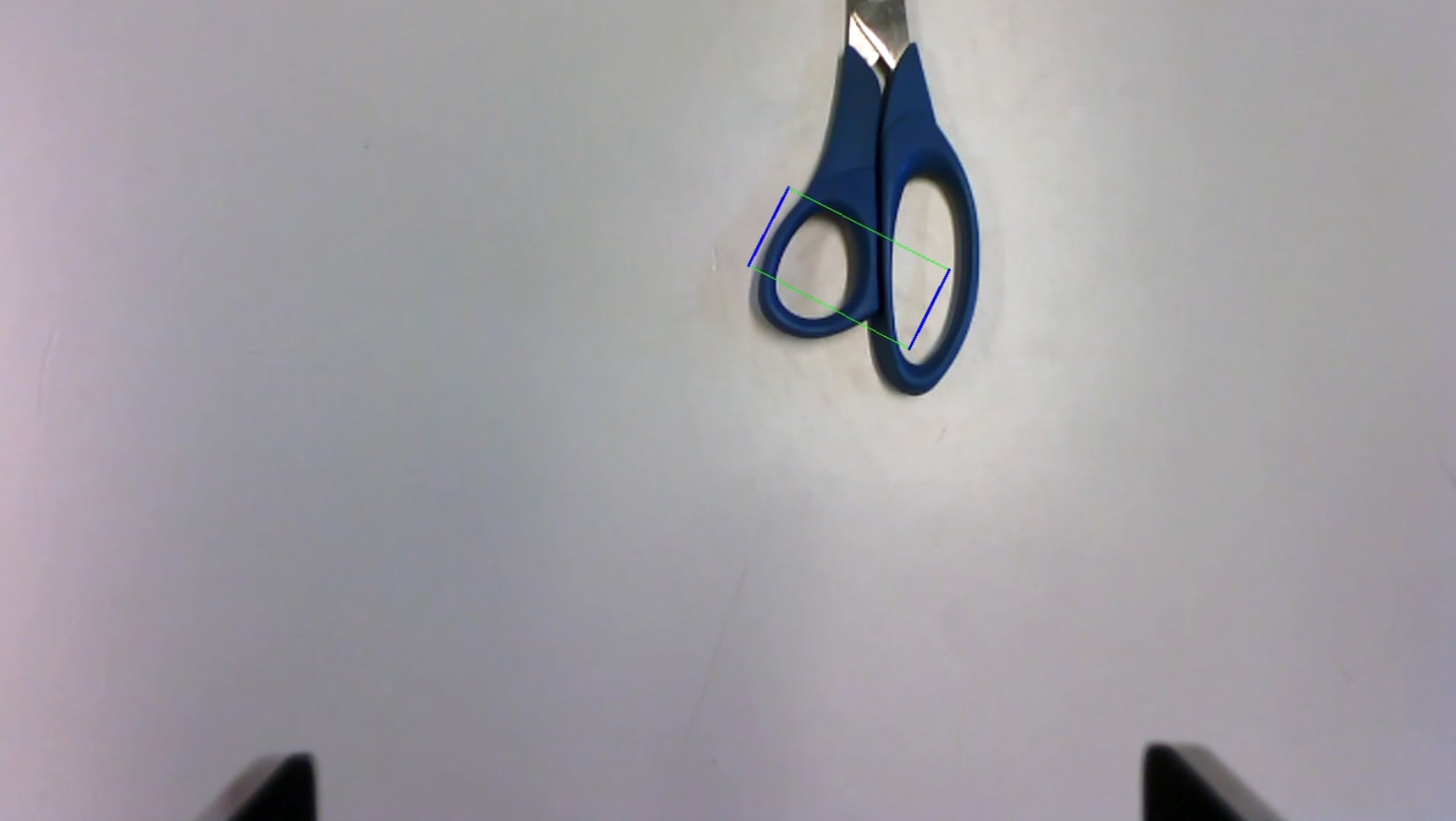}
         \caption{Scissors 2}
         \label{fig:tesoura2}
     \end{subfigure}
     \begin{subfigure}[t]{0.31\textwidth}
         \centering
         \includegraphics[width=\textwidth]{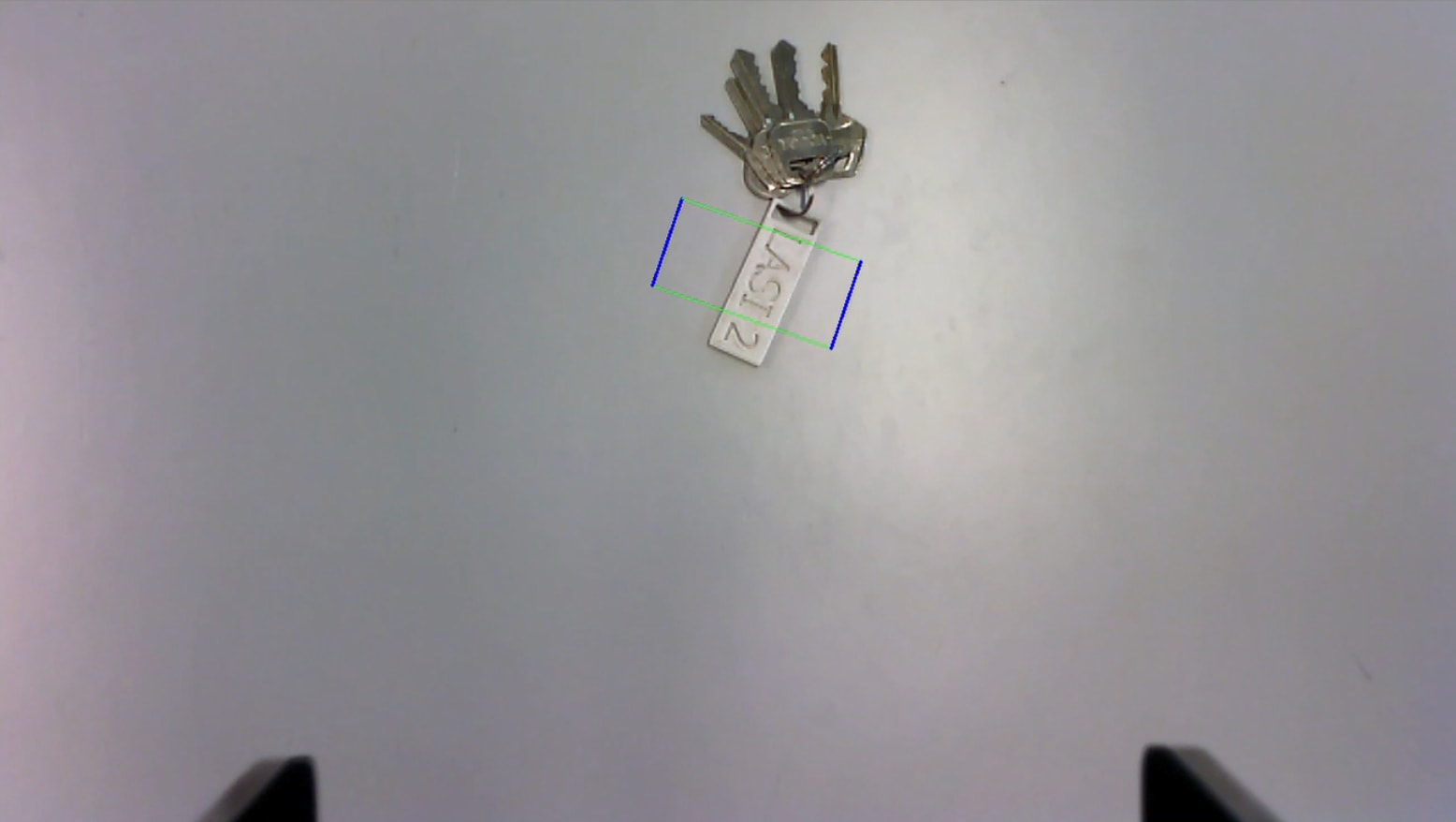}
         \caption{Keychain + Bunch of keys}
         \label{fig:chave}
     \end{subfigure}
     \hfill
     \begin{subfigure}[t]{0.31\textwidth}
         \centering
         \includegraphics[width=\textwidth]{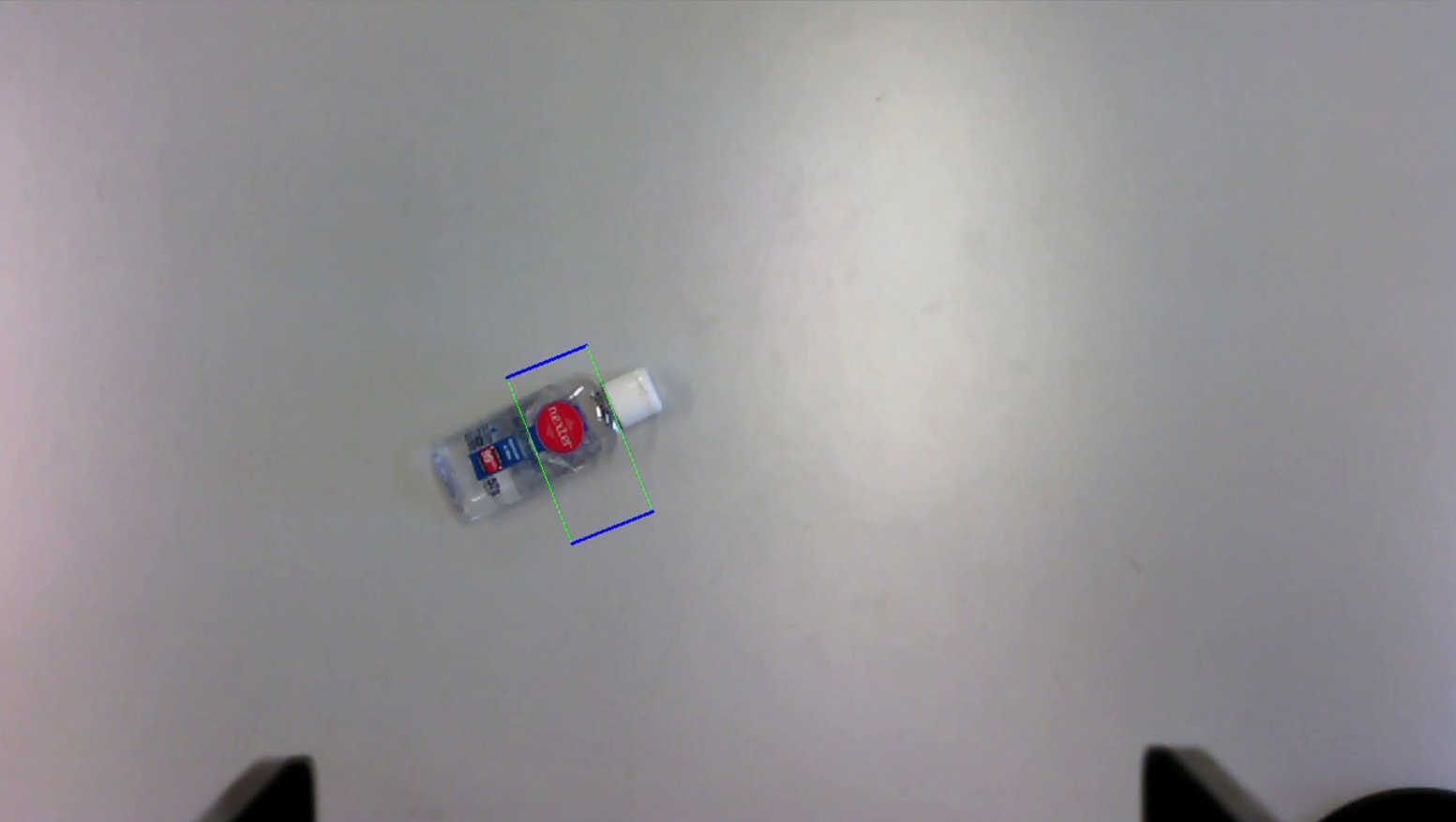}
         \caption{Hand sanitizer}
         \label{fig:alcoolgel}
     \end{subfigure}
     \hfill
     \begin{subfigure}[t]{0.31\textwidth}
         \centering
         \includegraphics[width=\textwidth]{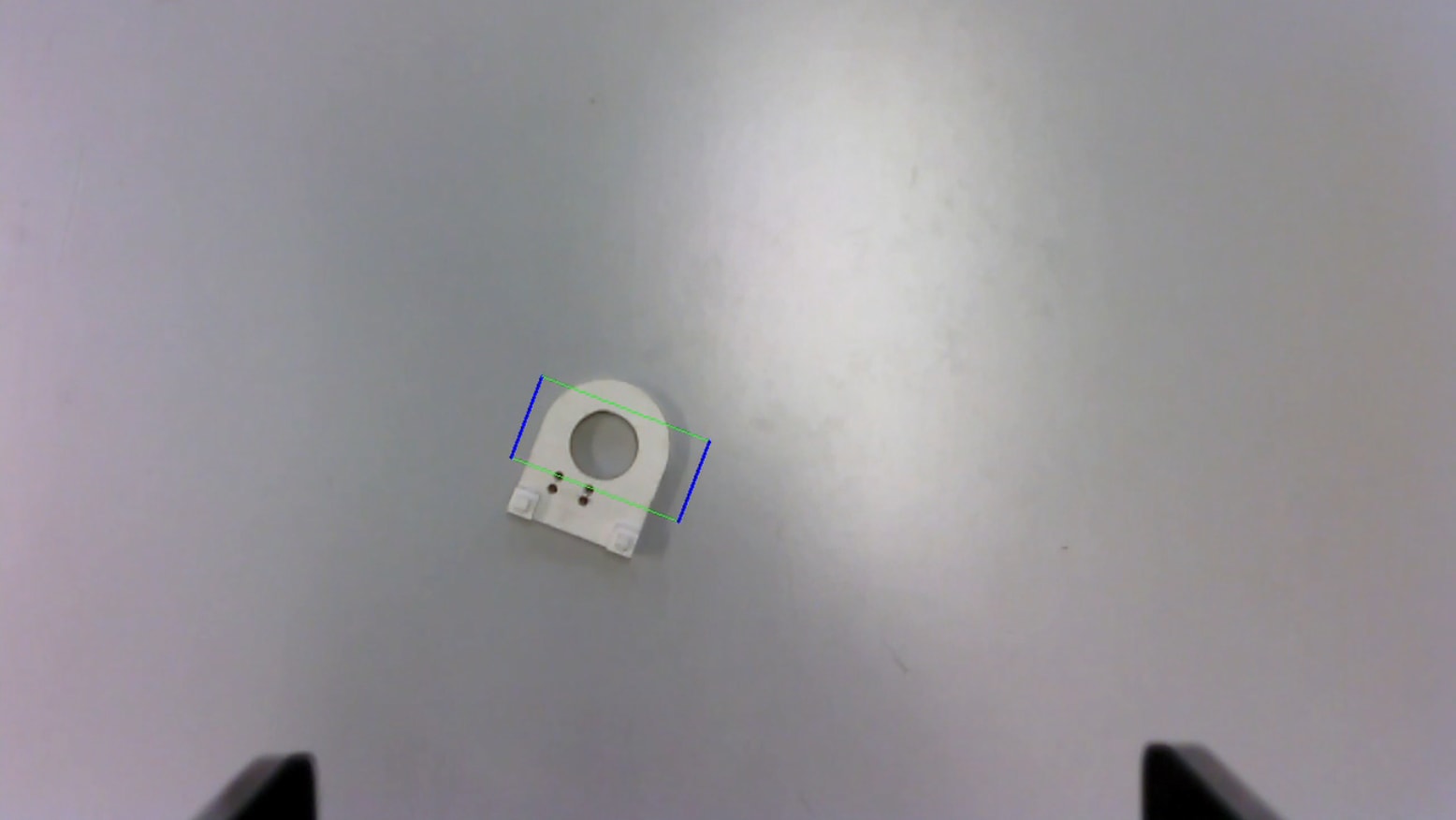}
         \caption{Random polymer device}
         \label{fig:pendulocirc}
     \end{subfigure}
     \caption{Network prediction for different objects first seen by the robot}
        \label{fig:grasprobot}
\end{figure*}

The grasp detection results are presented in a qualitative way, showing the predicted grasp rectangle in the image acquired by the robot for different objects. Once the rectangle is correctly predicted, the grasp is executed considering a $2D\rightarrow 3D$ mapping that uses the camera's intrinsic and extrinsic parameters, and a fixed depth for the gripper. The network application scenario is quite different from that represented in the CGD, primarily due to the different objects, and also due to the considerable difference in luminosity. Fig. \ref{fig:grasprobot} illustrates eighteen detection examples. 

In general, the figure makes it evident that the predicted rectangles (grippers indicated in blue) are consistent with the shape of the target objects, adapting to their sizes and orientations. Some particular cases, however, deserve more attention.

The objects represented in Figs. \ref{fig:fenda} and \ref{fig:espatula} are the easiest for the network to detect grasps, as they are large in length and small in thickness. Most of the objects that are commonly handled follow this pattern, and therefore have a large number of instances in the CGD (at least 50 objects following this pattern).

The chalkboard eraser (Fig. \ref{fig:apagador}) and pencil case (Fig. \ref{fig:estojo}) are not as simple as the previous objects because they are thicker. The network needs to reason whether the object width exceeds the represented gripper opening. This same situation occurs with circular objects, as illustrated by the detections in the hand massage ball (Fig. \ref{fig:bola}) and anti-stress ball (Fig. \ref{fig:globo}), in which the smaller ball is easier to detect. However, circular objects can also confuse the network due to the infinite possibilities of orientation.

\begin{figure*}[!htb]
\centering
     \begin{subfigure}[b]{0.31\textwidth}
         \centering
         \includegraphics[width=\textwidth]{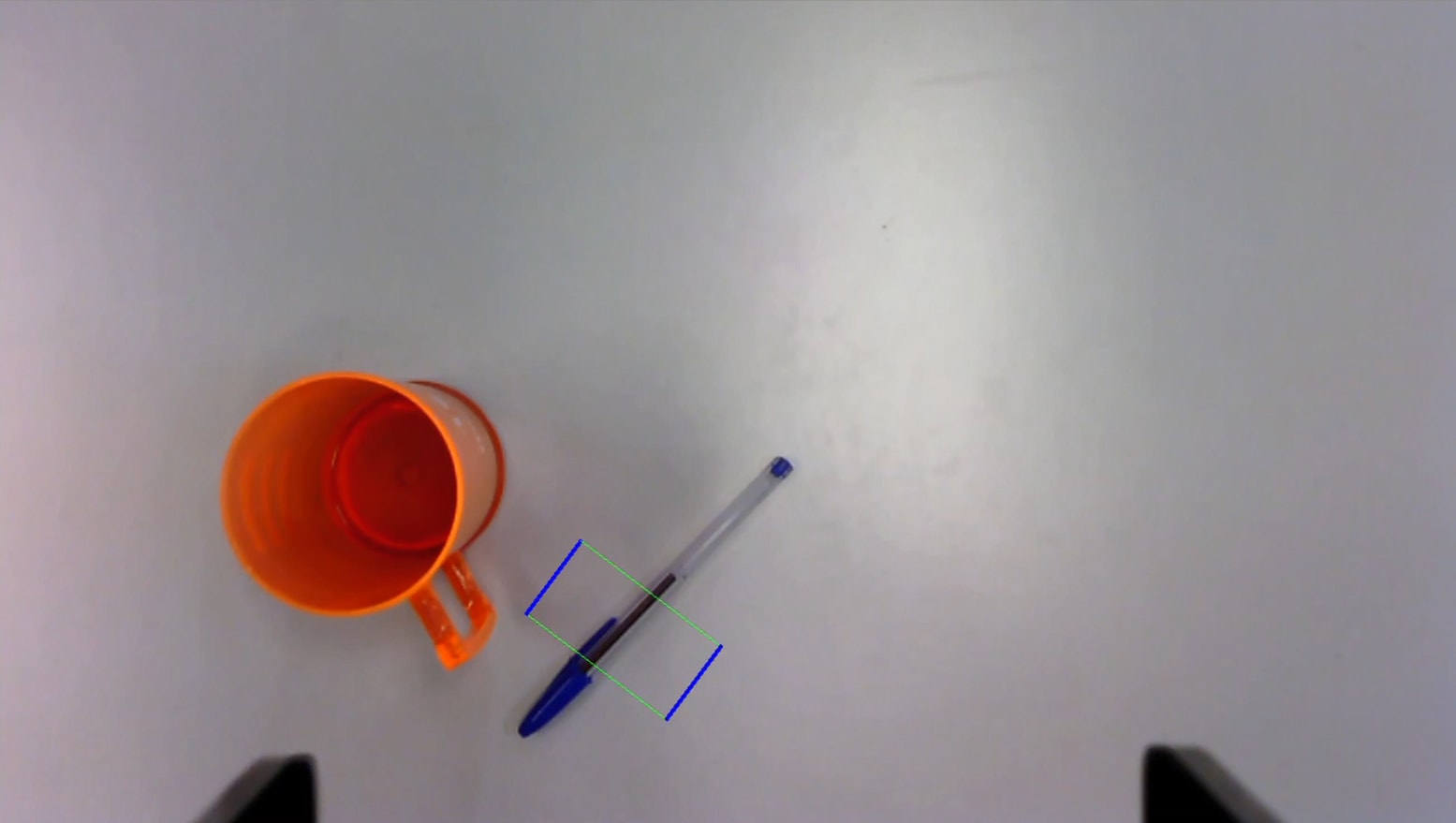}
         \caption{Cup and pen A}
         \label{fig:copocaneta}
     \end{subfigure}
     \hfill
     \begin{subfigure}[b]{0.31\textwidth}
         \centering
         \includegraphics[width=\textwidth]{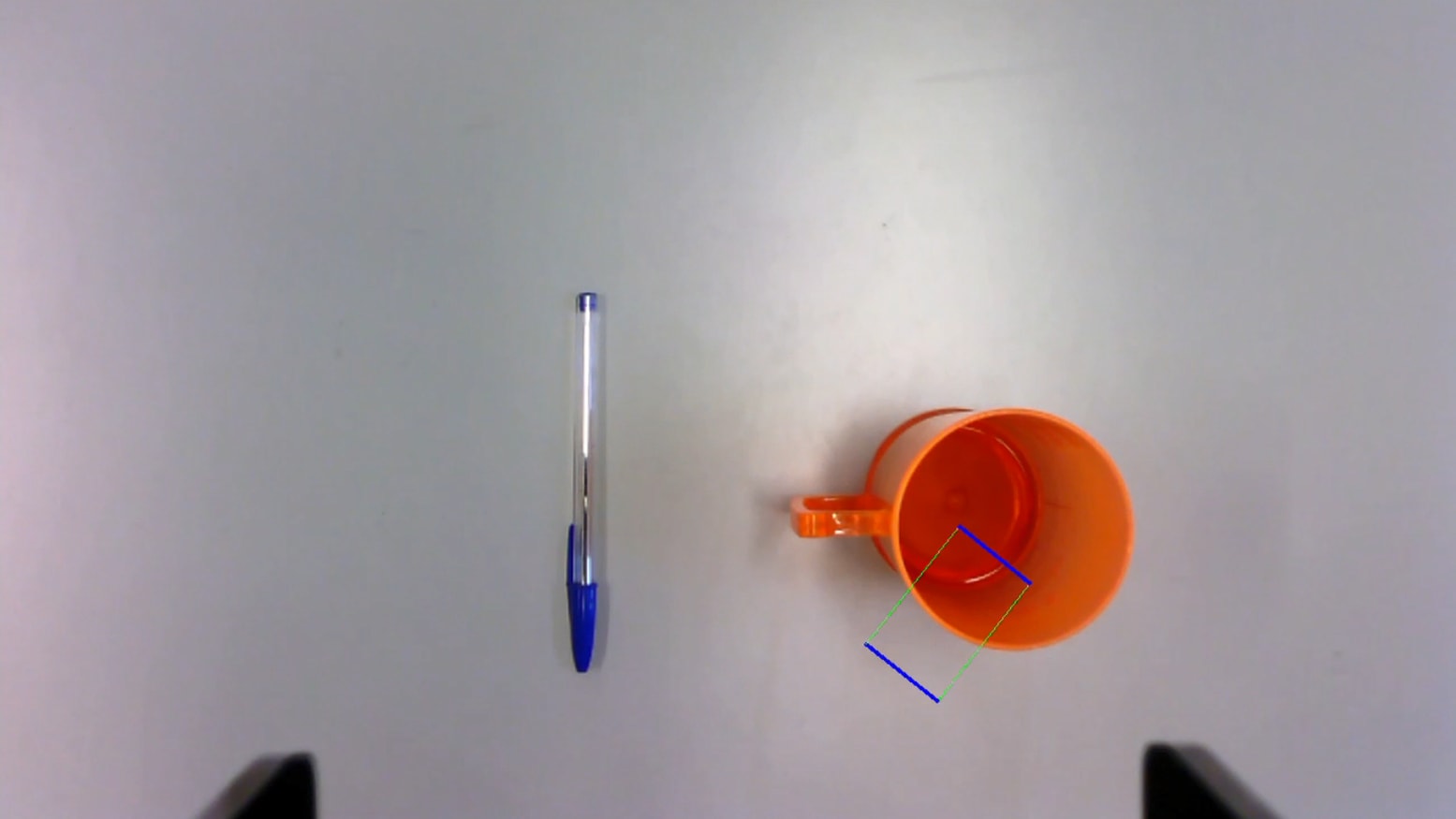}
         \caption{Cup and pen B}
         \label{fig:canetacopo}
     \end{subfigure}
     \hfill
     \begin{subfigure}[b]{0.31\textwidth}
         \centering
         \includegraphics[width=\textwidth]{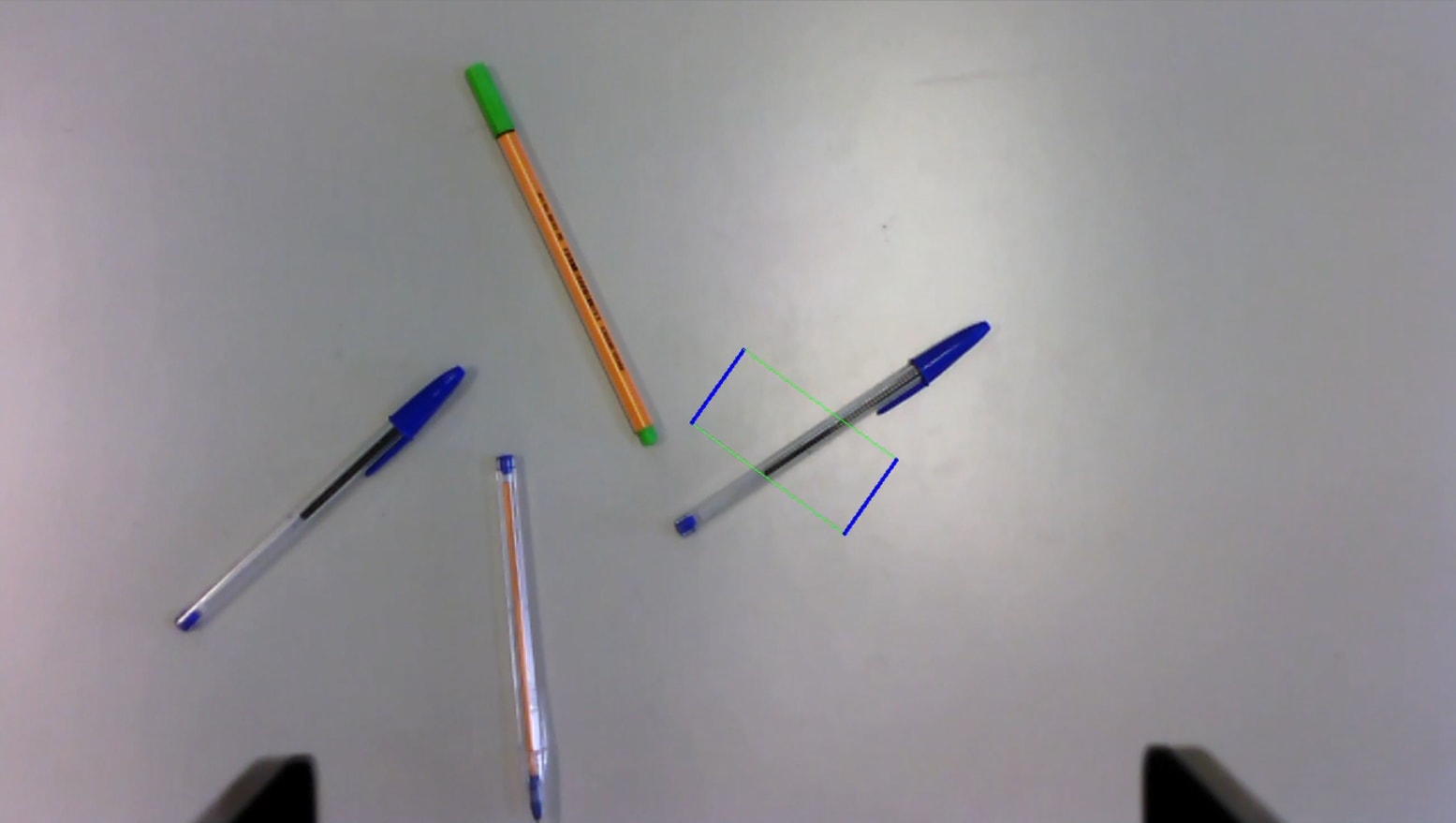}
         \caption{Pens}
         \label{fig:canetas}
     \end{subfigure}
     \caption{Grasp network prediction when confronted with multiple objects} 
        \label{fig:grasprobot2}
\end{figure*}

The masking tapes in Figs. \ref{fig:crepefina} and \ref{fig:crepe} also exemplify the thickness problem. But the great challenge regarding these objects is the ring format. The network may often reason that the middle of the object is a constituent part of it, especially if the object and the background have similar colors. Besides, the prediction of a grasp rectangle in the middle of a circular object (given the arrangement of its labels) is only surpassed because only one label is used during training.

Cups and mugs may also offer a challenge to the network, as they can be caught both by the edges and by the handles. In both Figs. \ref{fig:copo} and \ref{fig:caneca}, the CNN predicts a rectangle on the handles, which, especially in the case of Fig. \ref{fig:caneca}, can be justified by the prominence of the handle and the fact that the edges are not so distinguishable.

The ruler (Fig. \ref{fig:regua}) and hand sanitizer (Fig. \ref{fig:alcoolgel}) are transparent and may be reflective, however, the network manages to distinguish them. The scissors in Figs. \ref{fig:tesoura} and \ref{fig:tesoura2} and the device in Fig. \ref{fig:pendulocirc}, test the network's ability to reason on scale. The first pair of scissors is further away from the camera, hence the network prefers to predict a large rectangle. The second pair is closer, and the network can have a better understanding of how big the hole between the handles is, so it can make a more skillful grasp. The device in Fig. \ref{fig:pendulocirc} exemplifies the same situation. However, whether near or far, the network always prefers the larger opening, since it understands that the gripper does not fit in the hole.

To conclude the discussion on Fig. \ref{fig:grasprobot}, the objects in Figs. \ref{fig:oculos}, \ref{fig:carregador} and \ref{fig:chave} test the network's ability to find the best grasp location amid some  possibilities. It is possible to note that the CNN makes wise choices, especially regarding the laptop charger, when choosing the part of the object that is most likely to result in a stable grasp.

A final analysis comprises the behavior of the CNN when the input image has multiple objects, as presented in Fig. \ref{fig:grasprobot2}. The network is designed to deal with only one object, which can be seen as a limitation or as a strategy, since the prediction of multiple rectangles needs an additional post-processing step for choosing one of them. The important thing to note is that the presence of two or more objects does not confuse the network as it chooses one of the objects as a target. Figs. \ref{fig:copocaneta} and \ref{fig:canetacopo} show that this choice is made at random. It is important since, although the application space involves only one object, the CNN remains operational if another object enters the camera's field of view.

It should be noted, however, that a cluttered scenario is not allowed in the experiment. According to Morisson et al. \cite{morrison2019learning}, a challenge in this type of environment is that many visual occlusions occur, so that multi-viewpoint, or some other information fusion strategy, must be employed in order to improve accuracy of grasp detection.

\subsubsection{Visual Servoing}
\label{vsonlinesection}

To test the networks in a real visual servoing scenario, and assess their robustness to lighting changes (global or local), their ability to generalize to first-seen objects and to ascertain whether the prediction speed ensures real-time operation, these trained CNNs were implemented to control the Kinova Gen3 robot. As it will be shown, the results differ from the offline results, since the network is tested with feedback. Thus, the graph curve is exponential and subject to real-world noise. The experiment's setup for all models begins at pose [0.276m, 0.390m, 0.411m, $177.734^{\circ}$, $-4.326^{\circ}$, $90.252^{\circ}$], where the reference image is taken, and then the robot is repositioned at pose [0.344m, 0.326m, 0.372m, $-179,089^{\circ}$, $-3,201^{\circ}$, $91,907^{\circ}$], where the control process starts. At each iteration, a new current image $I_c$ is considered by the network. The target object for the visual servoing task is a scientific calculator.

The differences between achieved and desired poses using visual servoing with the developed CNN model are presented in Table \ref{errorm3}, considering $ \lambda_ {lin} = 0.05 $ and $ \lambda_ {ang} = 2 $. The gains chosen are less than those used in the offline experiment, as the network must deal with a noisy scenario. In the final dynamic grasp experiment, the convergence time becomes a relevant factor and is better assessed by increasing $\lambda$.

\begin{table}[!h]
\centering
\caption{Differences between achieved and desired poses using the velocity control predicted by VS CNN models}
\label{errorm3}
\resizebox{\linewidth}{!}{
\begin{tabular}{@{}c|c|c|c|c|c|c@{}}
\hline
\hline
Error & $x \left[mm\right]$ & $y \left[mm\right]$& $z \left[mm\right]$& $\alpha \left[^{\circ}\right]$& $\beta \left[^{\circ}\right]$ & $\gamma \left[^{\circ}\right]$ \\
 \hline
 \textit{Model 1}  & 1.3  & 2.3  & 4  & 0.56  & 0.29  & 0.57  \\
 \textit{Model 2} & 26.2  & 16.2  & 3  & 1.20  & 0.14  & 0.58  \\
\textit{Model 3}  & 1.1  & 8.4  & 31.8  & 0.75  & 0.15  & 1.86  \\
\textit{Model 4} & 44.5  & 3.6  & 31.1  & 0.11  & 0.61  & 0.95  \\
\hline
\hline
\end{tabular}
}
\end{table}

In general, it is possible to note that all networks have more difficulty in inferring the necessary velocity in $z$, to reach the desired pose. This can be seen both in the final positioning error and in the behavior of control signal $ \dot{z}$, more oscillatory than the others.

\begin{figure*}[htb!]
\centering
     \begin{subfigure}[b]{0.35\textwidth}
         \centering
         \includegraphics[width=\textwidth]{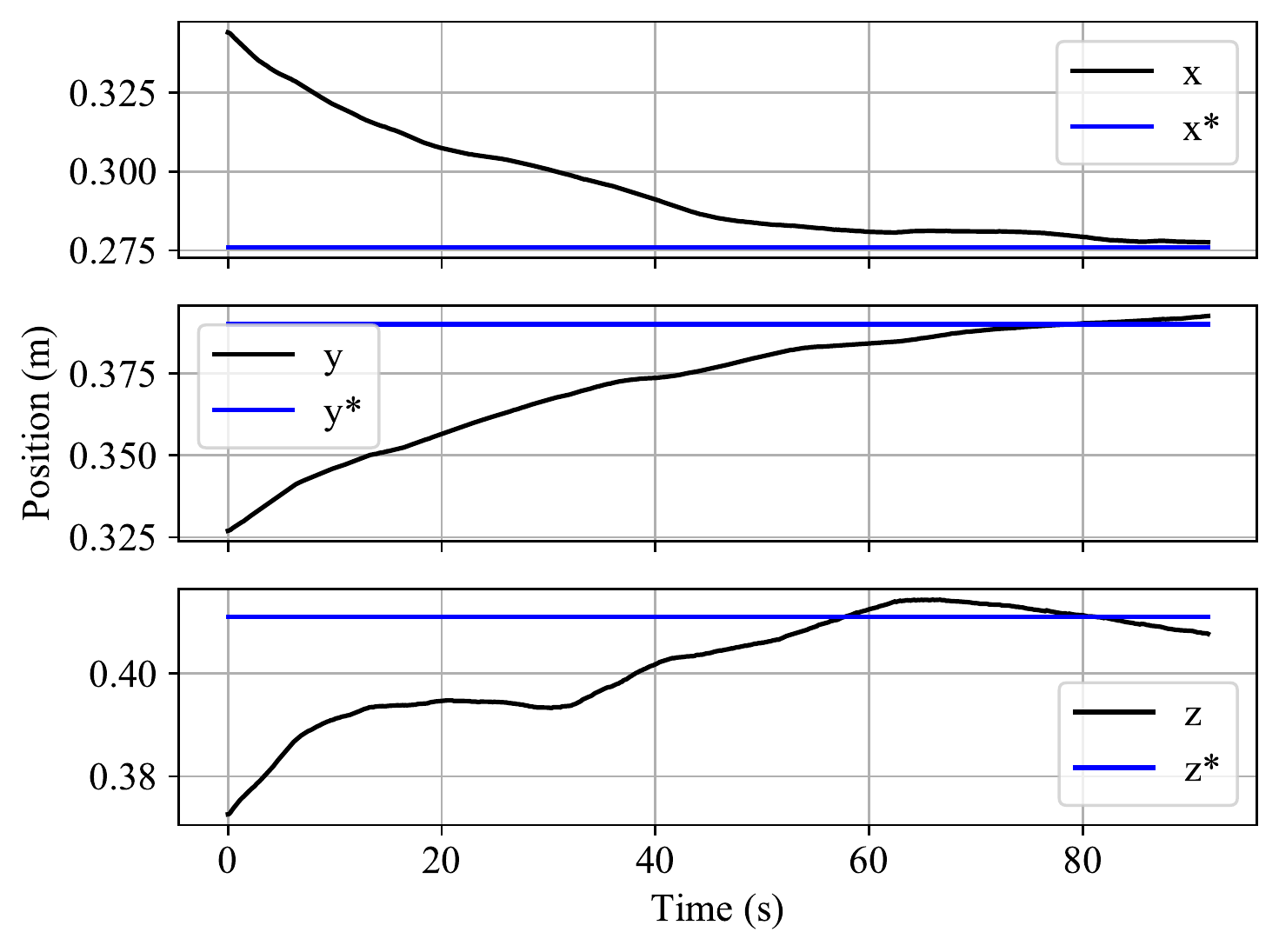}
         \caption{Position in time - Model 1}
         \label{fig:onl11}
     \end{subfigure}
     \begin{subfigure}[b]{0.35\textwidth}
         \centering
         \includegraphics[width=\textwidth]{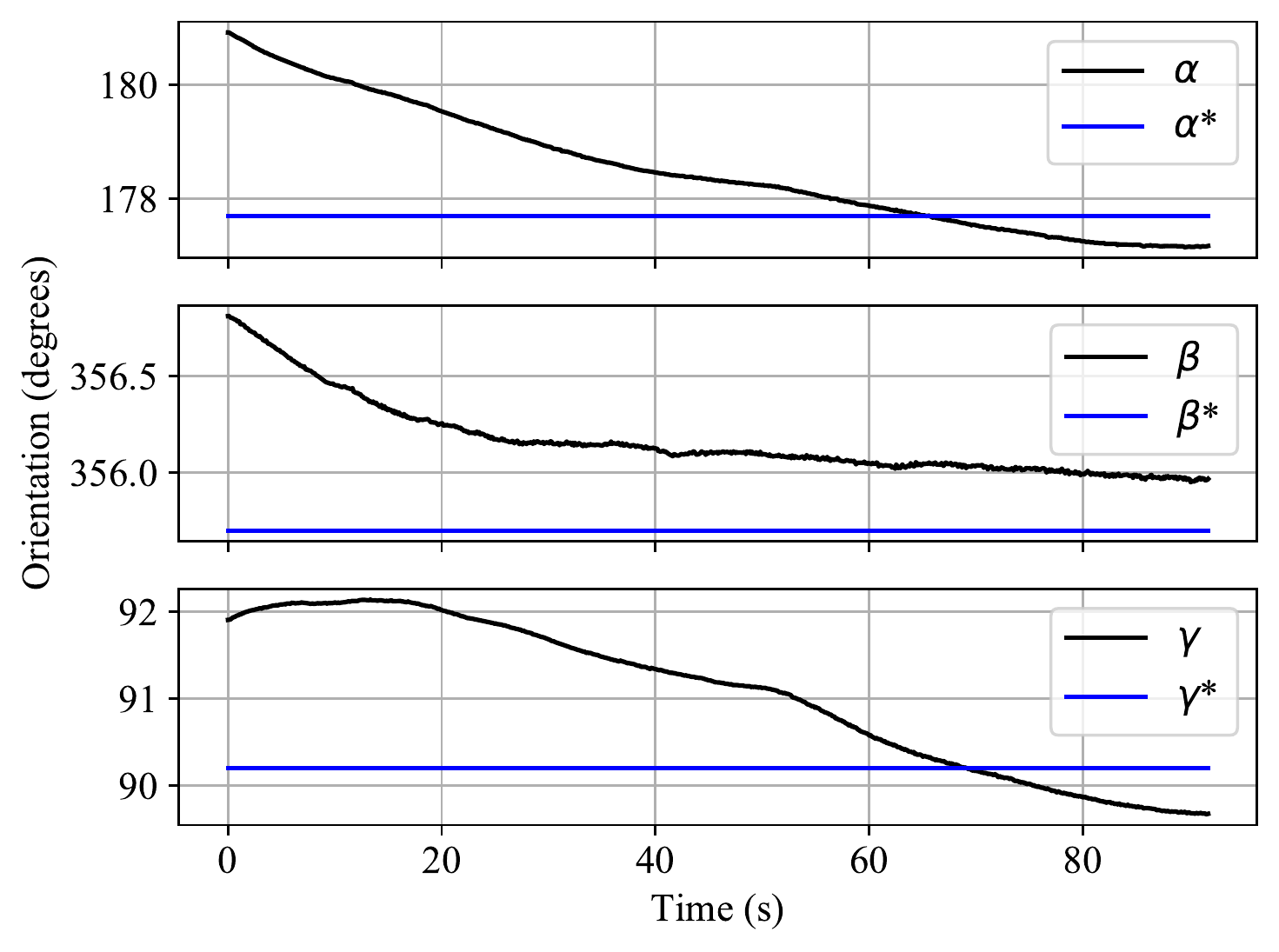}
         \caption{Orientation in time - Model 1}
         \label{fig:onl12}
     \end{subfigure} 
     \hfill
     \begin{subfigure}[b]{0.35\textwidth}
         \centering
         \includegraphics[width=\textwidth]{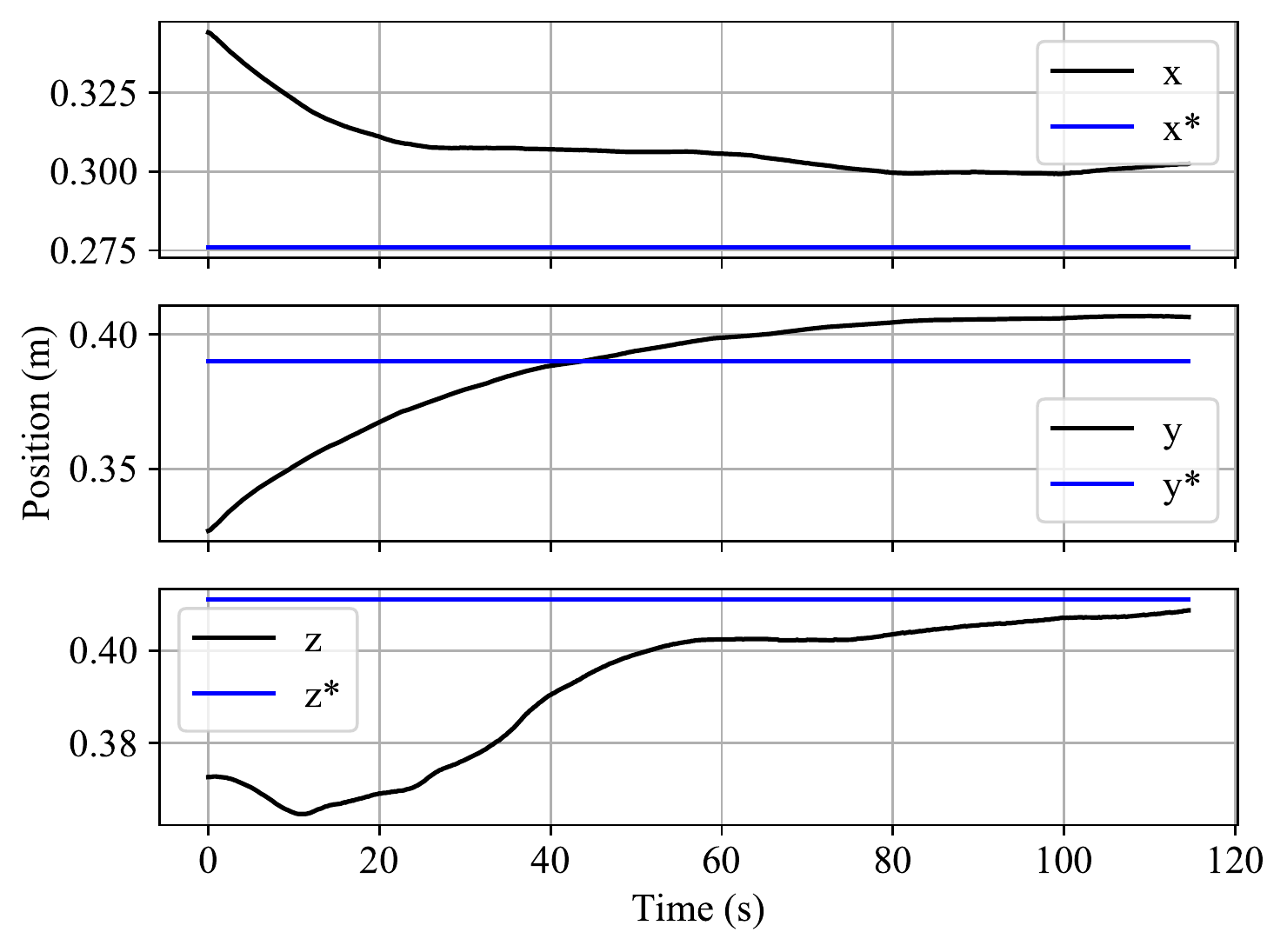}
         \caption{Position in time - Model 2}
         \label{fig:onl21}
     \end{subfigure}
     \begin{subfigure}[b]{0.35\textwidth}
         \centering
         \includegraphics[width=\textwidth]{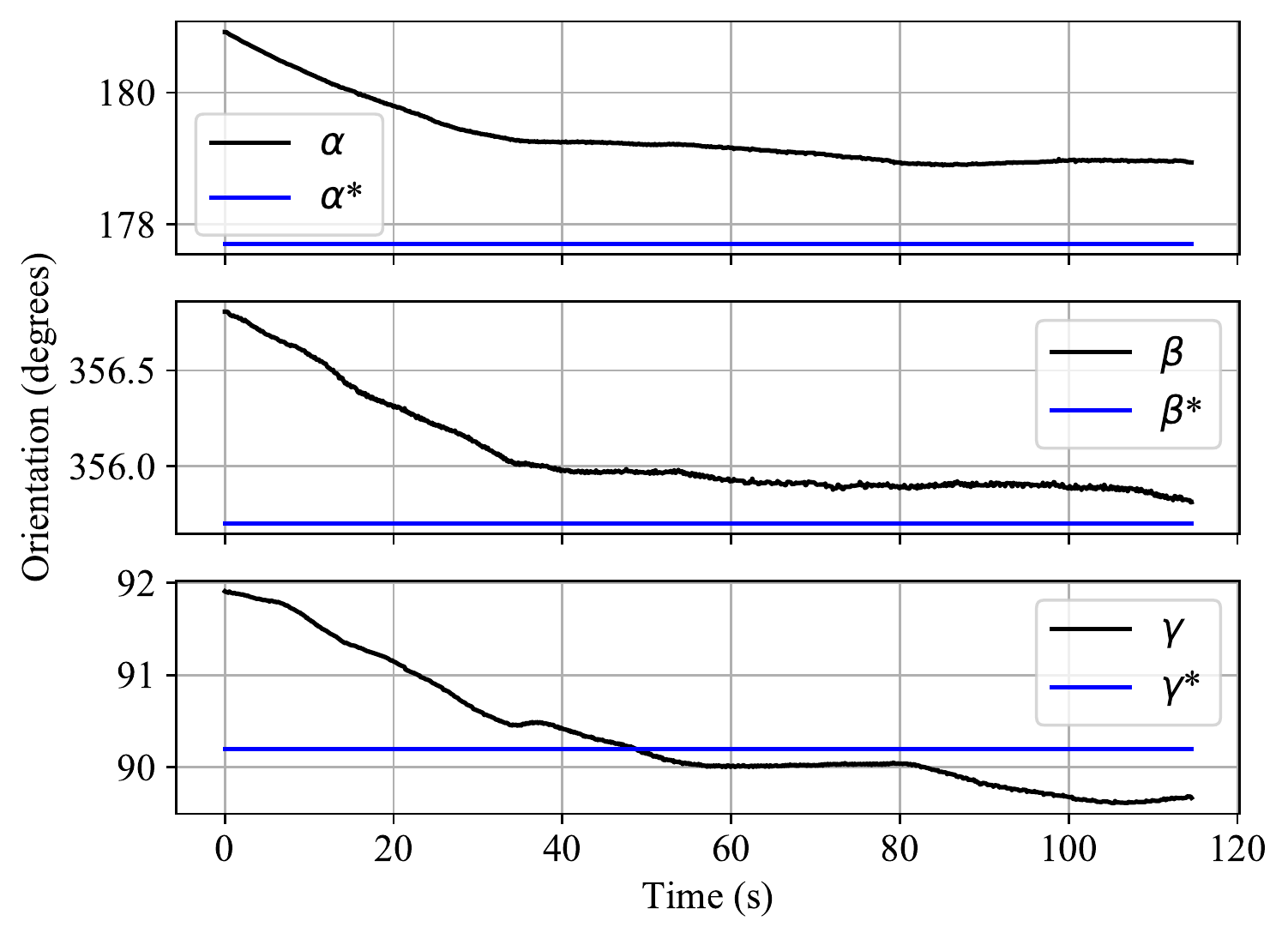}
         \caption{Orientation in time - Model 2}
         \label{fig:onl22}
     \end{subfigure} 
     \hfill
     \begin{subfigure}[b]{0.35\textwidth}
         \centering
         \includegraphics[width=\textwidth]{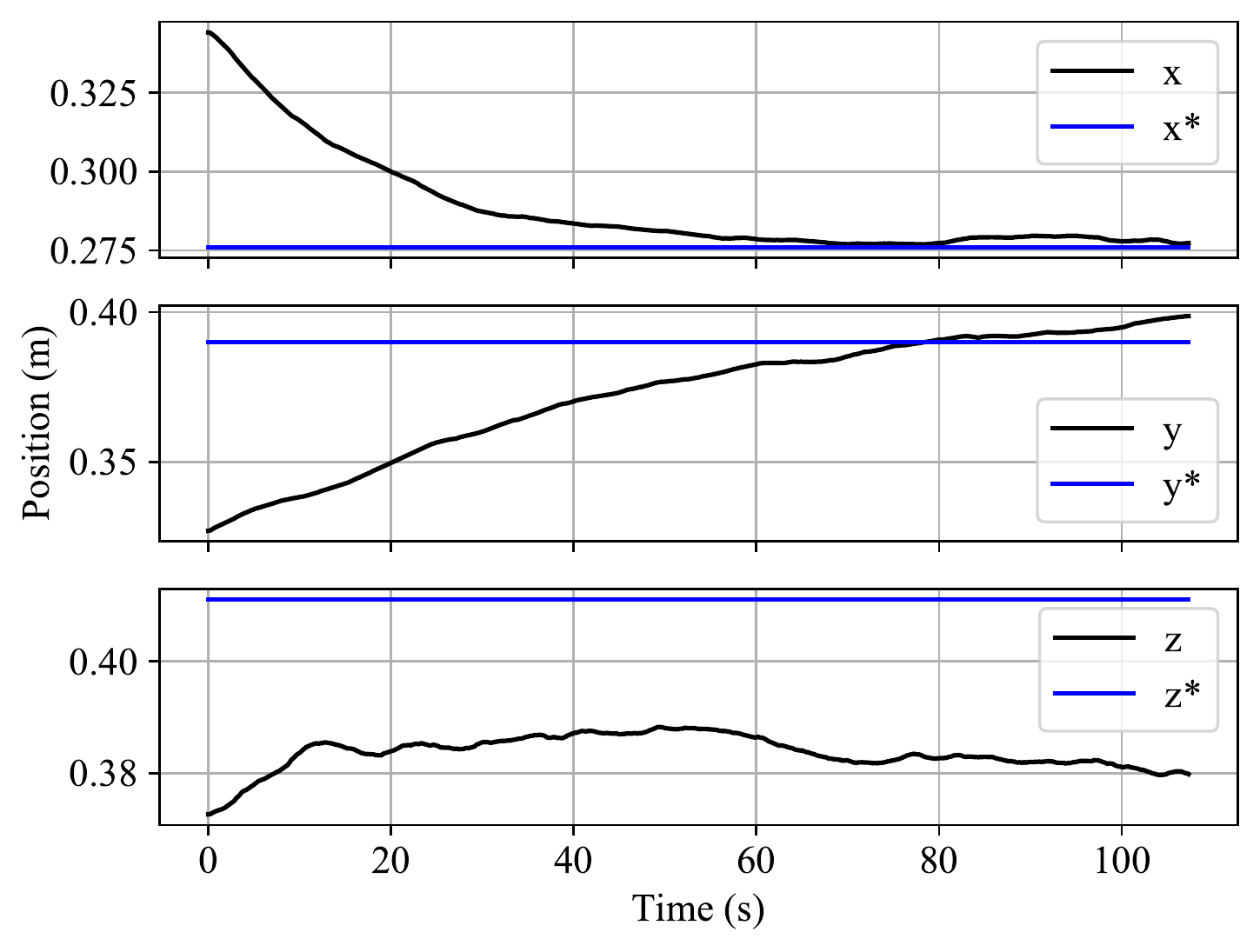}
         \caption{Position in time - Model 3}
         \label{fig:onl31}
     \end{subfigure}
     \begin{subfigure}[b]{0.35\textwidth}
         \centering
         \includegraphics[width=\textwidth]{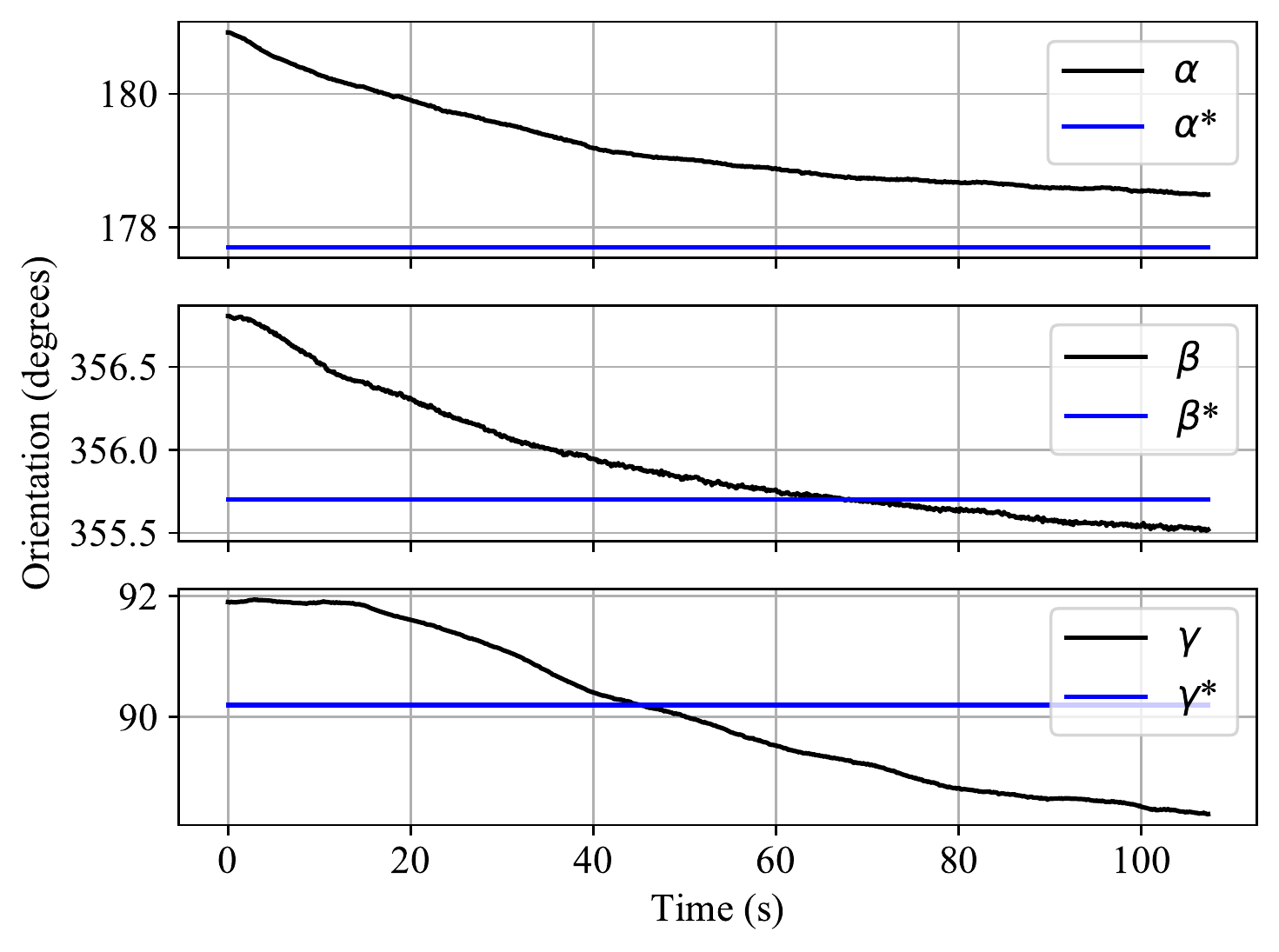}
         \caption{Orientation in time - Model 3}
         \label{fig:onl32}
     \end{subfigure} 
     \hfill
     \begin{subfigure}[b]{0.35\textwidth}
         \centering
         \includegraphics[width=\textwidth]{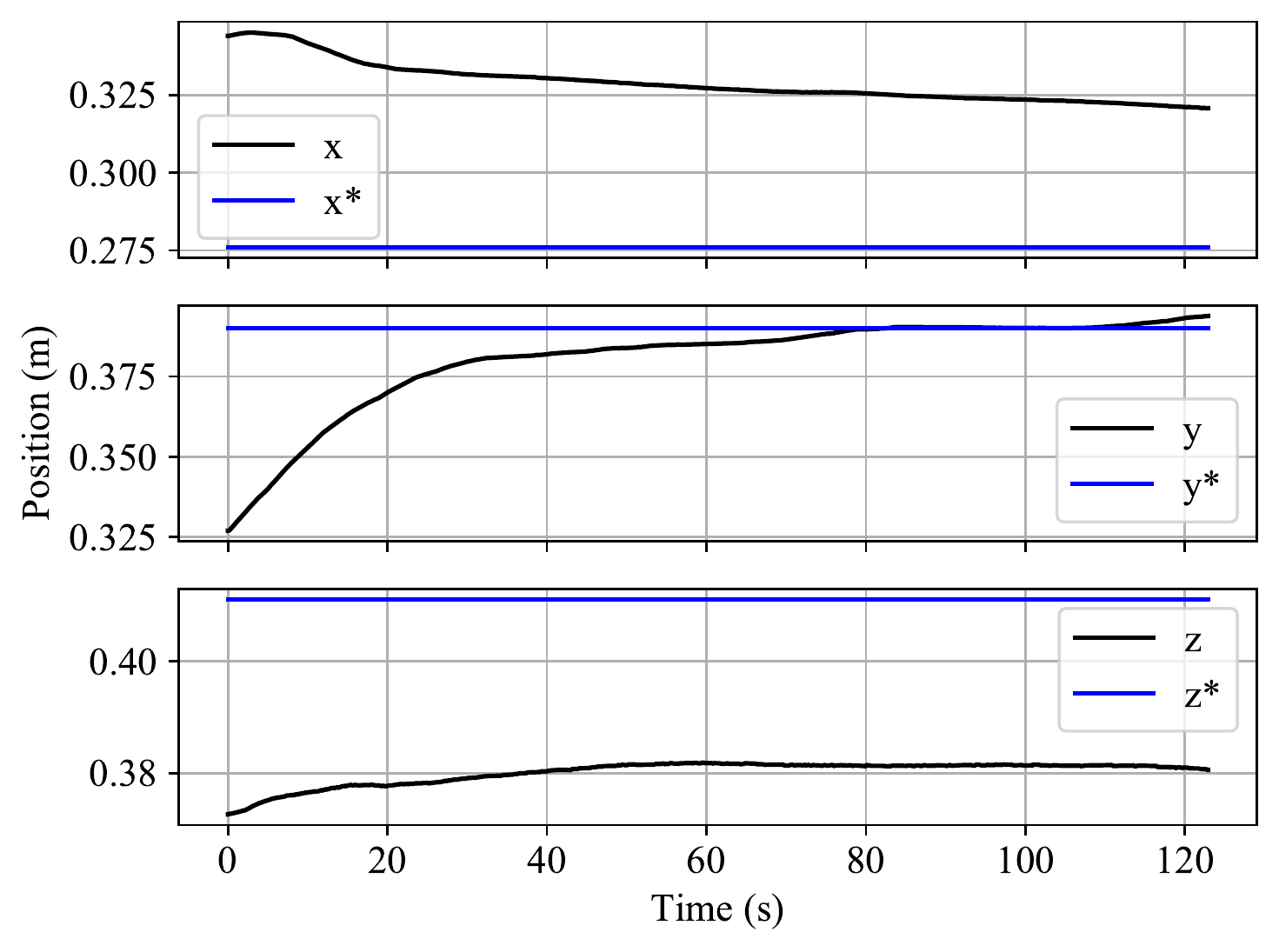}
         \caption{Position in time - Model 4}
         \label{fig:onl41}
     \end{subfigure}
     \begin{subfigure}[b]{0.35\textwidth}
         \centering
         \includegraphics[width=\textwidth]{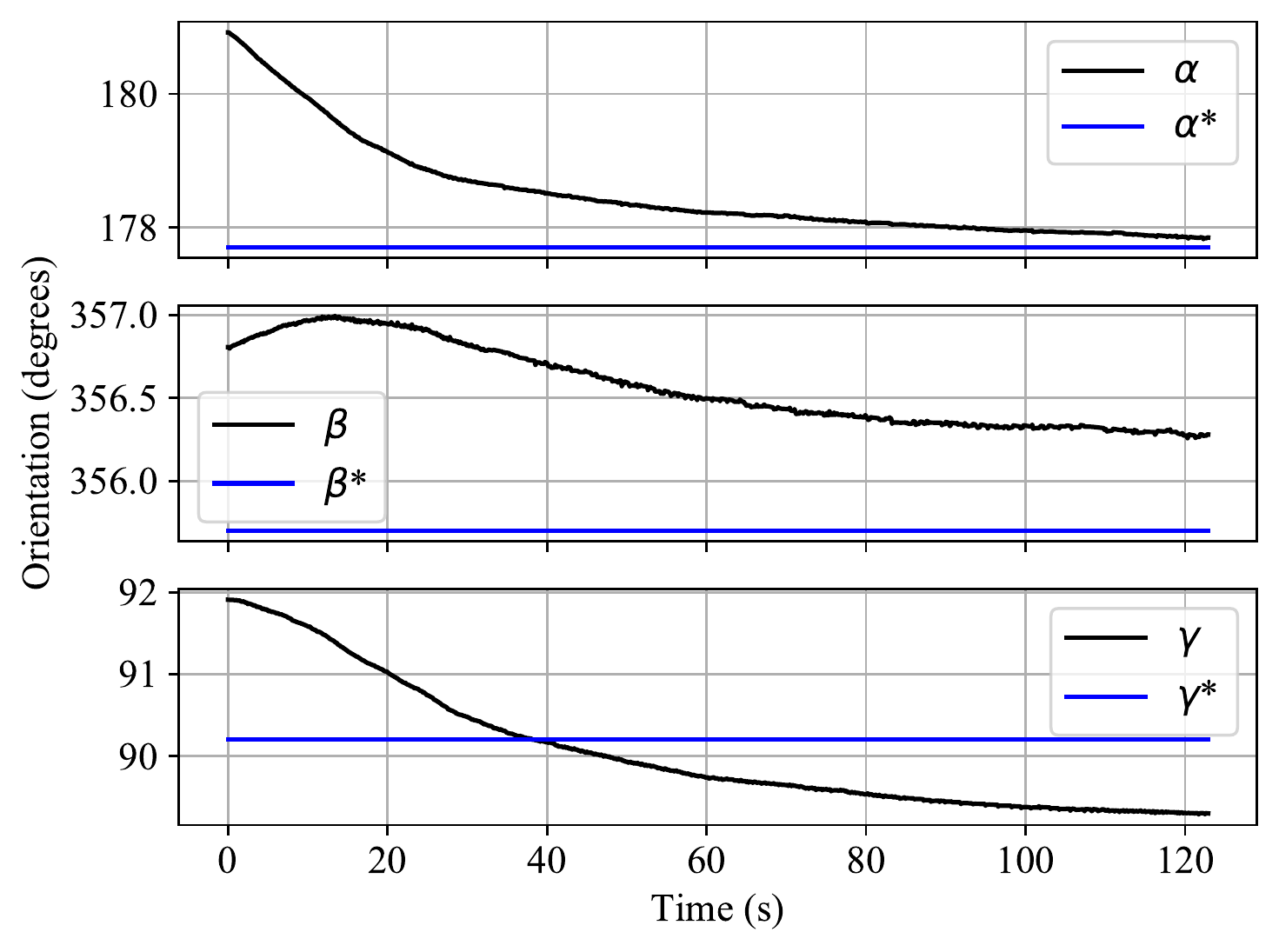}
         \caption{Orientation in time - Model 4}
         \label{fig:onl42}
     \end{subfigure} 
\caption{Robot's position and orientation in time (setpoint in blue) during visual servoing} 
\label{pos_ori}
\end{figure*}

\begin{figure*}[htb!]
\centering
\begin{subfigure}[b]{0.49\textwidth}
         \centering
         \includegraphics[width=\textwidth]{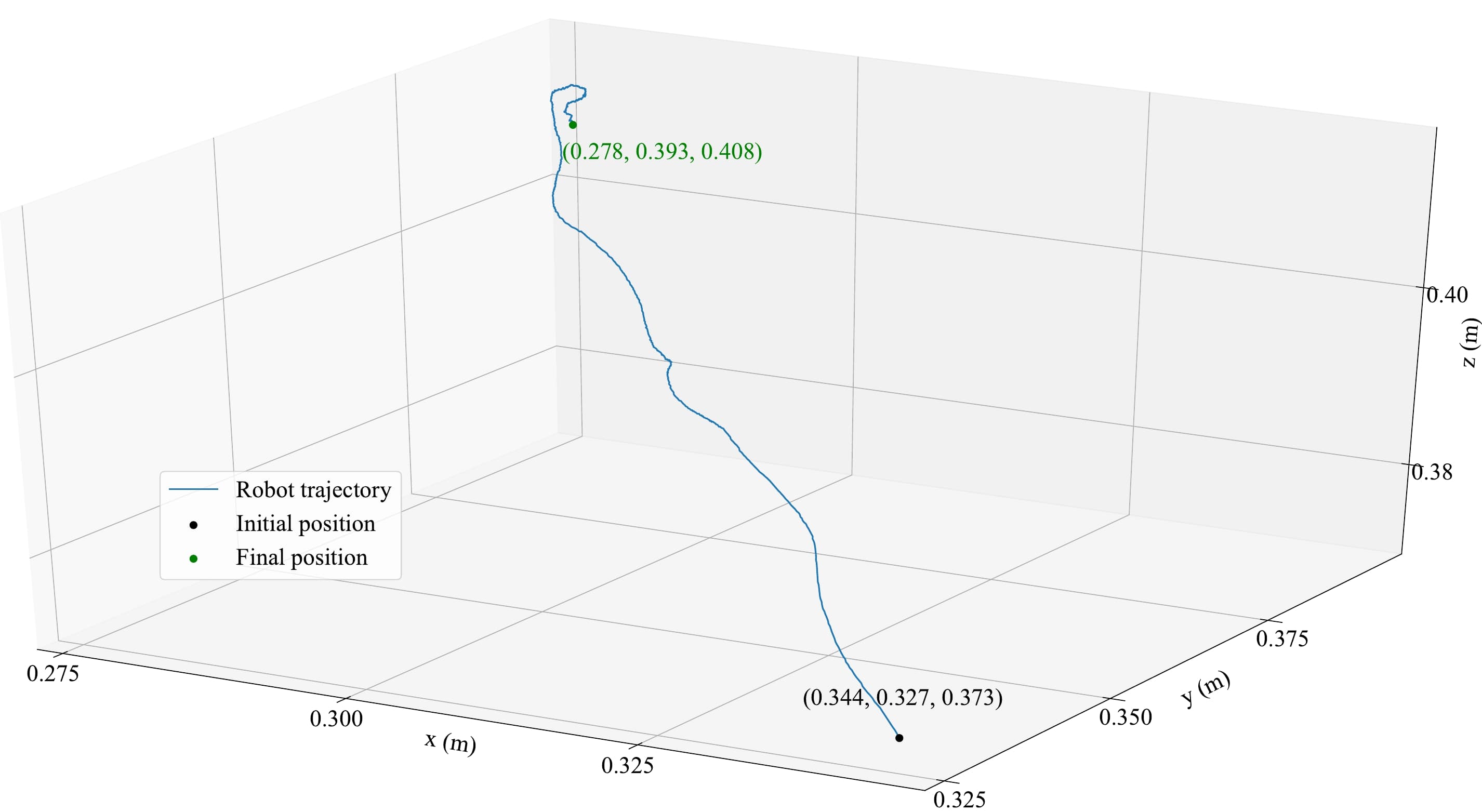}
         \caption{Robot's trajectory during VS with Model 1}
         \label{fig:onl15}
\end{subfigure}
\begin{subfigure}[b]{0.49\textwidth}
         \centering
         \includegraphics[width=\textwidth]{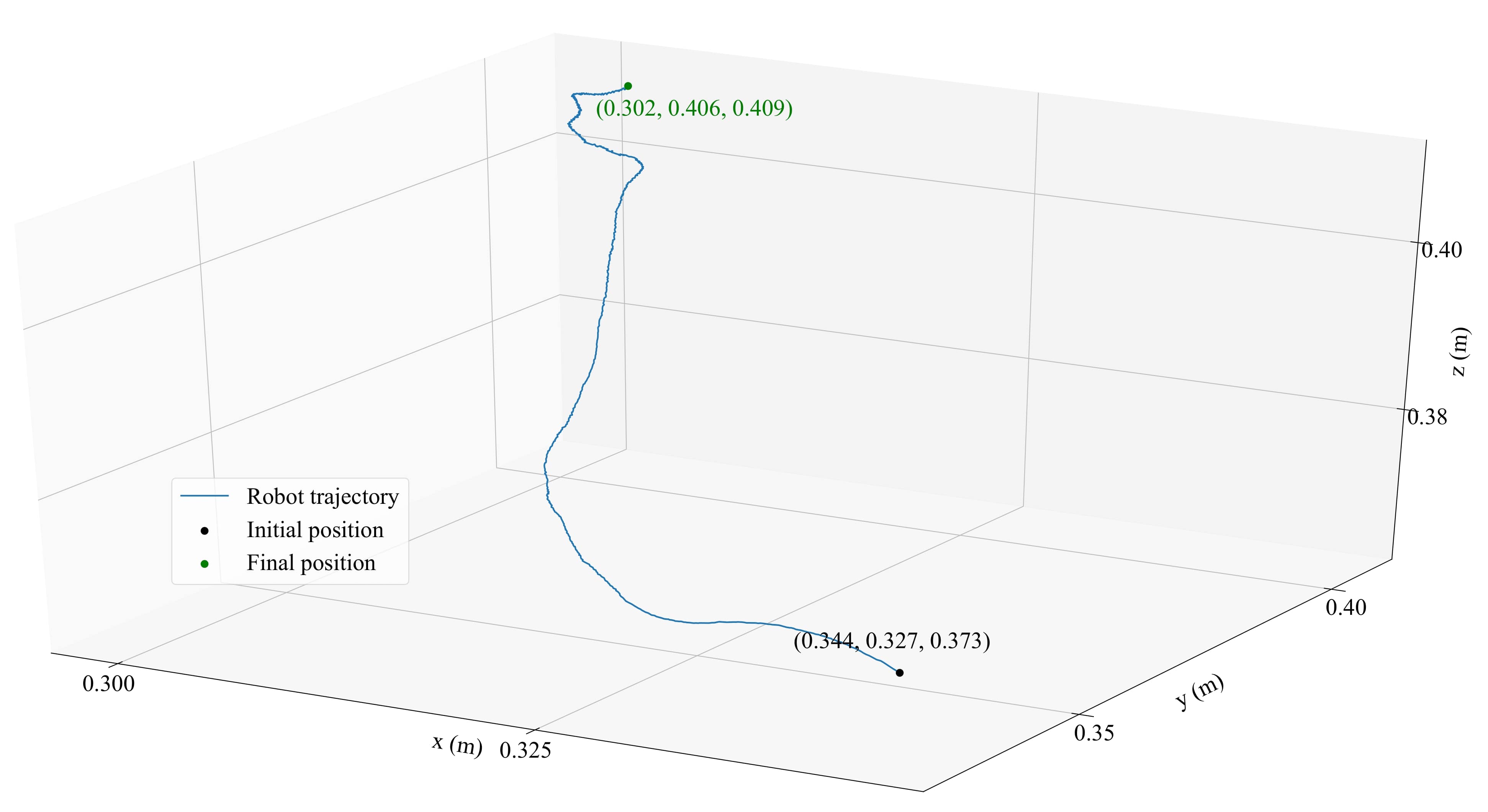}
         \caption{Trajectory performed by the robot with Model 2}
         \label{fig:onl25}
     \end{subfigure} 
\begin{subfigure}[b]{0.49\textwidth}
         \centering
         \includegraphics[width=\textwidth]{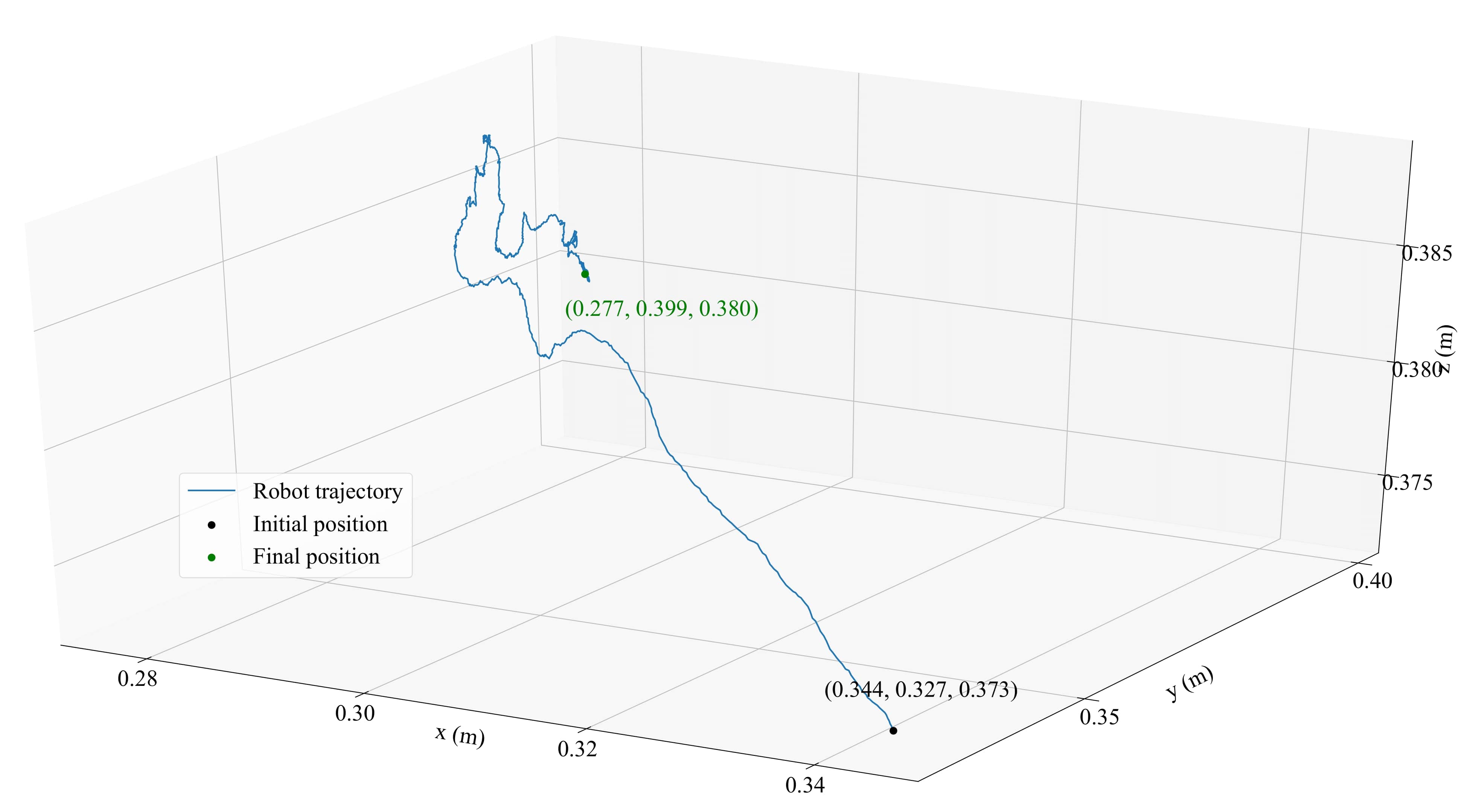}
         \caption{Trajectory performed by the robot with Model 3}
         \label{fig:onl35}
     \end{subfigure} 
\begin{subfigure}[b]{0.49\textwidth}
         \centering
         \includegraphics[width=\textwidth]{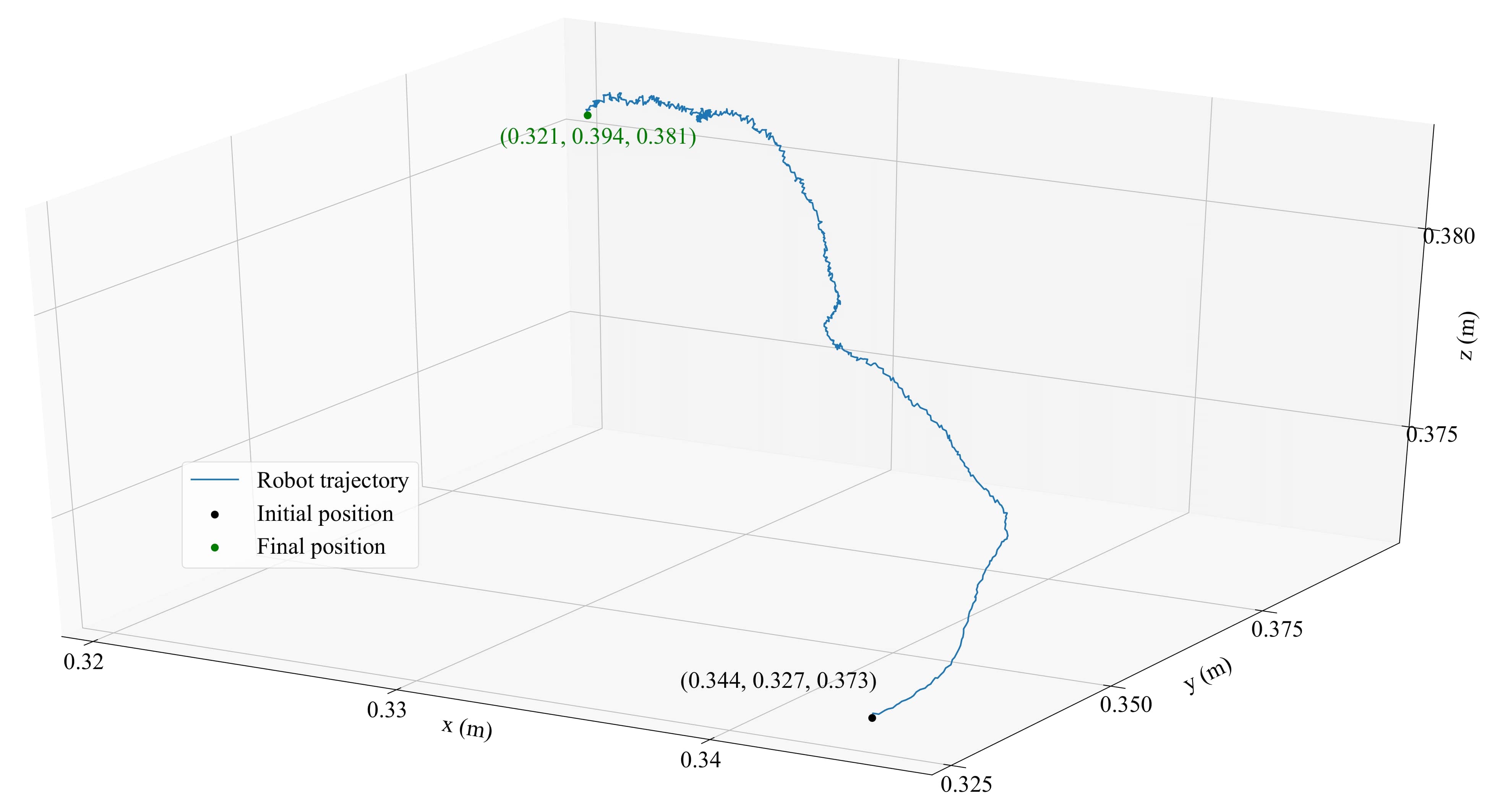}
         \caption{Trajectory performed by the robot with Model 4}
         \label{fig:onl45}
     \end{subfigure} 
\caption{Trajectory performed by the robot during visual servoing} 
\label{fig:onl4}
\end{figure*}

The efficiency of the CNNs must be analyzed assuming that the target object was not an instance of training (it was considered only among other items not as a single object). Thus, the final errors of positioning and orientation express the network's ability to generalize a task that, in the literature, is usually performed relying on strong modeling and \textit{a priori} knowledge of the object and camera parameters. Then, a VS controller designed using prior knowledge will certainly outperform one learned from scratch \cite{lampe2013acquiring}. We aim to show that this VS CNN can be applied in the field of autonomous manipulation in situations where knowledge about the process is scarce or difficult to generate.

\begin{figure*}[htb!]
\centering
\begin{subfigure}[b]{0.35\textwidth}
         \centering
         \includegraphics[width=\textwidth]{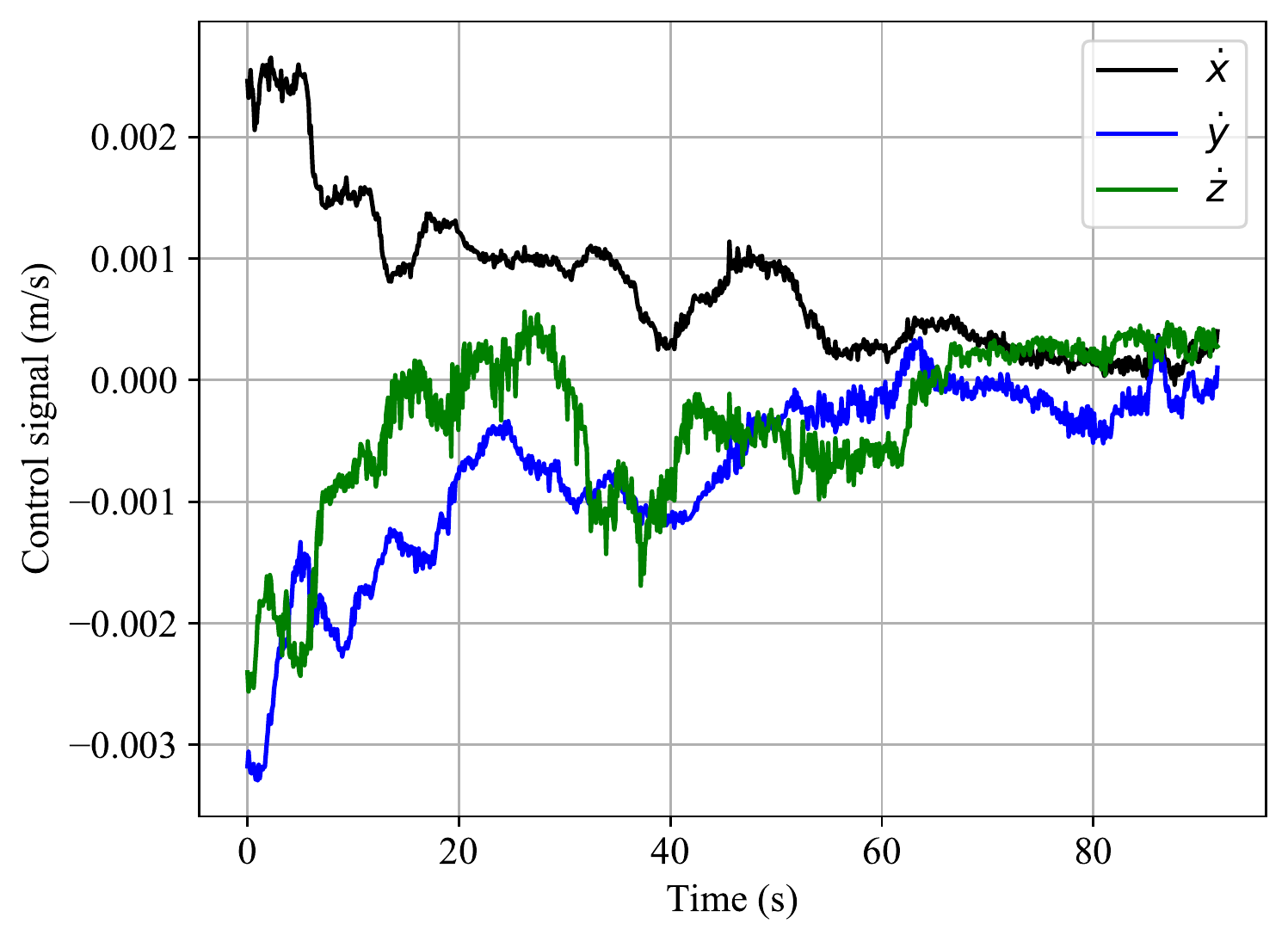}
         \caption{Linear velocity for Model 1}
         \label{fig:onl13}
     \end{subfigure}
     \begin{subfigure}[b]{0.35\textwidth}
         \centering
         \includegraphics[width=\textwidth]{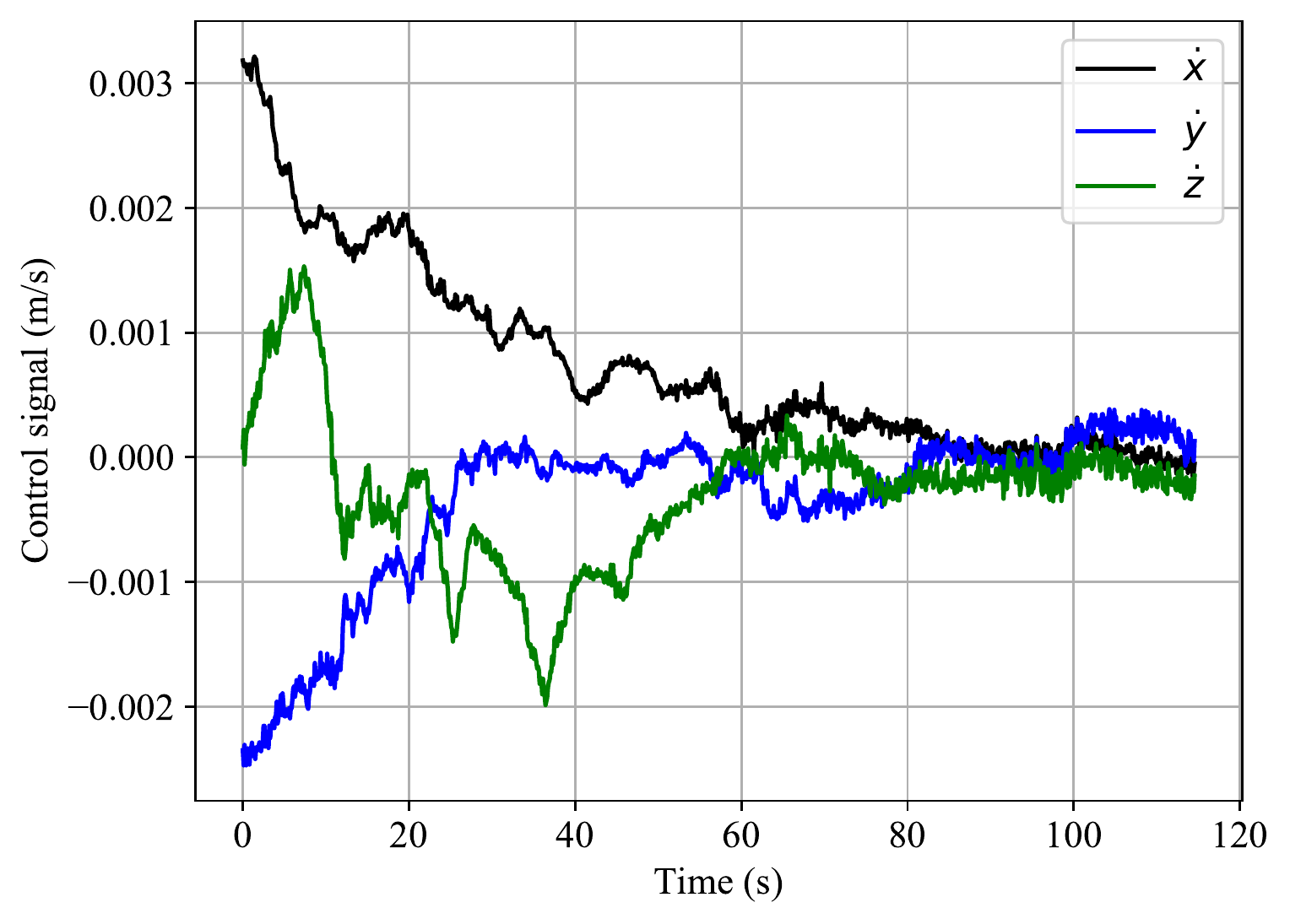}
         \caption{Linear velocity for Model 2}
         \label{fig:onl23}
     \end{subfigure}
     \begin{subfigure}[b]{0.35\textwidth}
         \centering
         \includegraphics[width=\textwidth]{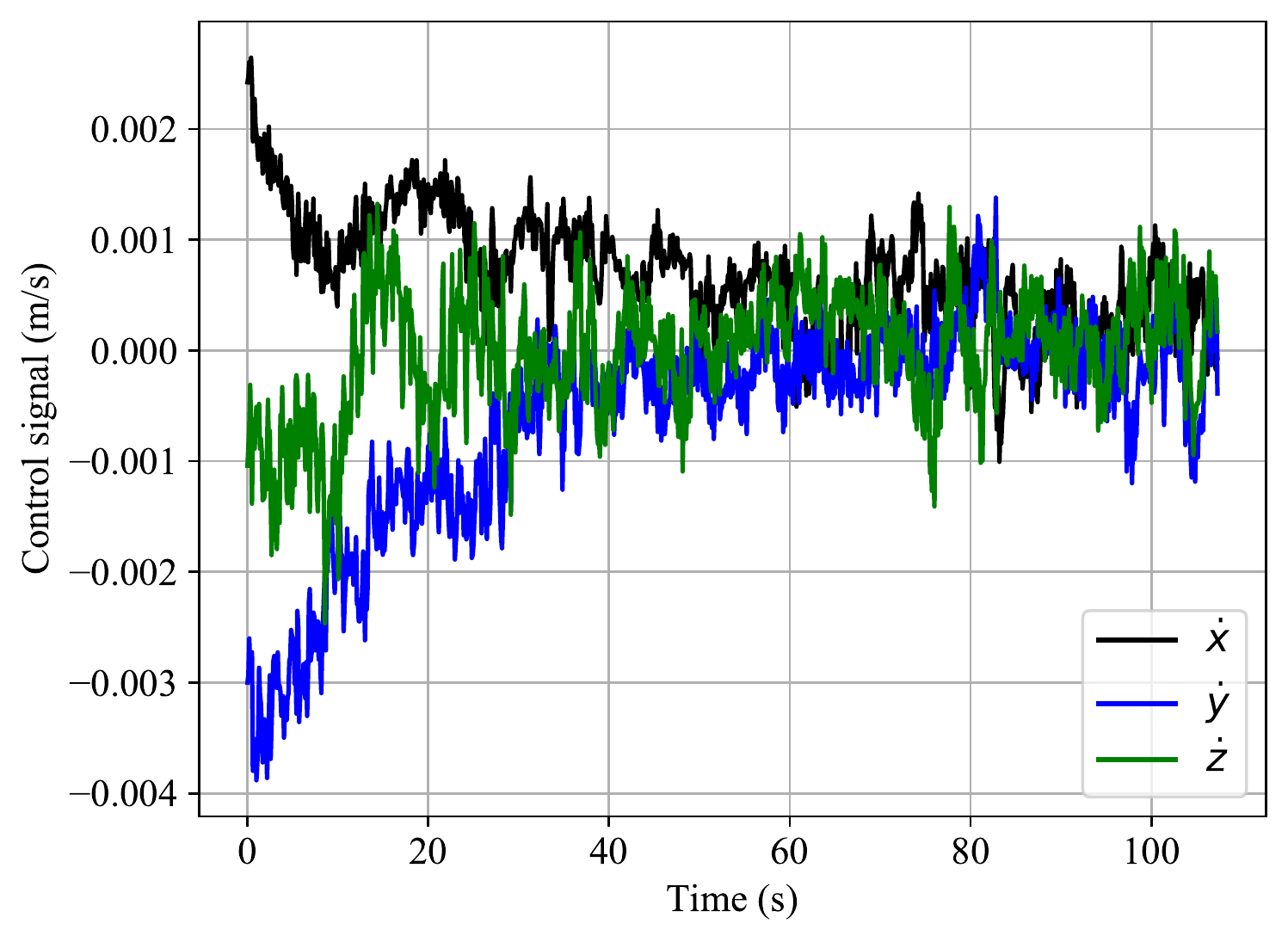}
         \caption{Linear velocity for Model 3}
         \label{fig:onl33}
     \end{subfigure}
     \begin{subfigure}[b]{0.35\textwidth}
         \centering
         \includegraphics[width=\textwidth]{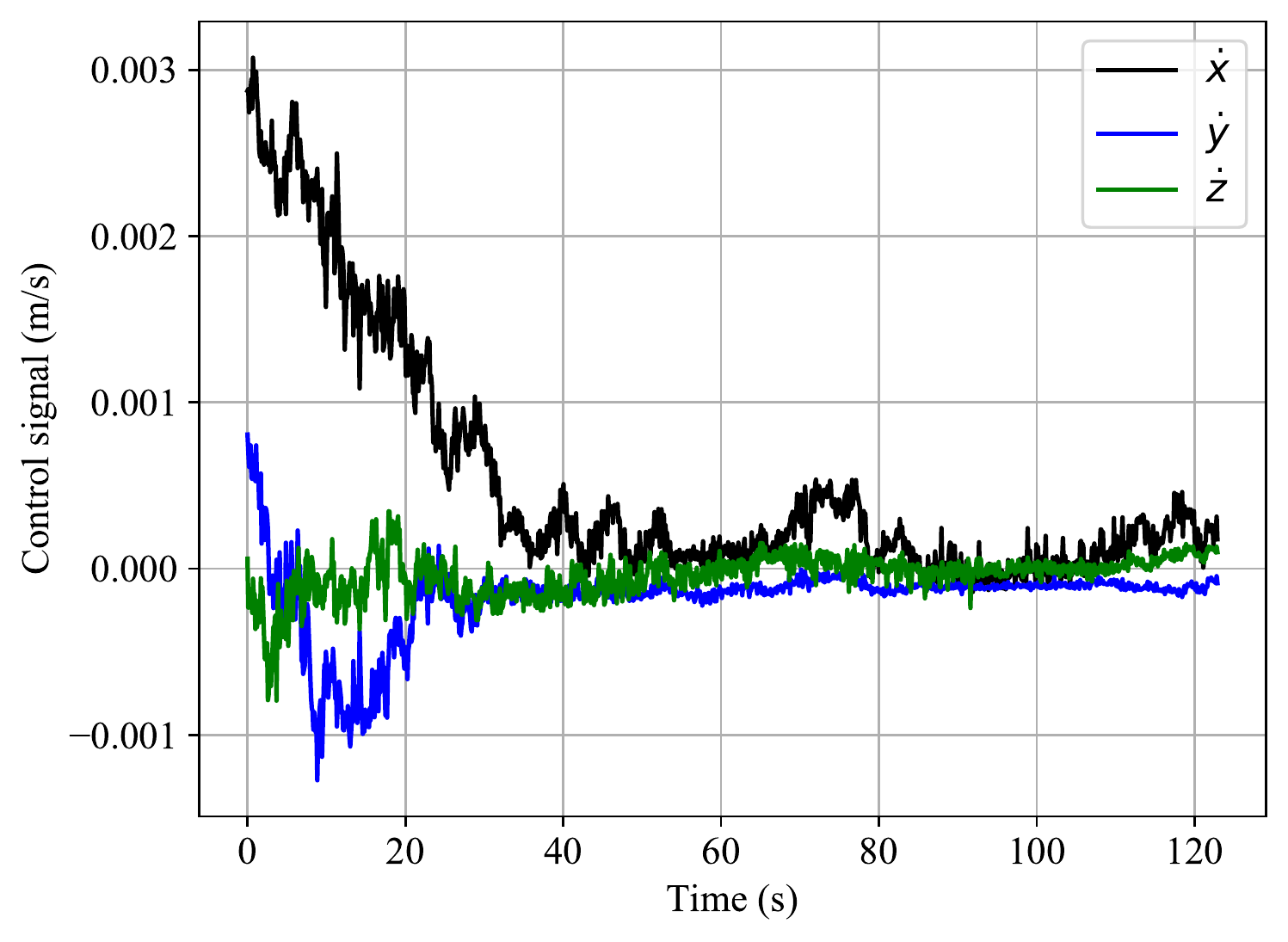}
         \caption{Linear velocity for Model 4}
         \label{fig:onl43}
     \end{subfigure}
     \begin{subfigure}[b]{0.35\textwidth}
         \centering
         \includegraphics[width=\textwidth]{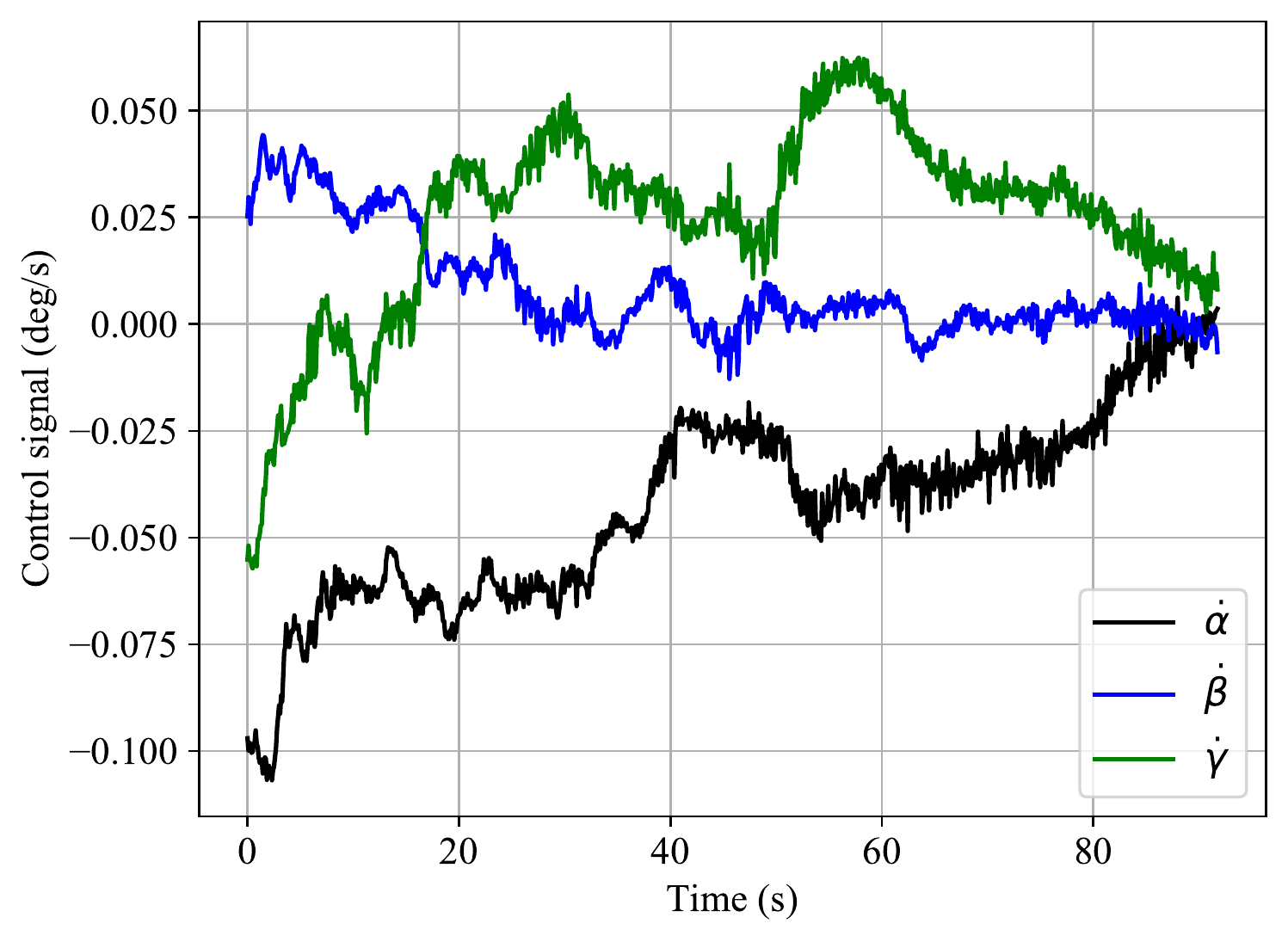}
         \caption{Angular velocity for Model 1}
         \label{fig:onl14}
     \end{subfigure} 
     \begin{subfigure}[b]{0.35\textwidth}
         \centering
         \includegraphics[width=\textwidth]{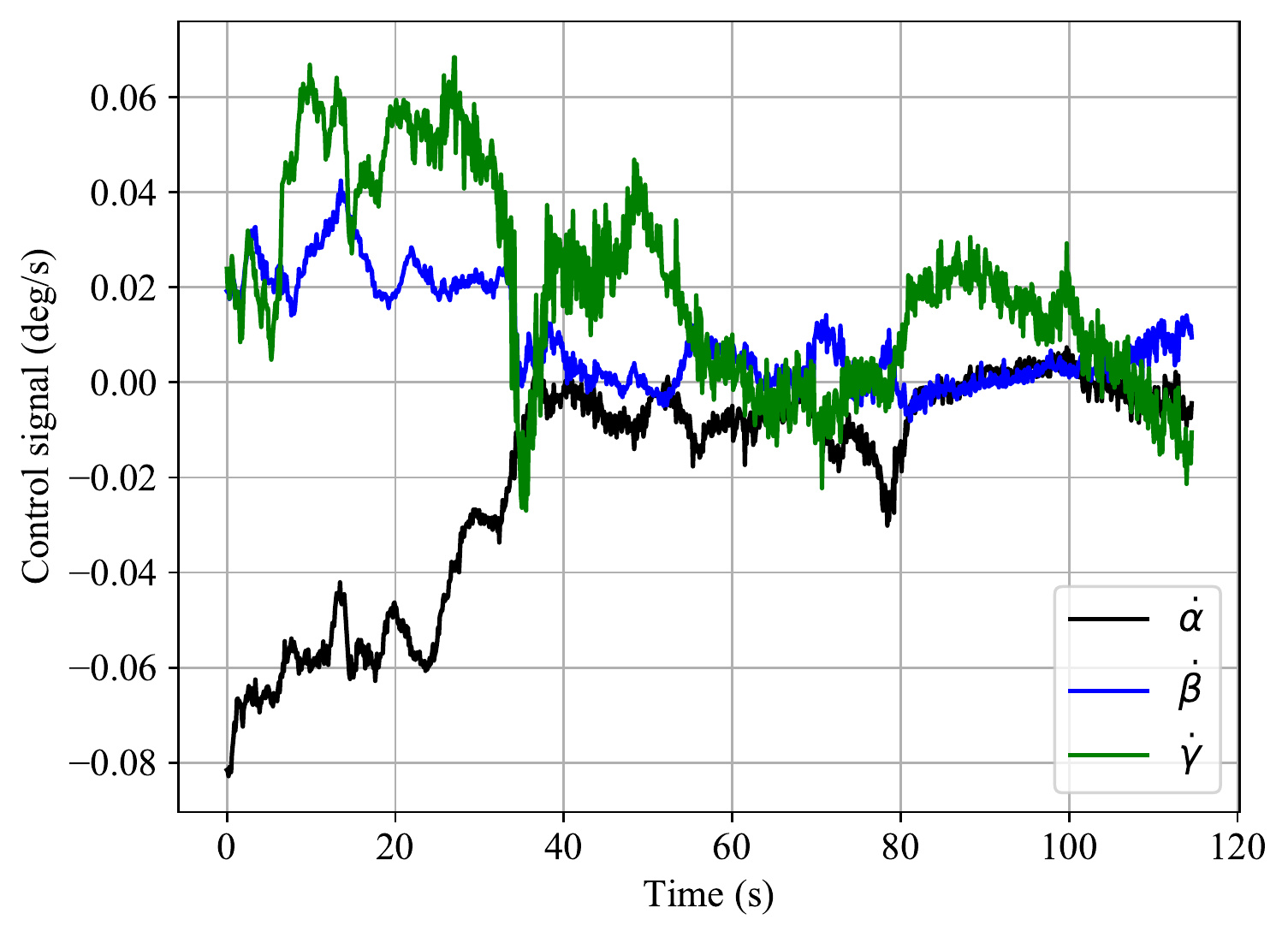}
         \caption{Angular velocity for Model 2}
         \label{fig:onl24}
     \end{subfigure} 
     \begin{subfigure}[b]{0.35\textwidth}
         \centering
         \includegraphics[width=\textwidth]{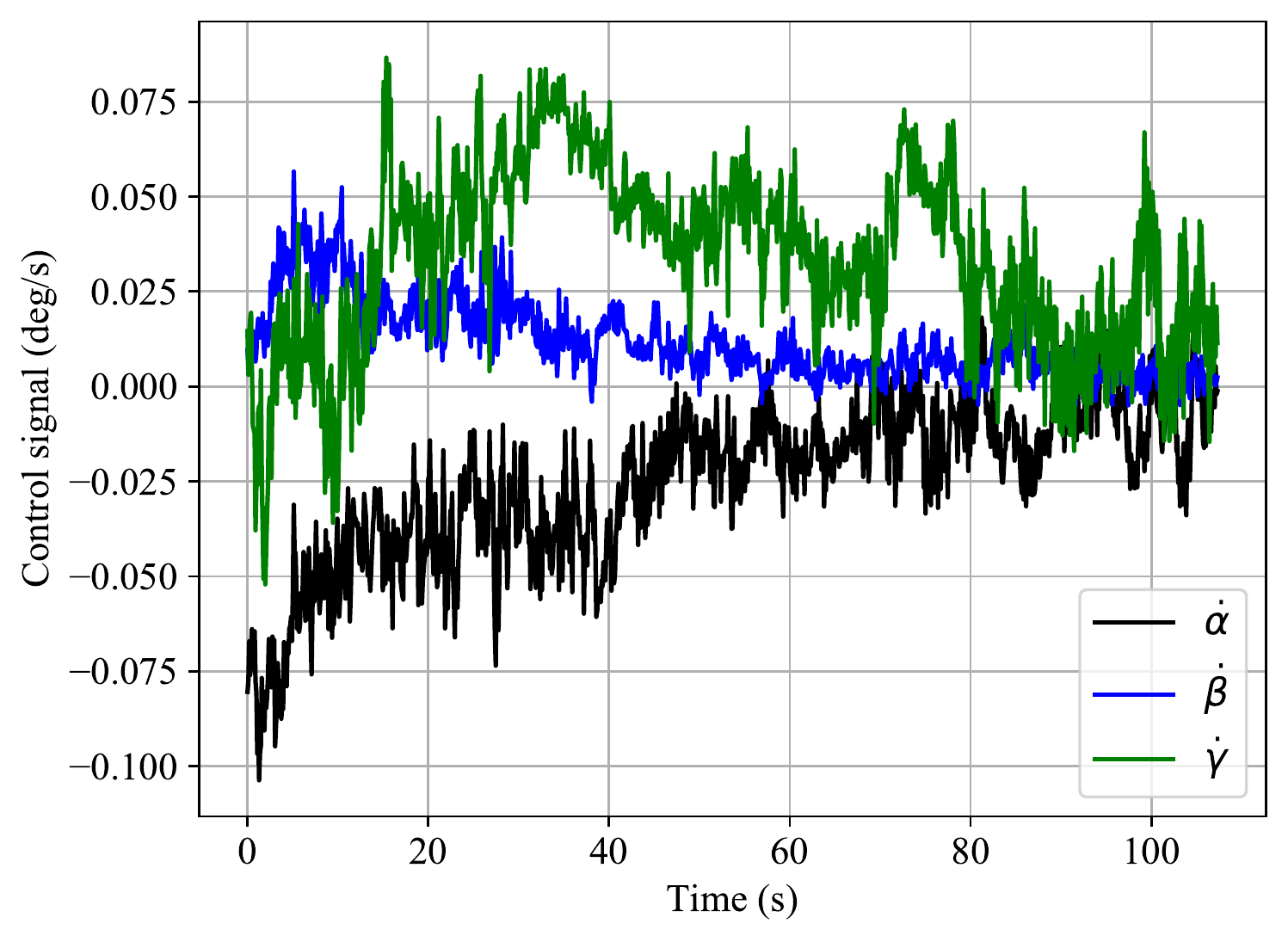}
         \caption{Angular velocity for Model 3}
         \label{fig:onl34}
     \end{subfigure} 
     \begin{subfigure}[b]{0.35\textwidth}
         \centering
         \includegraphics[width=\textwidth]{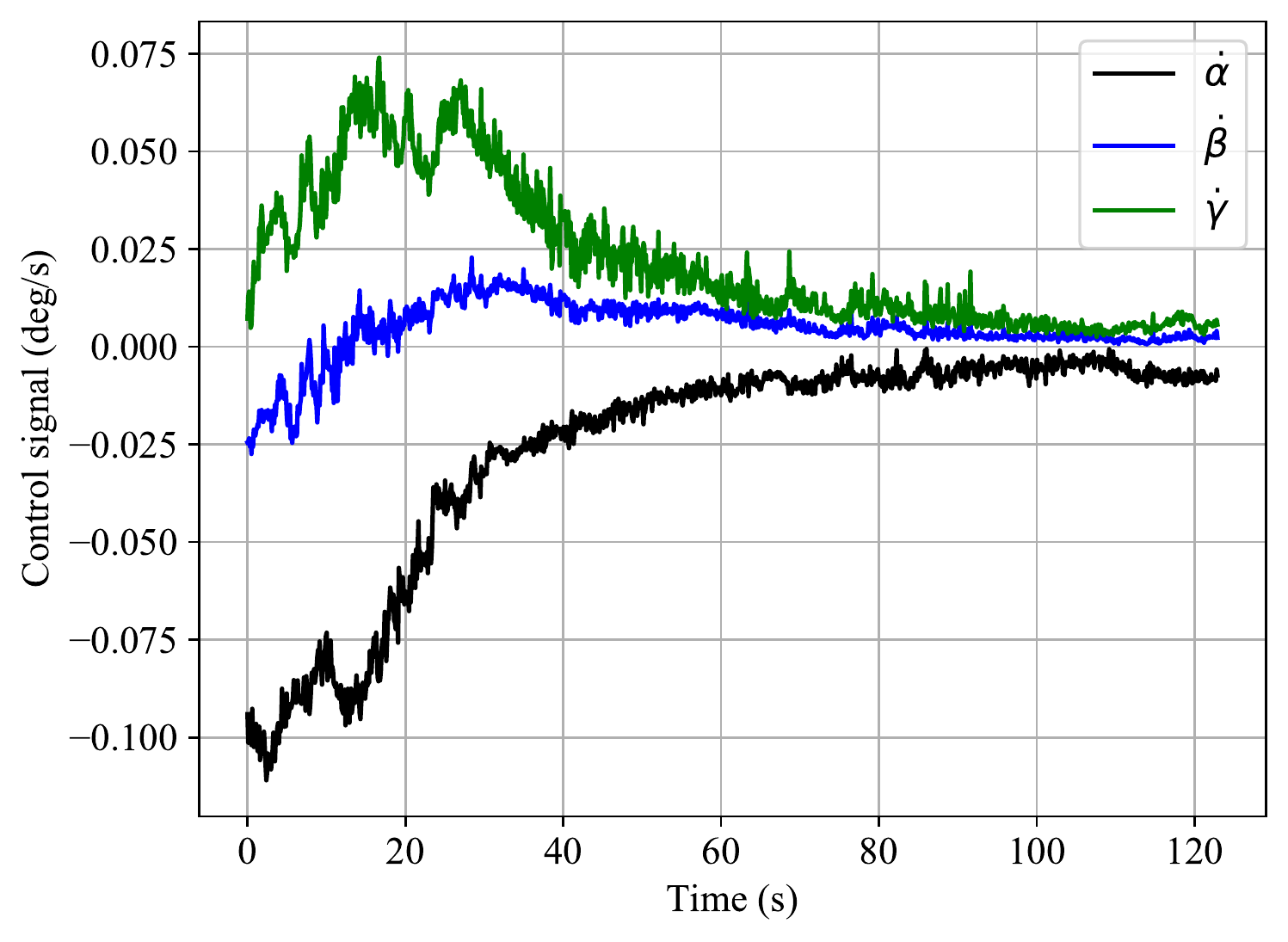}
         \caption{Angular velocity for Model 4}
         \label{fig:onl44}
     \end{subfigure}
\caption{Control signals} 
\label{fig:onl8}
\end{figure*}

However, our designed networks are still able to obtain final millimeter positioning and less than one degree orientation errors. Especially Model 1 (Figs. \ref{fig:onl11} and \ref{fig:onl12}), which is simpler and faster, surprisingly manages to obtain a final positioning error of less than 5mm. All other networks can achieve this same performance in only one dimension, while others reach the order of centimeters.

Regarding the trajectory made by the robot during the control, Model 1 is also the one that has the most desired behavior (Fig. \ref{fig:onl15}). This is the closest to a pure straight line in 3D space, something obtained in classic PBVS, using $ \boldsymbol{s} = (^ {c ^ *} \boldsymbol{t} _c, \theta \boldsymbol{u})$, only when the parameters involved in the control law are perfectly estimated.

Graphs in Figs. \ref{pos_ori} and \ref{fig:onl4} present the trajectory, position and orientation in time for all other models. Model 2, which obtained the lowest MSE in the offline phase, ends the control with a relatively high error in positioning, however, it is the one that converges better in $z$. Regarding the orientation, Model 2 leads to the faster stabilization of the desired angles, something that can be attributed to the dissociation of the network's outputs between linear and angular control signals.

Model 3 converges well for $x$ and $y$, but has a large final error in $z$, whereas Model 4, considered a better version of Model 3, converges well only in $y$. However, the behavior of the control signal, presented in Fig. \ref{fig:onl8}, generated by Model 4, is the closest to a default controller, in which speeds tend to zero quickly and no longer oscillate after that. Model 3, on the other hand, generates an extremely noisy control signal, which penalizes the trajectory made by the robot.

In comparison, Bateux et al. \cite{bateux2018training} also presented results for visual servoing in a real robot, using a VGG network \cite{simonyan2014very}, considering scenes not seen in the training. The authors mention that the network is able to position the robot within a few centimeters of the desired pose. At this point, they switch the control to a classic direct visual servoing in order to reduce the positioning error. Our Model 1, in contrast, achieves millimeter positioning accuracy without any additional control steps, with almost 90 times fewer parameters than VGG.

\subsubsection{Dynamic Grasping}

The last experiment carried out on the robot is the test of the entire system, as illustrated in Fig. \ref{fig:dynamic_grasp_system}. To do this, the robot takes a reference image of an anti-stress ball and then performs a displacement $[\Delta x, \Delta y, \Delta z, \Delta \alpha, \Delta \beta, \Delta \gamma] = [0.15m, -0.15m, 0.05m, 5^{\circ}, -5^{\circ}, 5^{\circ}]$, from where the control starts. To mimic a dynamic object that changes its desired position after the grasping attempt begins, it was preferred to change the pose of the robot, which has the same effect and allows for better monitoring and reproducibility.

The specific ball was not used in the grasping dataset, or in the control dataset. Therefore, the experiment evaluates the adaptation of the robot to a dynamic first-seen object using a method entirely based on learning. Only Model 1 is evaluated since this is the one that achieves the best results in the robot.

\begin{figure}[t]
\centering
     \begin{subfigure}[b]{\linewidth}
         \centering
         \includegraphics[width=\linewidth]{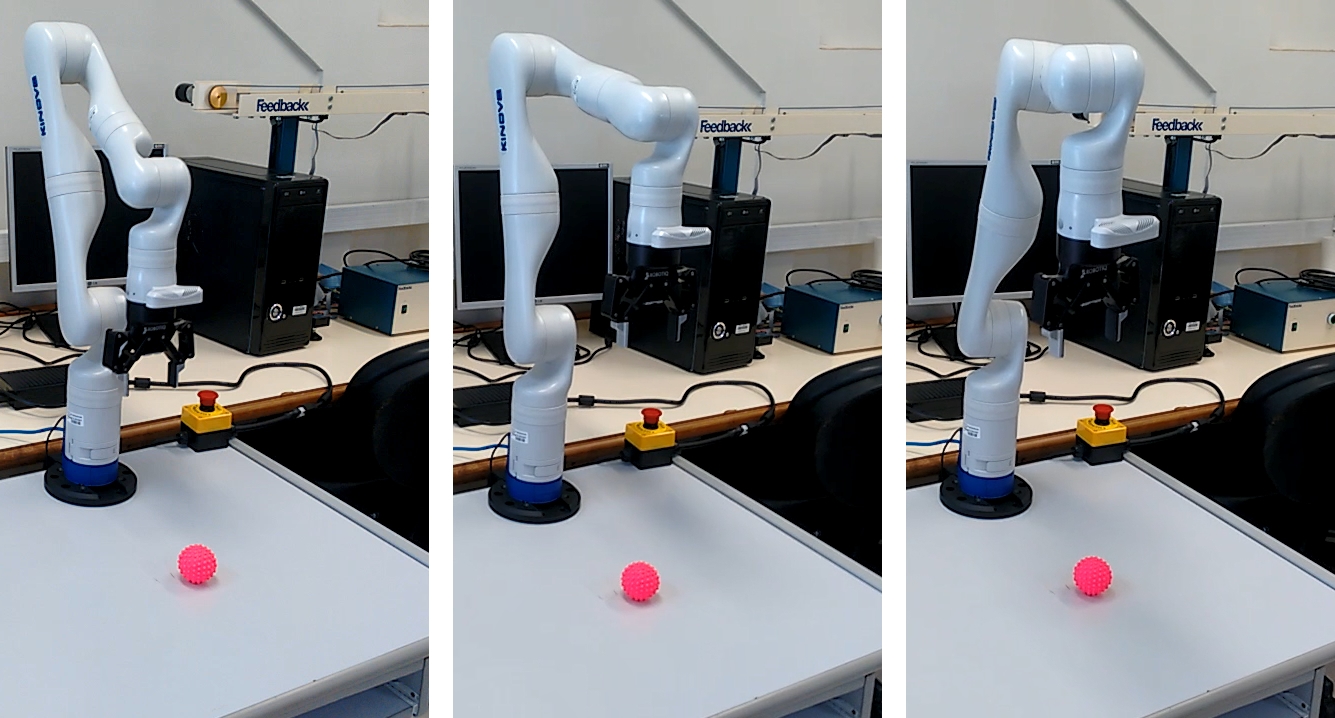}
         \caption{Initial setup for VS}
         \label{fig:din21}
     \end{subfigure}
     \hfill
     \begin{subfigure}[b]{\linewidth}
         \centering
         \includegraphics[width=\linewidth]{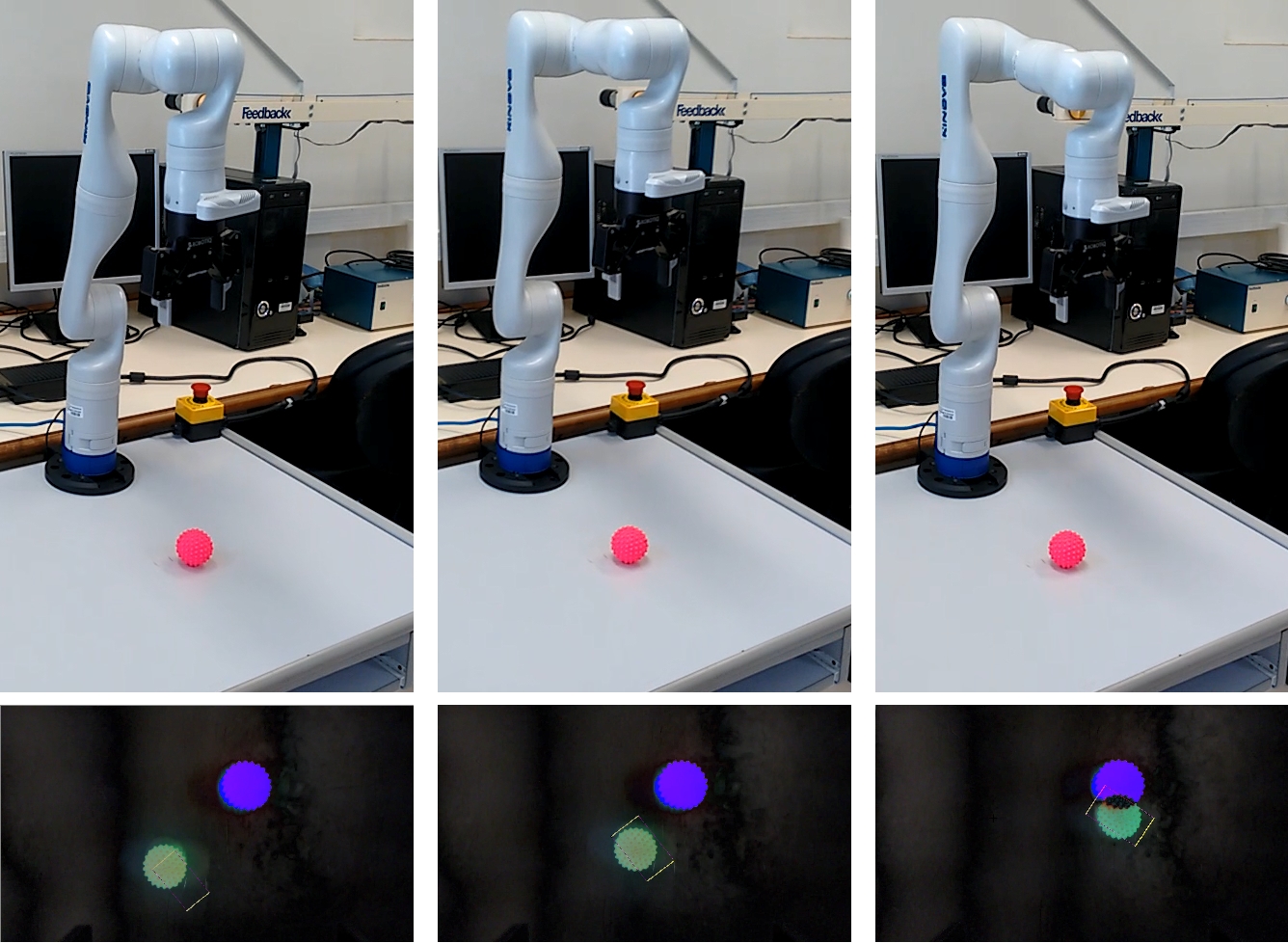}
         \caption{VS execution both in Cartesian and image spaces}
         \label{fig:din22}
     \end{subfigure} 
     \caption{Demonstration of the VS stage in the dynamic grasping} 
\label{fig:din2}
\end{figure}

\begin{figure}[]
\centering
\includegraphics[width=\linewidth]{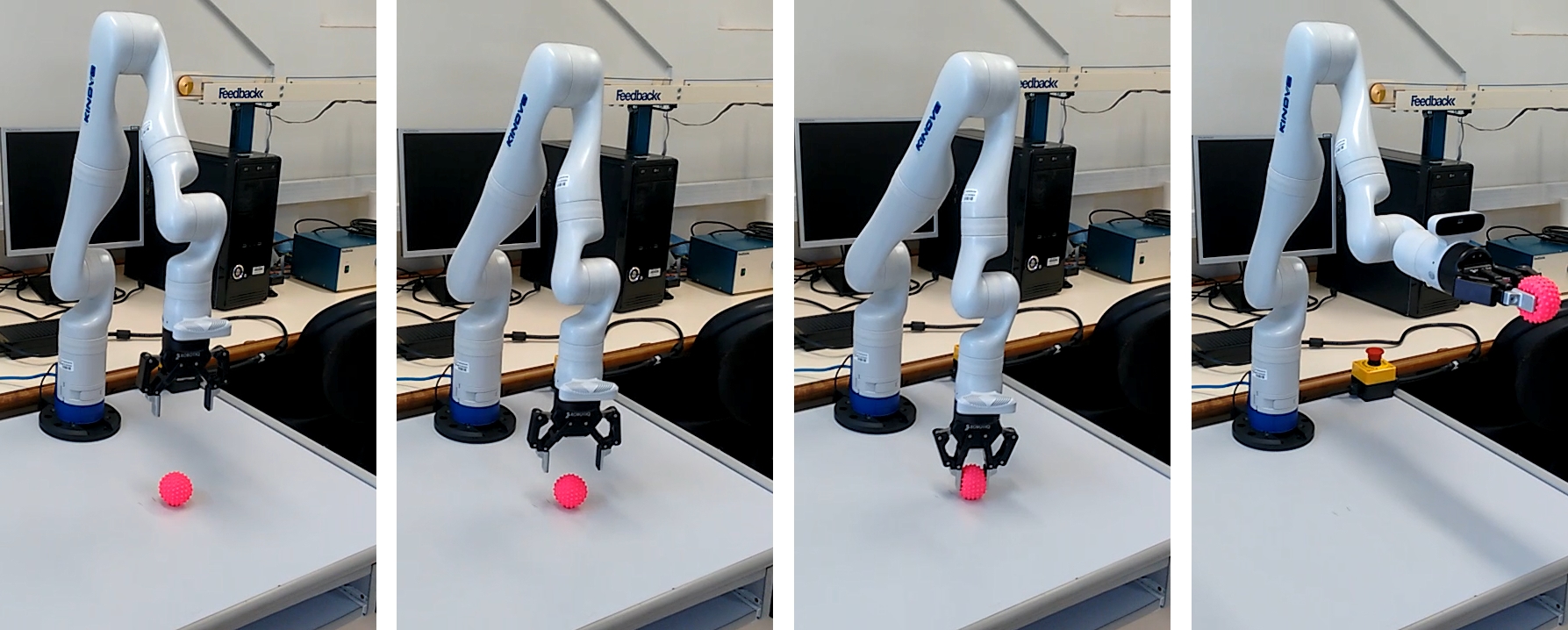}\\  
\caption{Demonstration of the grasp execution right after VS is done}
\label{fig:graspball}
\end{figure}

Fig. \ref{fig:din2} illustrates the environment in which the robot is assembled and shows each step of the process. In Fig. \ref{fig:din21} the robot setup is shown, highlighting the initial pose, followed by the pose in which the robot captures the reference image and the final pose, from where the dynamic grasping starts.

In Fig. \ref{fig:din22}, three images of the robot are shown at different moments of the visual servoing, with the last image illustrating the robot's pose in the instant before the grasp execution when the stop condition is reached. Below each image of the robot, the differences between the current images of each instant and the desired image are also illustrated. The desired position of the ball is represented in blue and the current position in green. During the entire process, it is possible to see the prediction of the grasp network, so that the user can intervene at any time s/he thinks that the grasping can be performed, according to the prediction seen in real-time.

When the stop condition is reached, where the current position of the ball is close enough to the desired position, the robot starts to approach the object for grasping. Fig. \ref{fig:graspball} shows some moments of the grasp execution. In the first and second images, the robot adjusts the orientation and opening of the grippers, and also moves towards the $ (x, y) $ position after mapping the network prediction. Depth information is not used, hence the final $ z $ value is fixed. In the third image, the robot closes its grippers until the applied force reaches a threshold, and then takes the grasped ball to a pre-defined position.

\begin{figure*}[h]
\centering
     \begin{subfigure}[b]{0.35\textwidth}
         \centering
         \includegraphics[width=\textwidth]{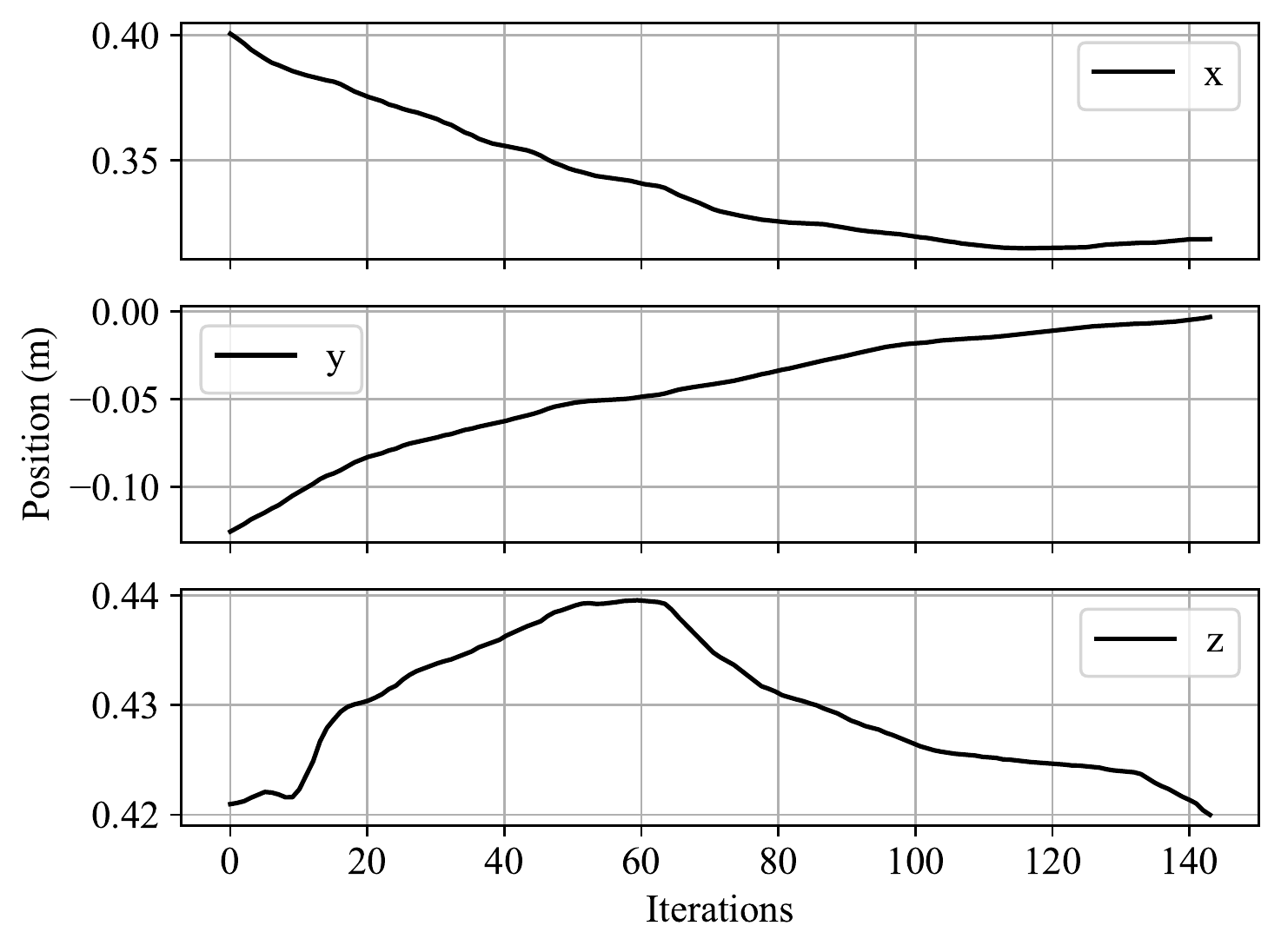}
         \caption{Position per iteration}
         \label{fig:din11}
     \end{subfigure}
     \begin{subfigure}[b]{0.35\textwidth}
         \centering
         \includegraphics[width=\textwidth]{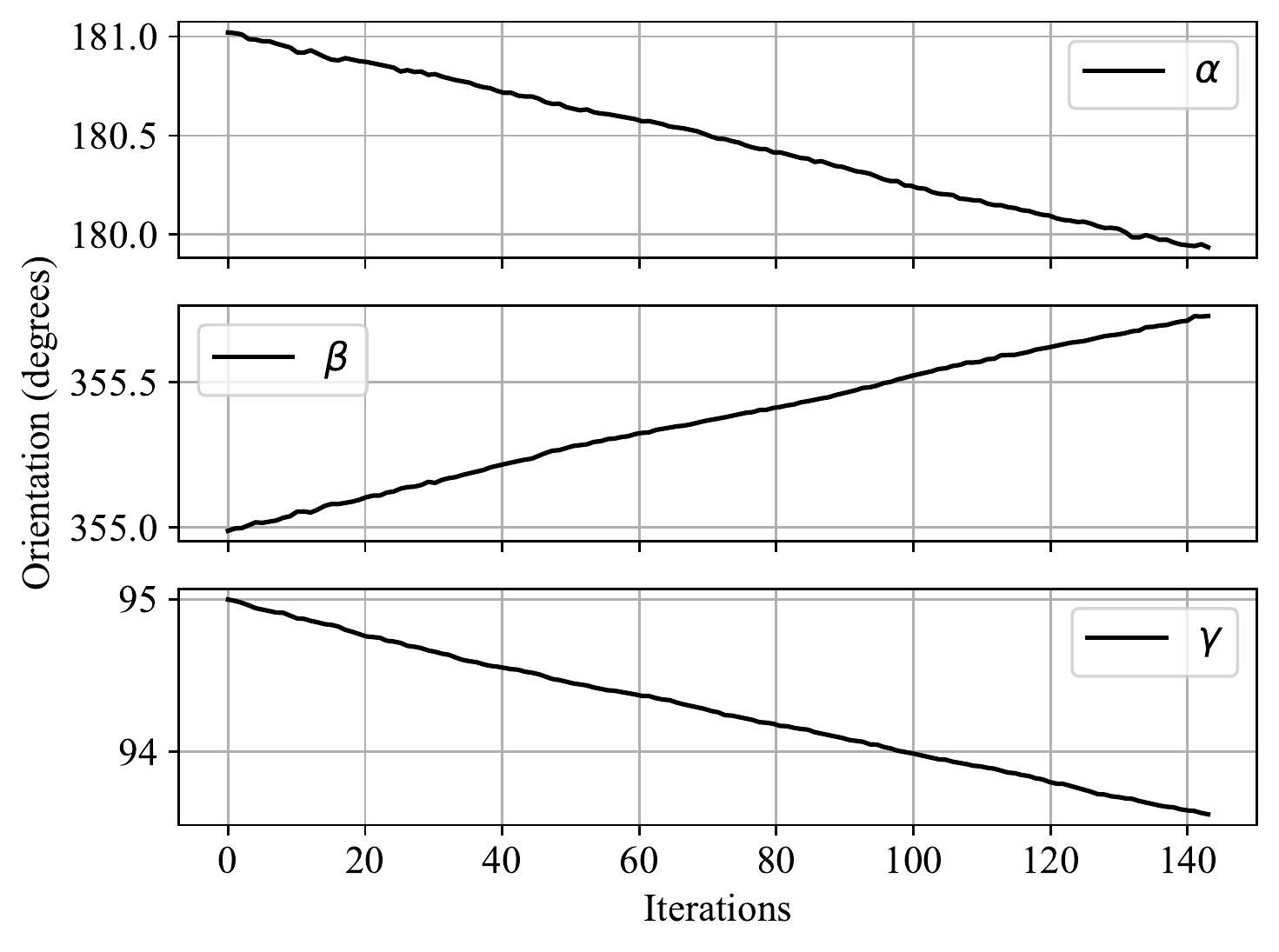}
         \caption{Orientation  per iteration}
         \label{fig:din12}
     \end{subfigure} 
     \begin{subfigure}[b]{0.35\textwidth}
         \centering
         \includegraphics[width=\textwidth]{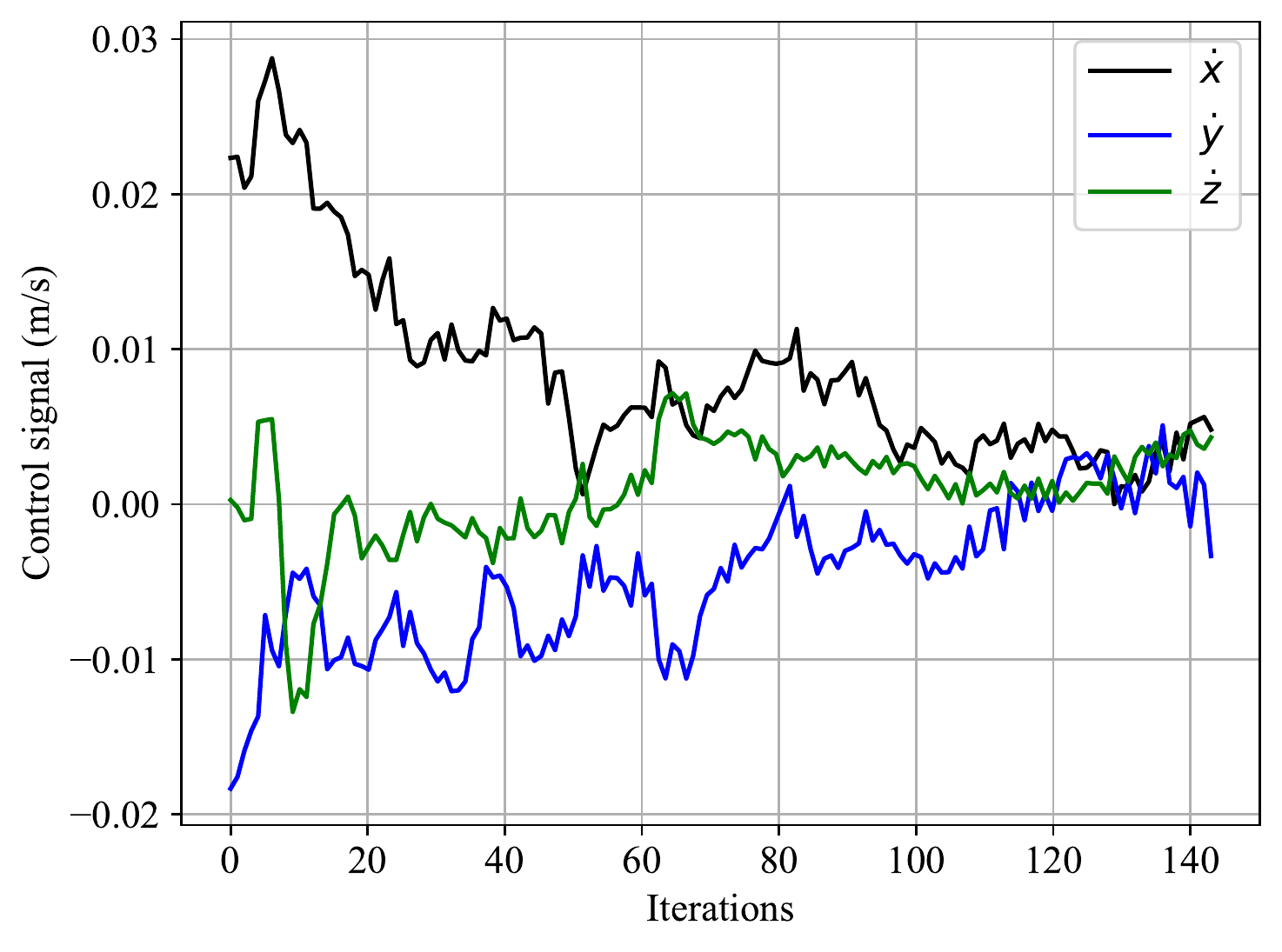}
         \caption{Linear velocity  per iteration}
         \label{fig:din13}
     \end{subfigure}
     \begin{subfigure}[b]{0.35\textwidth}
         \centering
         \includegraphics[width=\textwidth]{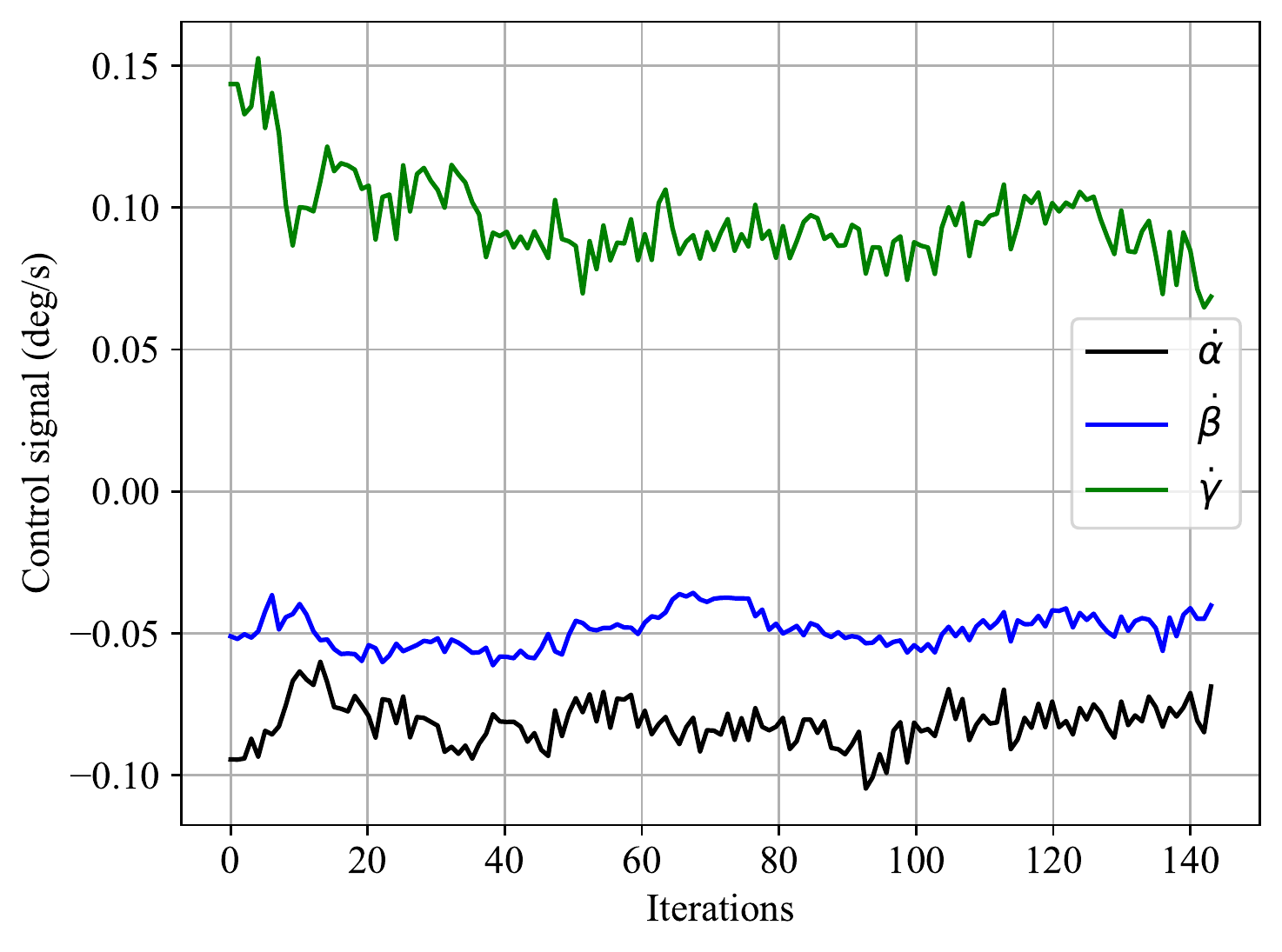}
         \caption{Angular velocity per iteration}
         \label{fig:din14}
     \end{subfigure} 
     \caption{Results for VS in the dynamic grasping scenario: the robot is positioned  $[\Delta x, \Delta y, \Delta z, \Delta \alpha, \Delta \beta, \Delta \gamma] = [0.15m, -0.15m, 0.05m, 5^\circ, -5^\circ, 5^\circ]$ from the desired pose, but as soon as the control signal's norm is below 0.05, the system perceives that it no longer needs to track the object, so the grasp can be executed}
\label{fig:din1}
\end{figure*}

The robot's behavior and the applied control signal are illustrated in Fig. \ref{fig:din1}. The robot comes close to convergence only in $ y $, but as seen by the difference between the current and desired images in Fig. \ref{fig:din2}, it was enough to considerably reduce the distance to the desired position. The behavior in $ z $ starts in the direction of divergence, but after iteration 60, it starts to converge. The angular velocity signal was still relatively large in magnitude, when the VS was interrupted.

Therefore, the choice of the L1-norm threshold must be the trade-off between the positioning accuracy and the speed with which grasping is performed. In a dynamic grasping scenario, speed naturally plays a more important role, so the system should perform the grasping as soon as possible, whereas the visual servoing step serves the task of keeping the object in the camera's field of view, rather than positioning accuracy.

For this reason, a relatively high threshold was chosen, so that the execution of the control does not take too long. It took 14s for the robot to stop visual servoing and perform grasping. After some fine-tuning experiments, we find that threshold to be optimal. However, if time constraints are more strict in particular situations, it can be reduced by considering higher gains and/or thresholds.

\begin{figure}[!htb]
\centering
\includegraphics[width=0.6\linewidth]{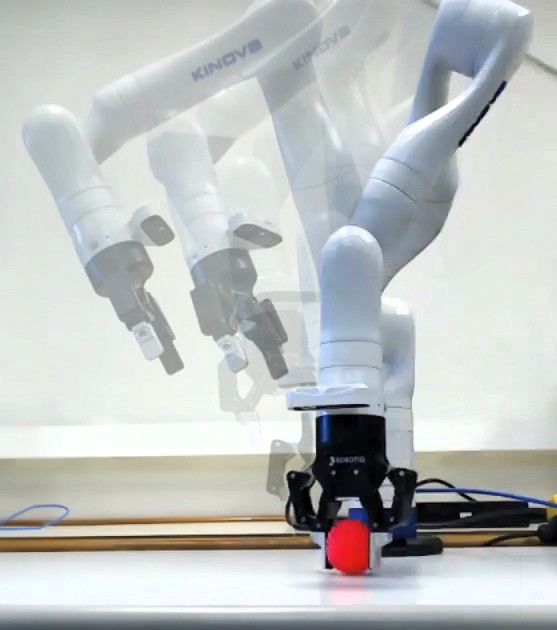}\\  
\caption{3D motion of the robot during grasping attempt}
\label{fig:robot3d}
\end{figure}

In Figs. \ref{fig:din2} and \ref{fig:graspball}, the robot makes a small displacement during the VS and then performs the grasping, since we aim to show the convergence of the control in image space. To illustrate the movement of the robot in Cartesian space for a greater displacement, a sequence of poses is presented in Fig. \ref{fig:robot3d}, to highlight that all 6 dimensions are considered in the experiment.

\begin{figure*}[!htb]
\centering
\includegraphics[width=\textwidth]{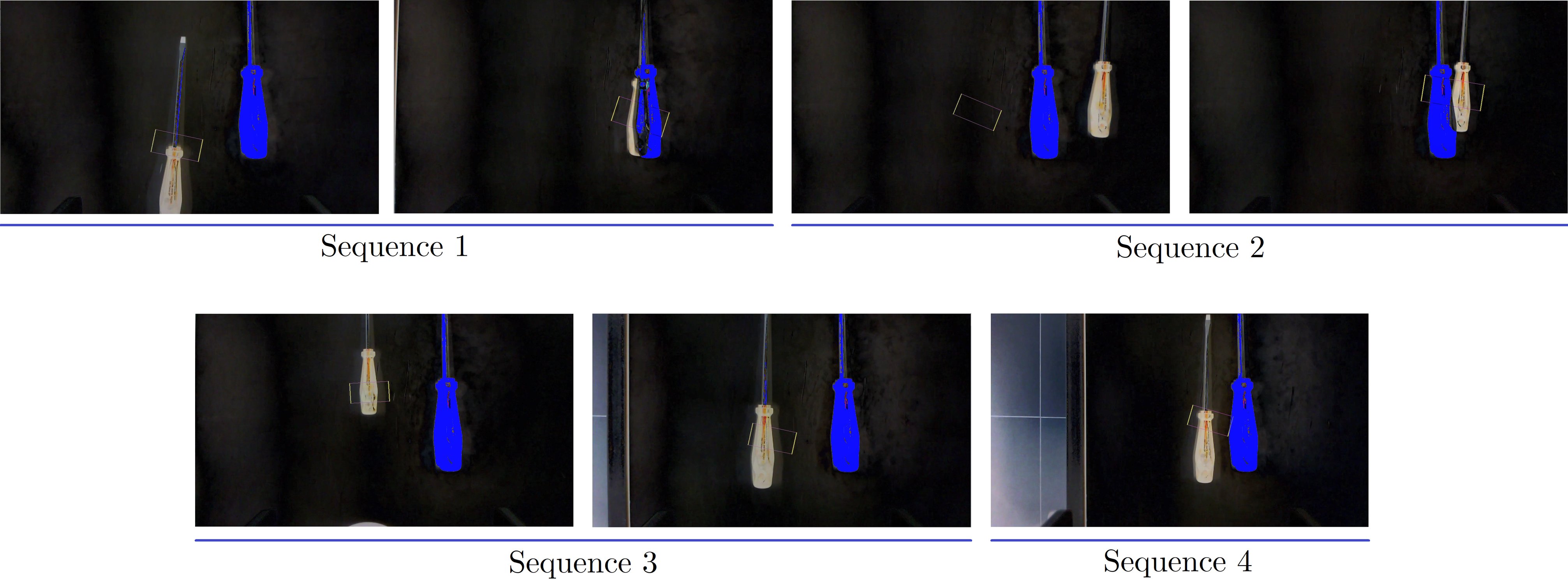}\\  
\caption{Variation of difference between expected and current images during VS} 
\label{fig:sequence}
\end{figure*}

\begin{figure}[h]
\centering
         \includegraphics[width=\linewidth]{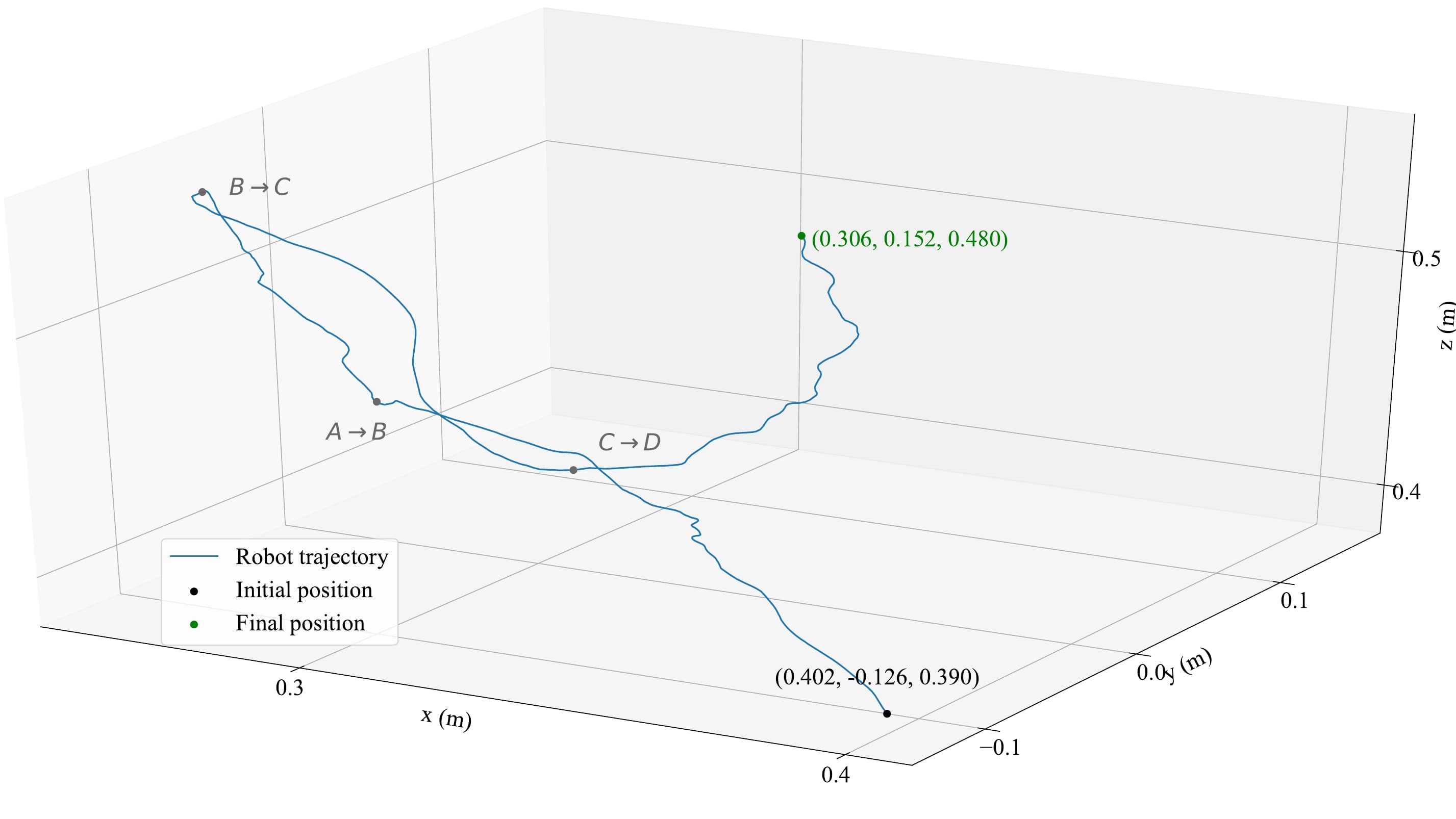}
         \caption{Trajectory performed by the robot during visual servoing for dynamic grasping}
         \label{fig:din35}
\end{figure}

A second experiment to analyze the robot's behavior concerning a dynamic object is performed, in which a quick and unexpected change in the object's position is considered, as well as the case where the object moves with irregular speed. During the experiment, the object's trajectory was not registered, so the description of the motion performed by the object is purely approximate and visual, as illustrated by Fig. \ref{fig:sequence}. The trajectory performed by the robot is illustrated in Fig. \ref{fig:din35}.

The object considered for tracking is a screwdriver (not seen in training), whose desired position is represented in blue in the image. In the first sequence, the same conditions as in the previous experiment are maintained, that is, the robot takes an image in a pre-defined pose, which serves as the desired image, and then moves to another pose to start the control. Sequence 2 occurs right after the object is abruptly moved to the right, causing the robot to track the object and then adjust the position. 

During Sequence 3, the screwdriver is moved with approximately constant speed in $ y $ (robot base frame), causing the robot to follow the movement with an offset between the current and desired images. Finally, in Sequence 4, the robot adjusts its position after the object stops moving, and, as soon as the stop condition is met, the control ends and the grasping is performed.

\begin{figure}[!h]
\centering
     \begin{subfigure}[b]{0.7\linewidth}
         \centering
         \includegraphics[width=\linewidth]{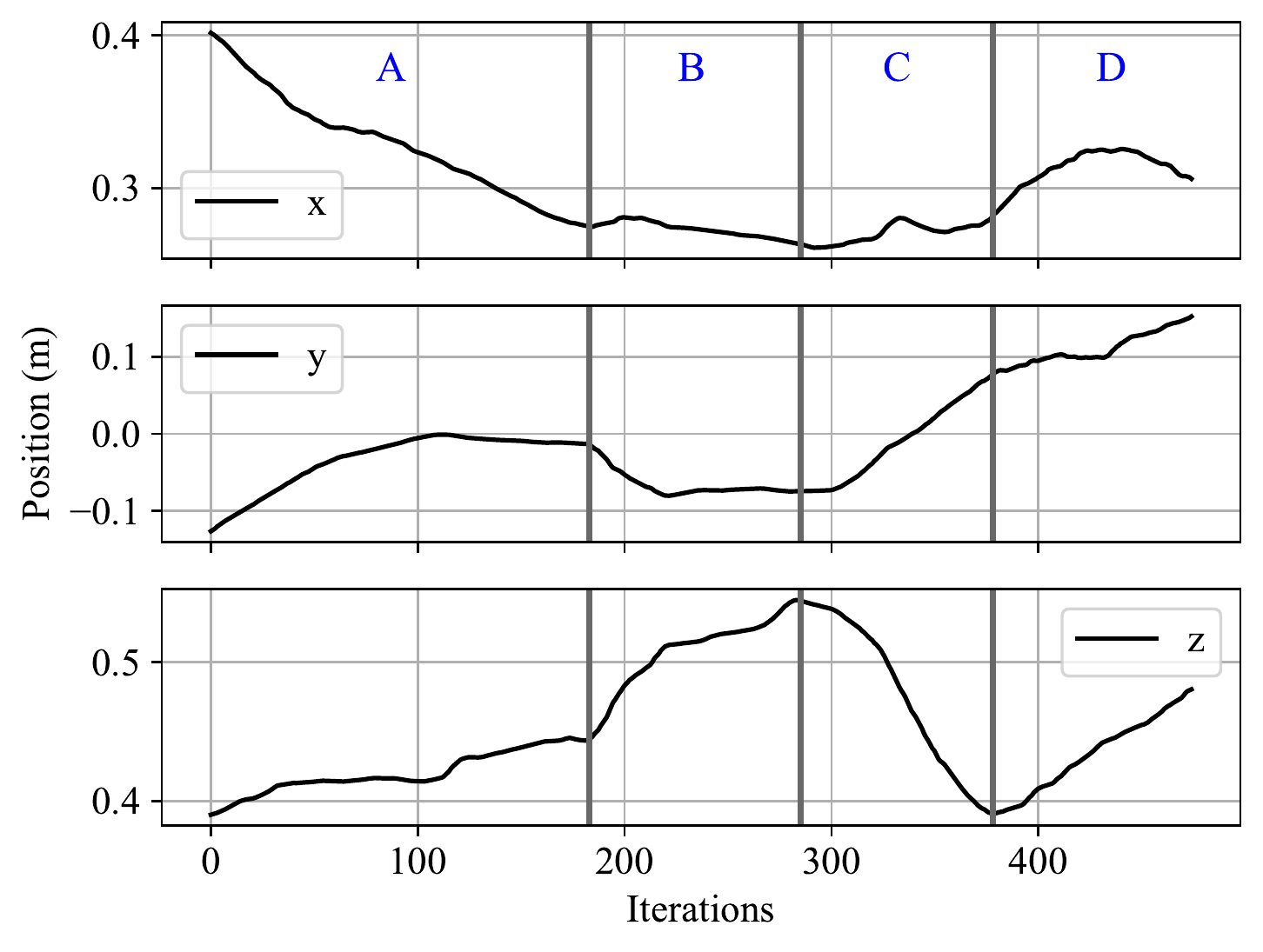}
         \caption{Position per iteration}
         \label{fig:din31}
     \end{subfigure}
     \hfill
     \begin{subfigure}[b]{0.7\linewidth}
         \centering
         \includegraphics[width=\linewidth]{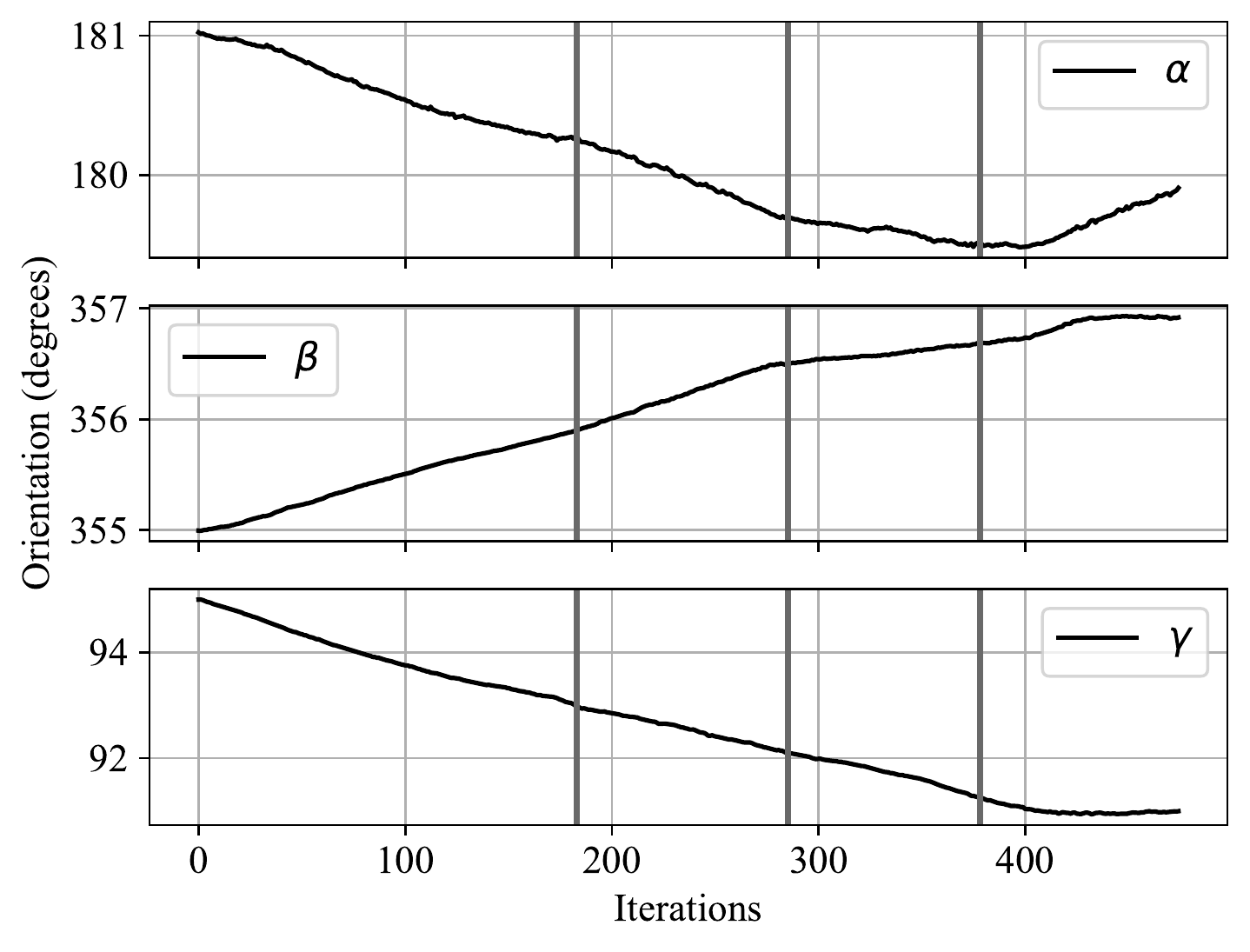}
         \caption{Orientation per iteration}
         \label{fig:din32}
     \end{subfigure} 
     \begin{subfigure}[b]{0.7\linewidth}
         \centering
         \includegraphics[width=\linewidth]{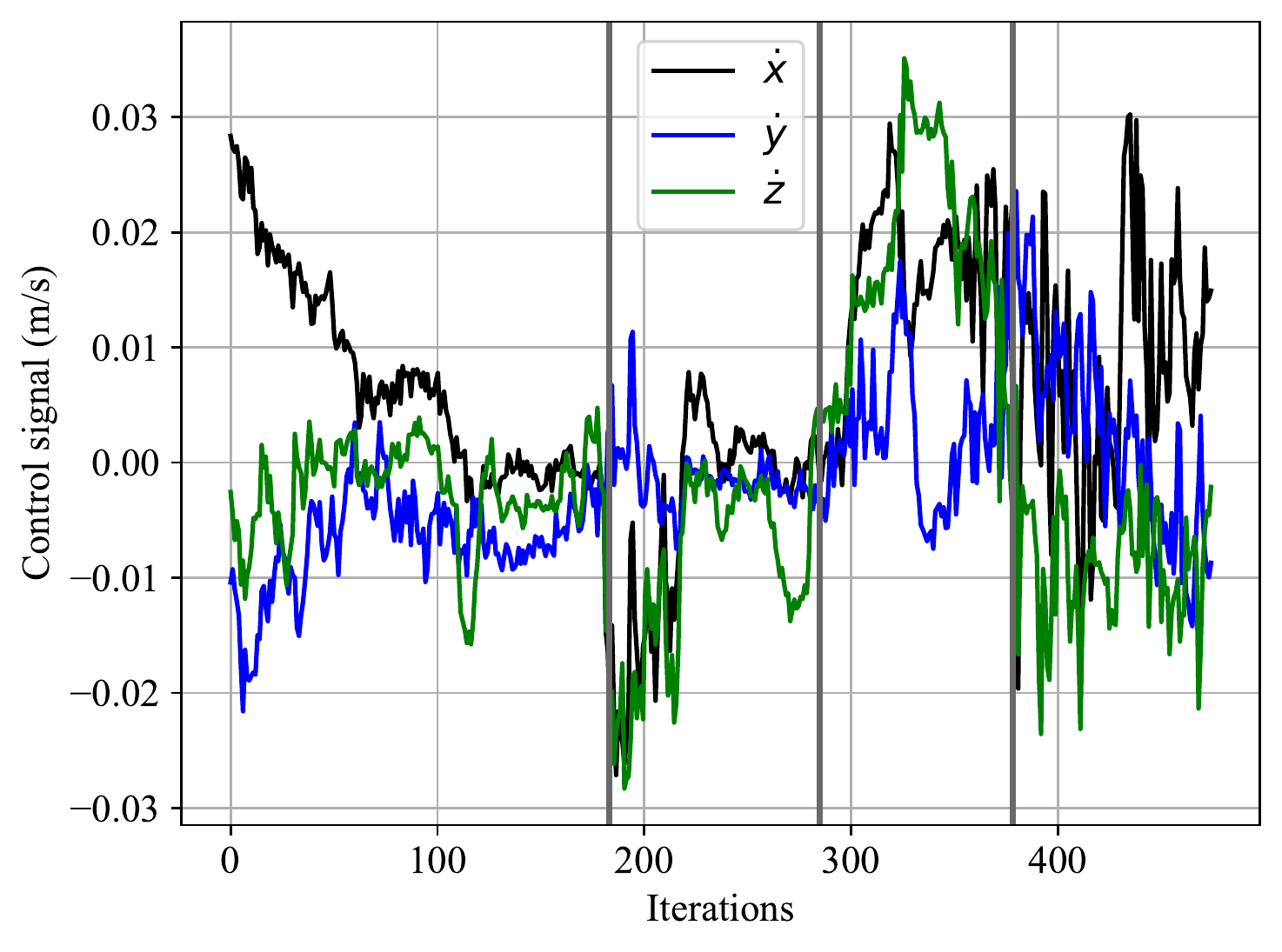}
         \caption{Linear velocity per iteration}
         \label{fig:din33}
     \end{subfigure}
     \hfill
     \begin{subfigure}[b]{0.7\linewidth}
         \centering
         \includegraphics[width=\linewidth]{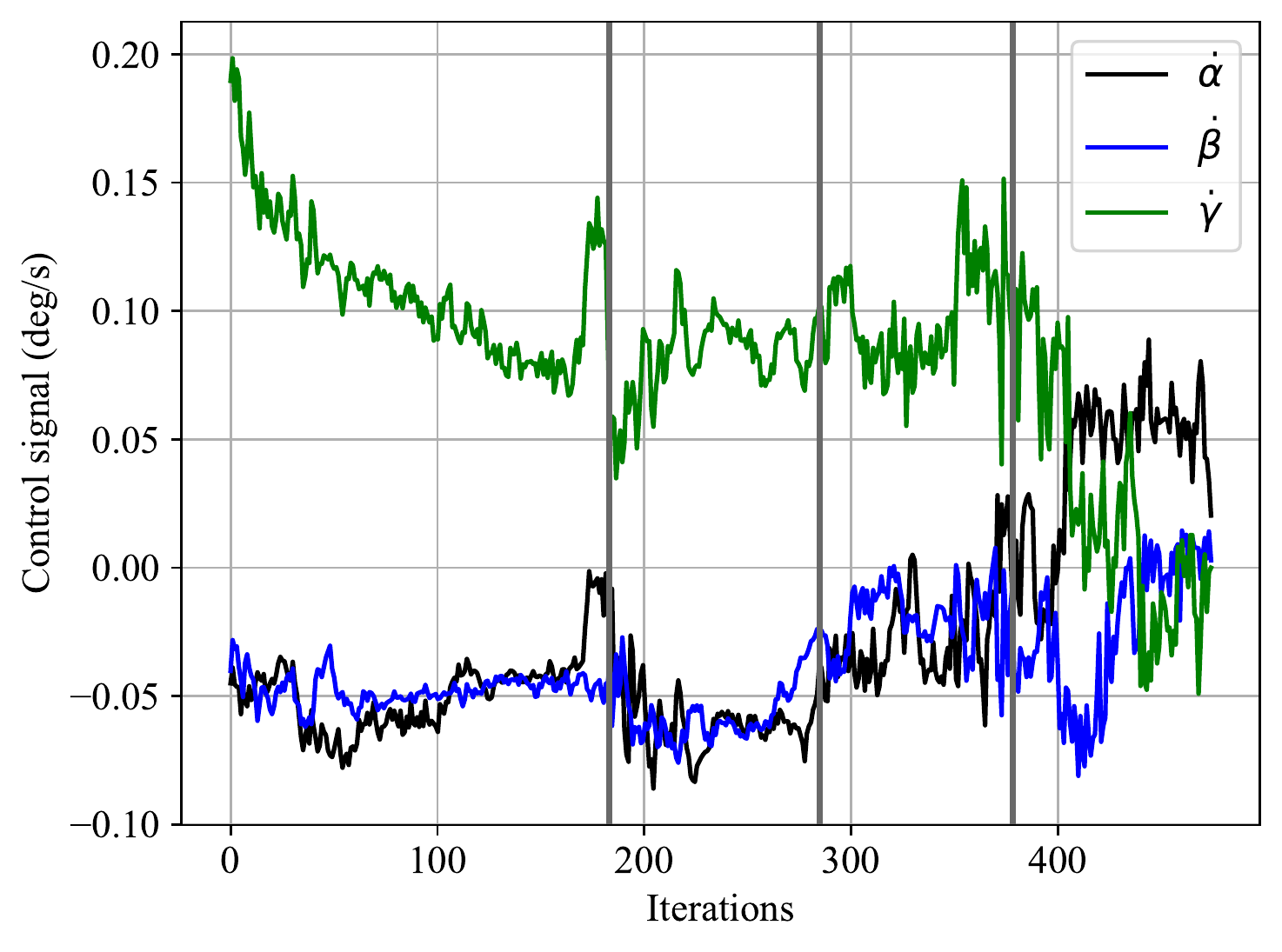}
         \caption{Angular velocity per iteration}
         \label{fig:din34}
     \end{subfigure} 
     \caption{Robot's behaviour in the face of a quick and unexpected change in the object's position (A to B), followed by a displacement with lower, but irregular speed (C) and later adjustment as soon as the object is motionless (D)} 
\label{fig:din3}
\end{figure}

Fig. \ref{fig:din3} represents all stages of the experiment highlighting the transition between sequences. From the linear speed signal, it is possible to notice that the robot was about to stop the visual servoing and perform the grasping, but the sudden change in the screwdriver position causes an overshoot in the signal. The position graph shows that the correct $y$ position is quickly reached and only maintained until the beginning of Sequence 3. From this moment on, the $y$ position of the robot progressively increases as the object moves in that same direction.

From the beginning of Sequence 4, the $y$ position is only adjusted to match the desired image, until the control ends. It is interesting to note that $z$ varies greatly in response to these changes, even though it is not directly involved in them, while $x$ remains almost constant, as expected. This may indicate that it is necessary to adjust a particular gain for $\dot{z}$, or to retrain the network considering larger variations in $z$. At the end of the control, the speeds become very noisy since the floor appears in the image, confusing the network. Even so, the stop condition is reached and the grasping is successfully performed.

A final experiment to evaluate the dynamic grasping system was performed in order to assess the efficiency of the network for different objects and different initial poses. To validate the network's applicability, 70 grasping attempts in a dynamic scenario were considered for 10 single objects and 4 sets of multiple objects. The sets of multiple objects are: 1- five scattered pens; 2- USB cable and glue stick; 3- three transparent whiteboard cleaners; 4- screwdriver, scissors and battery charger.

The robot takes an image at a pre-defined position and then is repositioned considering a Gaussian distribution with SD $[\sigma x, \sigma y, \sigma z, \sigma \alpha, \sigma \beta, \sigma \gamma]$ = [50mm, 50mm, 50mm, $2^{\circ}$, $2^{\circ}$, $2^{\circ}$]. The results are shown in Table \ref{tab:resVS}, indicating a success rate of 85.7\%, in which, of the ten failure cases, six occurred due to errors in the grasp network prediction, and four due to errors in the VS CNN.

\begin{table}[!htb]
\centering
\caption{Result of the dynamic grasping system for 70 attempts involving different objects. G and VS associated with unsuccessful attempts indicate whether the error occurred due to failure of the grasp or visual servoing network prediction.}
\label{tab:resVS}
\resizebox{\linewidth}{!}{
\begin{tabular}{@{}l|c|c|c|c|c@{}}
\hline
\hline
Target object & Attempt 1 & Attempt 2 & Attempt 3 & Attempt 4 & Attempt 5\\
\hline
Screwdriver    & \textcolor{cyan}{\cmark}     & \textcolor{cyan}{\cmark}    & \textcolor{cyan}{\cmark} & \textcolor{cyan}{\cmark} & \textcolor{cyan}{\cmark} \\
Scissors         & \textcolor{cyan}{\cmark}     & \textcolor{cyan}{\cmark}     & \textcolor{cyan}{\cmark} & \textcolor{red}{\xmark} (G) & \textcolor{cyan}{\cmark} \\
Pen     &  \textcolor{cyan}{\cmark}     & \textcolor{cyan}{\cmark}    & \textcolor{cyan}{\cmark} & \textcolor{cyan}{\cmark} & \textcolor{cyan}{\cmark}\\
Masking tape    & \textcolor{red}{\xmark} (VS)     & \textcolor{cyan}{\cmark}     & \textcolor{cyan}{\cmark} & \textcolor{red}{\xmark} (G) & \textcolor{cyan}{\cmark} \\
Pliers    & \textcolor{cyan}{\cmark}     & \textcolor{cyan}{\cmark}    & \textcolor{red}{\xmark} (G) & \textcolor{cyan}{\cmark} & \textcolor{cyan}{\cmark} \\
USB cable         & \textcolor{cyan}{\cmark}     & \textcolor{cyan}{\cmark}     & \textcolor{cyan}{\cmark} & \textcolor{cyan}{\cmark} & \textcolor{cyan}{\cmark} \\
Cup      &  \textcolor{red}{\xmark} (G)     & \textcolor{cyan}{\cmark}    & \textcolor{cyan}{\cmark} & \textcolor{cyan}{\cmark} & \textcolor{cyan}{\cmark}\\
Staples box    & \textcolor{cyan}{\cmark}     & \textcolor{cyan}{\cmark}     & \textcolor{cyan}{\cmark} & \textcolor{cyan}{\cmark} & \textcolor{cyan}{\cmark} \\
Bottle    & \textcolor{cyan}{\cmark}     & \textcolor{cyan}{\cmark}    & \textcolor{cyan}{\cmark} & \textcolor{red}{\xmark} (VS) & \textcolor{cyan}{\cmark} \\
Ball         & \textcolor{cyan}{\cmark}     & \textcolor{cyan}{\cmark}     & \textcolor{cyan}{\cmark} &  \textcolor{cyan}{\cmark} & \textcolor{red}{\xmark} (G) \\
Multiple 1     &  \textcolor{cyan}{\cmark}     & \textcolor{cyan}{\cmark}    & \textcolor{cyan}{\cmark} & \textcolor{cyan}{\cmark} & \textcolor{cyan}{\cmark}\\
Multiple 2   & \textcolor{red}{\xmark} (VS)     & \textcolor{cyan}{\cmark}     & \textcolor{cyan}{\cmark} & \textcolor{cyan}{\cmark} & \textcolor{cyan}{\cmark} \\
Multiple 3   & \textcolor{cyan}{\cmark}     & \textcolor{red}{\xmark} (VS)    & \textcolor{cyan}{\cmark} & \textcolor{cyan}{\cmark} & \textcolor{cyan}{\cmark} \\
Multiple  4       & \textcolor{cyan}{\cmark}     & \textcolor{cyan}{\cmark}     & \textcolor{cyan}{\cmark} & \textcolor{red}{\xmark} (G) & \textcolor{cyan}{\cmark} \\
\hline
\hline
\end{tabular}
}
\end{table}

\subsubsection{Real-World Comparison}

Table \ref{tab:realcomp} lists other 8 works that also use learning tools to perform grasping with parallel grippers. The presented success rates refer to the efficiency of the algorithms in real robots. Many of them deal only with static scenarios, in which it is not possible to handle changes in the environment and dynamic objects.

Studying reactive systems that are able to tackle dynamic object problems, Gualtieri et al. \cite{gualtieri2016high} were one of the first researchers to report the success rate for their method, which reached an accuracy of $22.5\%$. In a recent work, Julian et al. \cite{julian2020stop} investigated the effects of continuous learning to improve the success rate of grasping. However, to accomplish such an objective, the authors used a pre-trained model with 580,000 grasp attempts and, afterwards, it still required 15,000 more attempts. It took a month to collect the additional attempts only. The average efficiency of all experiments presented by the authors in the dynamic scenario was of $80.4\%$.

Viereck et al. \cite{viereck2017learning} reported a success rate of $77\% (58/75)$ when approaching dynamic objects in a reinforcement learning framework. Morrison, Corke and Leitner \cite{morrison2019learning}, on the other hand, developed a reactive grasping system that is similar to ours, except for the fact that their approach for visual servoing is a classic PBVS dependent on grasping prediction. The authors obtained a promising accuracy of $81\% (76/94)$. Both works addressed cluttered scenes, which may hinder grasp detection, yet the dynamics of the set are not complex. Such dynamics consist of moving the observed object 10cm in one direction and rotating it $25^{\circ}$ in $z$, which does not reproduce a challenging scenario. Considering single dynamic objects, the method of Morrison, Corke and Leitner \cite{morrison2019learning} reached an efficiency of $86\%$ in 200 attempts, yet the control is performed by the same classic PBVS, relying on the pose predicted by the grasping network. The dynamics of the set are the same as the ones presented for the cluttered scenario. 

The average final efficiency of all experiments presented by Morrison, Corke and Leitner is of $84\%$. This is the work that comes closest to the final efficiency of our method - $85.7\%$.

We highlight that, in addition to having the highest success rate among the listed works, our approach also has the advantage of being lightweight and agnostic to the robotic platform, in addition to not requiring months of data collection. Furthermore, errors in the grasp detection do not induce errors in the control, since these systems are dissociated.

\begin{table}
\centering
\caption{Comparison of other systems using learning tools to perform grasping with parallel grippers. The reported success rate is the average of all real robot experiments.}
\label{layers}
\resizebox{\linewidth}{!}{
\begin{tabular}{@{}>{}c|>{}c|>{}c@{}}
\hline
\hline
Approach                  & Environment  & Success rate \\ 
\hline
Lenz, Lee and Saxena (2015) \cite{lenz2015deep}   & Static     & 89\%        \\
Johns, Leutenegger and Davison (2016) \cite{johns2016deep}           & Static     & 80\%        \\
Pinto and Gupta (2016) \cite{pinto2016supersizing}              & Static       & 73\%          \\

Mahler et al. (2017) \cite{mahler2017dex}   & Static      & 86.5\%      \\

Morrison, Corke and Leitner (2019) \cite{morrison2019learning}           & Static       & 92\%          \\
Julian et al. (2020) \cite{julian2020stop}   & Static      & 84.7\%     \\
\hline
Gualtieri et al. (2016) \cite{gualtieri2016high} & Dynamic/Reactive & 22.5\% \\
Viereck et al. (2017) \cite{viereck2017learning}              & Dynamic/Reactive      & 77.3\%          \\
Morrison, Corke and Leitner (2019) \cite{morrison2019learning}   & Dynamic/Reactive       & 84\%     \\
Julian et al. (2020) \cite{julian2020stop}              & Dynamic/Reactive        &  80.4\%         \\
Ours   & Dynamic/Reactive       & \textbf{85.7\%}     \\
\hline
\hline
\end{tabular}
}
\label{tab:realcomp}
\end{table}

\section{Conclusions}

In this work, we addressed the problems of grasp detection and visual servoing using deep learning, and we also applied them as an approach to the problem of grasping dynamic objects.

A convolutional neural network is employed to obtain the points where a robot with parallel grippers can grasp an object using only RGB information. In order to use a simple network with a small number of parameters, the data augmentation technique is employed innovatively, allowing to extract as much visual information from the dataset images as possible. As a result, the trained network is able to predict grasp rectangles at speeds that surpass the state-of-the-art in the CGD, without harshly penalizing accuracy. The trained network is also evaluated on a robot, with its own camera to test its generalization to different objects that were not seen in the training. The visual results demonstrate that the network is robust to real-world noise and capable of detecting grasps on objects of different shapes, colors, in several orientations and subjected to variable lighting conditions.

Four CNN models were designed as potential candidates for end-to-end visual servo controllers. The networks do not use any additional information other than a reference image and the current image to regress the control signal. These networks were tested both offline and on a robot to assess applicability in a real-world and real-time scenario. The simplest model was able to achieve a millimeter accuracy in the final position considering a target object seen for the first time. To the best of our knowledge, we have not found other works that achieve such precision with a controller learned from scratch in the literature.

Finally, the trained networks that obtained the best results in each task, were embedded in a final system to grasp dynamic objects. In different scenarios, all \textit{a priori} unknown to the network, the robot was able to keep the target object in the camera's field of view using the real-time VS CNN. When the robot understands that it is close enough to the desired position, the prediction of the grasping network, which is also processed in real-time, is mapped to world coordinates and the robot approaches the object to execute grasp.

The system at the current stage has the deficiency of not considering depth information during the execution of the grasping. In future work, we aim to use information from an infrared sensor in the operational stage, maintaining training only with RGB images, as this is seen as an advantage. Other considered adjustments relate to the determination of control gains, adaptation of the grasping network to predict multiple rectangles, and obtaining a grasping dataset in a self-supervised manner with the robot.

The trained algorithms were not designed to surpass those modeled from prior knowledge, but to be applied in situations where the systems's knowledge is limited or impossible to be obtained. To this end, it is concluded that the developed methodology has superior applicability due to its demonstrated capacity for generalization, simplicity, speed, and accuracy.


\section*{Acknowledgment}

This research was financed in part by the Brazilian National Council for Scientific and Technological Development - CNPq (grants 465755/2014-3 and 130602/2018-3), by the Coordination of Improvement of Higher Education Personnel - Brazil - CAPES (Finance Code 001, grants 88887.136349/2017-00 and 88887.339070/2019-00), and the S\~{a}o Paulo Research Foundation - FAPESP (grant {2014/50851-0}).


\begin{thebibliography}{10}
\expandafter\ifx\csname url\endcsname\relax
  \def\url#1{\texttt{#1}}\fi
\expandafter\ifx\csname urlprefix\endcsname\relax\def\urlprefix{URL }\fi
\expandafter\ifx\csname href\endcsname\relax
  \def\href#1#2{#2} \def\path#1{#1}\fi

\bibitem{kumra2017robotic}
S.~Kumra, C.~Kanan, Robotic grasp detection using deep convolutional neural
  networks, in: 2017 IEEE/RSJ International Conference on Intelligent Robots
  and Systems (IROS), IEEE, Vancouver, Canada, 2017, pp. 769--776.

\bibitem{ribeiro2019fast}
E.~G. Ribeiro, V.~Grassi, Fast convolutional neural network for real-time
  robotic grasp detection, in: 2019 19th International Conference on Advanced
  Robotics (ICAR), IEEE, Belo Horizonte, Brazil, 2019, pp. 49--54.

\bibitem{bohg2014data}
J.~Bohg, A.~Morales, T.~Asfour, D.~Kragic, Data-driven grasp synthesis—a
  survey, IEEE Transactions on Robotics 30~(2) (2014) 289--309.

\bibitem{shimoga1996robot}
K.~B. Shimoga, Robot grasp synthesis algorithms: A survey, The International
  Journal of Robotics Research 15~(3) (1996) 230--266.

\bibitem{rubert2017relevance}
C.~Rubert, D.~Kappler, A.~Morales, S.~Schaal, J.~Bohg, On the relevance of
  grasp metrics for predicting grasp success, in: Intelligent Robots and
  Systems (IROS), 2017 IEEE/RSJ International Conference on, IEEE, Vancouver,
  Canada, 2017, pp. 265--272.

\bibitem{levine2018learning}
S.~Levine, P.~Pastor, A.~Krizhevsky, J.~Ibarz, D.~Quillen, Learning hand-eye
  coordination for robotic grasping with deep learning and large-scale data
  collection, The International Journal of Robotics Research 37~(4-5) (2018)
  421--436.

\bibitem{pinto2016supersizing}
L.~Pinto, A.~Gupta, Supersizing self-supervision: Learning to grasp from 50k
  tries and 700 robot hours, in: Robotics and Automation (ICRA), 2016 IEEE
  International Conference on, IEEE, Stockholm, Sweden, 2016, pp. 3406--3413.

\bibitem{kappler2015leveraging}
D.~Kappler, J.~Bohg, S.~Schaal, Leveraging big data for grasp planning, in:
  2015 IEEE International Conference on Robotics and Automation (ICRA), IEEE,
  Seattle, USA, 2015, pp. 4304--4311.

\bibitem{lenz2015deep}
I.~Lenz, H.~Lee, A.~Saxena, Deep learning for detecting robotic grasps, The
  International Journal of Robotics Research 34~(4-5) (2015) 705--724.

\bibitem{saxena2008robotic}
A.~Saxena, J.~Driemeyer, A.~Y. Ng, Robotic grasping of novel objects using
  vision, The International Journal of Robotics Research 27~(2) (2008)
  157--173.

\bibitem{du2019vision}
G.~Du, K.~Wang, S.~Lian, Vision-based robotic grasping from object
  localization, pose estimation, grasp detection to motion planning: A review,
  arXiv preprint arXiv:1905.06658.

\bibitem{viereck2017learning}
U.~Viereck, A.~Pas, K.~Saenko, R.~Platt, Learning a visuomotor controller for
  real world robotic grasping using simulated depth images, in: Conference on
  Robot Learning, CoRL, Mountain View, USA, 2017, pp. 291--300.

\bibitem{lampe2013acquiring}
T.~Lampe, M.~Riedmiller, Acquiring visual servoing reaching and grasping skills
  using neural reinforcement learning, in: The 2013 international joint
  conference on neural networks (IJCNN), IEEE, Dallas, USA, 2013, pp. 1--8.

\bibitem{kragic2002survey}
D.~Kragic, H.~I. Christensen, et~al., Survey on visual servoing for
  manipulation, Computational Vision and Active Perception Laboratory,
  Fiskartorpsv 15 (2002) 2002.

\bibitem{saxena2017exploring}
A.~Saxena, H.~Pandya, G.~Kumar, A.~Gaud, K.~M. Krishna, Exploring convolutional
  networks for end-to-end visual servoing, in: 2017 IEEE International
  Conference on Robotics and Automation (ICRA), IEEE, Marina Bay Sands,
  Singapore, 2017, pp. 3817--3823.
\newblock \href {http://dx.doi.org/10.1109/ICRA.2017.7989442}
  {\path{doi:10.1109/ICRA.2017.7989442}}.

\bibitem{bicchi2000robotic}
A.~Bicchi, V.~Kumar, Robotic grasping and contact: A review, in: Intenational
  Conference on Robotics and Automation, Vol. 348, IEEE, San Francisco, USA,
  2000, p. 353.

\bibitem{jiang2011efficient}
Y.~Jiang, S.~Moseson, A.~Saxena, Efficient grasping from rgbd images: Learning
  using a new rectangle representation, in: Robotics and Automation (ICRA),
  2011 IEEE International Conference on, IEEE, Shanghai, China, 2011, pp.
  3304--3311.

\bibitem{redmon2015real}
J.~Redmon, A.~Angelova, Real-time grasp detection using convolutional neural
  networks, in: Robotics and Automation (ICRA), 2015 IEEE International
  Conference on, IEEE, Seattle, 2015, pp. 1316--1322.

\bibitem{krizhevsky2012imagenet}
A.~Krizhevsky, I.~Sutskever, G.~E. Hinton, Imagenet classification with deep
  convolutional neural networks, in: Twenty-sixth Conference on Neural
  Information Processing Systems, NIPSF, Lake Tahoe, USA, 2012, pp. 1097--1105.

\bibitem{he2016deep}
K.~He, X.~Zhang, S.~Ren, J.~Sun, Deep residual learning for image recognition,
  in: Proceedings of the IEEE conference on computer vision and pattern
  recognition, IEEE, Las Vegas, USA, 2016, pp. 770--778.

\bibitem{ren2015faster}
S.~Ren, K.~He, R.~Girshick, J.~Sun, Faster r-cnn: Towards real-time object
  detection with region proposal networks, in: Advances in neural information
  processing systems, NIPSF, Montréal, Canada, 2015, pp. 91--99.

\bibitem{chu2018real}
F.-J. Chu, R.~Xu, P.~A. Vela, Real-world multiobject, multigrasp detection,
  IEEE Robotics and Automation Letters 3~(4) (2018) 3355--3362.

\bibitem{zhou2018fully}
X.~Zhou, X.~Lan, H.~Zhang, Z.~Tian, Y.~Zhang, N.~Zheng, Fully convolutional
  grasp detection network with oriented anchor box, in: 2018 IEEE/RSJ
  International Conference on Intelligent Robots and Systems (IROS), IEEE,
  Madrid, Spain, 2018, pp. 7223--7230.

\bibitem{Morrison_2018}
D.~Morrison, J.~Leitner, P.~Corke, Closing the loop for robotic grasping: A
  real-time, generative grasp synthesis approach, Robotics: Science and Systems
  XIV\href {http://dx.doi.org/10.15607/rss.2018.xiv.021}
  {\path{doi:10.15607/rss.2018.xiv.021}}.

\bibitem{gu2019attention}
Q.~Gu, J.~Su, X.~Bi, Attention grasping network: A real-time approach to
  generating grasp synthesis, in: 2019 IEEE International Conference on
  Robotics and Biomimetics (ROBIO), IEEE, Yunnan, China, 2019, pp. 3036--3041.

\bibitem{lee2017learning}
A.~X. Lee, S.~Levine, P.~Abbeel, Learning visual servoing with deep features
  and fitted q-iteration, arXiv preprint arXiv:1703.11000.

\bibitem{chaumette1998potential}
F.~Chaumette, Potential problems of stability and convergence in image-based
  and position-based visual servoing, in: D.~Kriegman, G.~Hager, A.~Morse
  (Eds.), The confluence of vision and control. Lecture Notes in Control and
  Information Sciences, Springer, London, 1998, pp. 66--78.

\bibitem{miller1987sensor}
W.~Miller, Sensor-based control of robotic manipulators using a general
  learning algorithm, IEEE Journal on Robotics and Automation 3~(2) (1987)
  157--165.

\bibitem{wei1999multisensory}
G.-Q. Wei, G.~Hirzinger, Multisensory visual servoing by a neural network, IEEE
  Transactions on Systems, Man, and Cybernetics, Part B (Cybernetics) 29~(2)
  (1999) 276--280.

\bibitem{ramachandram2003short}
D.~Ramachandram, M.~Rajeswari, A short review of neural network techniques in
  visual servoing of robotic manipulators, in: Malaysia-Japan Seminar On
  Artificial Intelligence Applications In Industry, AIAI, Kuala Lumpur,
  Malaysia, 2003, p. 2425.

\bibitem{pandya2015servoing}
H.~Pandya, K.~M. Krishna, C.~Jawahar, Servoing across object instances: Visual
  servoing for object category, in: 2015 IEEE International Conference on
  Robotics and Automation (ICRA), IEEE, Seattle, USA, 2015, pp. 6011--6018.

\bibitem{Deguchi2000}
K.~Deguchi, A direct interpretation of dynamic images with camera and object
  motions for vision guided robot control, International Journal of Computer
  Vision 37~(1) (2000) 7--20.
\newblock \href {http://dx.doi.org/10.1023/A:1008151528479}
  {\path{doi:10.1023/A:1008151528479}}.

\bibitem{caron2013photometric}
G.~Caron, E.~Marchand, E.~M. Mouaddib, Photometric visual servoing for
  omnidirectional cameras, Autonomous Robots 35~(2-3) (2013) 177--193.

\bibitem{silveira2012direct}
G.~Silveira, E.~Malis, Direct visual servoing: Vision-based estimation and
  control using only nonmetric information, IEEE Transactions on Robotics
  28~(4) (2012) 974--980.

\bibitem{silveira2014intensity}
G.~Silveira, On intensity-based nonmetric visual servoing, IEEE Transactions on
  Robotics 30~(4) (2014) 1019--1026.

\bibitem{zhang2015towards}
F.~Zhang, J.~Leitner, M.~Milford, B.~Upcroft, P.~Corke, Towards vision-based
  deep reinforcement learning for robotic motion control, arXiv preprint
  arXiv:1511.03791.

\bibitem{sampedro2018image}
C.~Sampedro, A.~Rodriguez-Ramos, I.~Gil, L.~Mejias, P.~Campoy, Image-based
  visual servoing controller for multirotor aerial robots using deep
  reinforcement learning, in: 2018 IEEE/RSJ International Conference on
  Intelligent Robots and Systems (IROS), IEEE, Madrid, Spain, 2018, pp.
  979--986.

\bibitem{shi2016decoupled}
H.~Shi, X.~Li, K.-S. Hwang, W.~Pan, G.~Xu, Decoupled visual servoing with
  fuzzyq-learning, IEEE Transactions on Industrial Informatics 14~(1) (2016)
  241--252.

\bibitem{dosovitskiy2015flownet}
A.~Dosovitskiy, P.~Fischer, E.~Ilg, P.~Hausser, C.~Hazirbas, V.~Golkov,
  P.~v.~d. Smagt, D.~Cremers, T.~Brox, Flownet: Learning optical flow with
  convolutional networks, in: 2015 International Conference on Computer Vision
  (ICCV), IEEE, Santiago, Chile, 2015.

\bibitem{glocker2013real}
B.~Glocker, S.~Izadi, J.~Shotton, A.~Criminisi, Real-time rgb-d camera
  relocalization, in: 2013 International Symposium on Mixed and Augmented
  Reality (ISMAR), IEEE, Adelaide, Australia, 2013, pp. 173--179.

\bibitem{bateux2018training}
Q.~Bateux, E.~Marchand, J.~Leitner, F.~Chaumette, P.~Corke, Training deep
  neural networks for visual servoing, in: ICRA 2018-IEEE International
  Conference on Robotics and Automation, IEEE, Brisbane, Australia, 2018, pp.
  1--8.

\bibitem{simonyan2014very}
K.~Simonyan, A.~Zisserman, Very deep convolutional networks for large-scale
  image recognition, arXiv preprint arXiv:1409.1556.

\bibitem{piater2011learning}
J.~Piater, S.~Jodogne, R.~Detry, D.~Kraft, N.~Kr{\"u}ger, O.~Kroemer,
  J.~Peters, Learning visual representations for perception-action systems, The
  International Journal of Robotics Research 30~(3) (2011) 294--307.

\bibitem{piepmeier2004uncalibrated}
J.~A. Piepmeier, G.~V. McMurray, H.~Lipkin, Uncalibrated dynamic visual
  servoing, IEEE Transactions on Robotics and Automation 20~(1) (2004)
  143--147.

\bibitem{shademan2010robust}
A.~Shademan, A.-M. Farahmand, M.~J{\"a}gersand, Robust jacobian estimation for
  uncalibrated visual servoing, in: 2010 IEEE International Conference on
  Robotics and Automation, IEEE, Anchorage, USA, 2010, pp. 5564--5569.

\bibitem{ratliff2007imitation}
N.~Ratliff, J.~A. Bagnell, S.~S. Srinivasa, Imitation learning for locomotion
  and manipulation, in: 2007 7th IEEE-RAS International Conference on Humanoid
  Robots, IEEE, Pittsburgh, USA, 2007, pp. 392--397.

\bibitem{stulp2011learning}
F.~Stulp, E.~Theodorou, J.~Buchli, S.~Schaal, Learning to grasp under
  uncertainty, in: 2011 IEEE International Conference on Robotics and
  Automation, IEEE, Shanghai, China, 2011, pp. 5703--5708.

\bibitem{levine2016learning}
S.~Levine, P.~Pastor, A.~Krizhevsky, D.~Quillen, Learning hand-eye coordination
  for robotic grasping with large-scale data collection, in: International
  symposium on experimental robotics, Springer, Roppongi, Tokyo, 2016, pp.
  173--184.

\bibitem{ten2018using}
A.~ten Pas, R.~Platt, Using geometry to detect grasp poses in 3d point clouds,
  in: Robotics Research, Springer, 2018, pp. 307--324.

\bibitem{wang2019homography}
A.~S. Wang, W.~Zhang, D.~Troniak, J.~Liang, O.~Kroemer, Homography-based deep
  visual servoing methods for planar grasps, in: 2019 IEEE/RSJ International
  Conference on Intelligent Robots and Systems (IROS), IEEE, Macau, China,
  2019, pp. 6570--6577.

\bibitem{morrison2019learning}
D.~Morrison, P.~Corke, J.~Leitner, Learning robust, real-time, reactive robotic
  grasping, The International Journal of Robotics Research (2019) 183--201.

\bibitem{haris_iqbal_2018_2526396}
H.~Iqbal,
  \href{https://doi.org/10.5281/zenodo.2526396}{Harisiqbal88/plotneuralnet
  v1.0.0} (Dec. 2018).
\newblock \href {http://dx.doi.org/10.5281/zenodo.2526396}
  {\path{doi:10.5281/zenodo.2526396}}.
\newline\urlprefix\url{https://doi.org/10.5281/zenodo.2526396}

\bibitem{tanner1987calculation}
M.~A. Tanner, W.~H. Wong, The calculation of posterior distributions by data
  augmentation, Journal of the American statistical Association 82~(398) (1987)
  528--540.

\bibitem{dozat2016incorporating}
T.~Dozat, Incorporating nesterov momentum into adam.

\bibitem{kendall2018multi}
A.~Kendall, Y.~Gal, R.~Cipolla, Multi-task learning using uncertainty to weigh
  losses for scene geometry and semantics, in: 2018 Conference on Computer
  Vision and Pattern Recognition (CVPR), IEEE, Salt Lake City, USA, 2018, pp.
  7482--7491.

\bibitem{chaumette2006visual}
F.~Chaumette, S.~Hutchinson, Visual servo control. i. basic approaches, IEEE
  Robotics \& Automation Magazine 13~(4) (2006) 82--90.

\bibitem{kingma2014adam}
D.~P. Kingma, J.~Ba, Adam: A method for stochastic optimization (2014).
\newblock \href {http://arxiv.org/abs/1412.6980} {\path{arXiv:1412.6980}}.

\bibitem{Wang2016robot}
Z.~Wang, Z.~Li, B.~Wang, H.~Liu, Robot grasp detection using multimodal deep
  convolutional neural networks, Advances in Mechanical Engineering 8~(9)
  (2016) 1--12.
\newblock \href {http://dx.doi.org/10.1177/1687814016668077}
  {\path{doi:10.1177/1687814016668077}}.

\bibitem{Asif2017rgbd}
U.~Asif, M.~Bennamoun, F.~A. Sohel, Rgb-d object recognition and grasp
  detection using hierarchical cascaded forests, IEEE Transactions on Robotics
  33~(3) (2017) 547--564.
\newblock \href {http://dx.doi.org/10.1109/TRO.2016.2638453}
  {\path{doi:10.1109/TRO.2016.2638453}}.

\bibitem{cbic-paper-97}
R.~Oliveira, E.~Alves, C.~Malqui, Redes neurais convolucionais aplicadas \`{a}
  preensão rob\'{o}tica, in: Anais do 13 Congresso Brasileiro de
  Intelig\^{e}ncia Computacional, ABRICOM, Niterói, RJ, 2017, pp. 1--11.

\bibitem{guo2017robotic}
D.~Guo, F.~Sun, B.~Fang, C.~Yang, N.~Xi, Robotic grasping using visual and
  tactile sensing, Information Sciences 417 (2017) 274--286.

\bibitem{asif2018graspnet}
U.~Asif, J.~Tang, S.~Harrer, Graspnet: An efficient convolutional neural
  network for real-time grasp detection for low-powered devices., in: 27th
  International Joint Conference on Artificial Intelligence, IJCAI, Stockholm,
  Sweden, 2018, pp. 4875--4882.

\bibitem{park2018classification}
D.~Park, S.~Y. Chun, Classification based grasp detection using spatial
  transformer network, arXiv preprint arXiv:1803.01356.

\bibitem{zhang2018roi}
H.~Zhang, X.~Lan, S.~Bai, X.~Zhou, Z.~Tian, N.~Zheng, Roi-based robotic grasp
  detection for object overlapping scenes, arXiv preprint arXiv:1808.10313.

\bibitem{ghazaei2018dealing}
G.~Ghazaei, I.~Laina, C.~Rupprecht, F.~Tombari, N.~Navab, K.~Nazarpour, Dealing
  with ambiguity in robotic grasping via multiple predictions, in: Asian
  Conference on Computer Vision, Springer, Perth, Australia, 2018, pp. 38--55.

\bibitem{chen2019convolutional}
L.~Chen, P.~Huang, Z.~Meng, Convolutional multi-grasp detection using grasp
  path for rgbd images, Robotics and Autonomous Systems 113 (2019) 94--103.

\bibitem{gualtieri2016high}
M.~Gualtieri, A.~Ten~Pas, K.~Saenko, R.~Platt, High precision grasp pose
  detection in dense clutter, in: 2016 IEEE/RSJ International Conference on
  Intelligent Robots and Systems (IROS), IEEE, Daejeon, Korea, 2016, pp.
  598--605.

\bibitem{julian2020stop}
R.~Julian, B.~Swanson, G.~S. Sukhatme, S.~Levine, C.~Finn, K.~Hausman, Never
  stop learning: The effectiveness of fine-tuning in robotic reinforcement
  learning\href {http://arxiv.org/abs/2004.10190} {\path{arXiv:2004.10190}}.

\bibitem{johns2016deep}
E.~Johns, S.~Leutenegger, A.~J. Davison, Deep learning a grasp function for
  grasping under gripper pose uncertainty, in: 2016 IEEE/RSJ International
  Conference on Intelligent Robots and Systems (IROS), IEEE, Daejeon, Korea,
  2016, pp. 4461--4468.

\bibitem{mahler2017dex}
J.~Mahler, J.~Liang, S.~Niyaz, M.~Laskey, R.~Doan, X.~Liu, J.~A. Ojea,
  K.~Goldberg, Dex-net 2.0: Deep learning to plan robust grasps with synthetic
  point clouds and analytic grasp metrics, arXiv preprint arXiv:1703.09312.

\end{thebibliography}

\end{document}